\documentclass[hidelinks]{article}
\newif\ifanonymous

\usepackage[T1]{fontenc}
\usepackage{arxiv}

\anonymousfalse

\usepackage{blindtext}
\usepackage{graphicx}
\usepackage[hidelinks]{hyperref}
\usepackage{url}
\usepackage{optidef}
\usepackage{times}
\usepackage{caption}
\usepackage{subcaption}
\usepackage[super]{nth}
\usepackage{ifdraft}
\usepackage{multirow}
\usepackage{placeins}
\usepackage{comment}
\usepackage{tabularx}
\usepackage[colorinlistoftodos,prependcaption]{todonotes}
\usepackage{amssymb}
\usepackage{algorithm}
\usepackage{algpseudocode}
\usepackage{color}
\usepackage{hyphenat}

\graphicspath{{./figures/}}

\usepackage{amsmath,amsfonts,bm,mathtools,xargs}

\def\Tableref#1{Table~\ref{#1}}
\def\Twotablerefs#1#2{Table~\ref{#1} and~\ref{#2}}

\def\Figref#1{Figure~\ref{#1}}

\def\Twofigref#1#2{Figures~\ref{#1} and~\ref{#2}}
\def\Threefigref#1#2#3{Figures~\ref{#1},~\ref{#2}, and~\ref{#3}}

\def\Figrangeref#1#2{Figures~\ref{#1} to~\ref{#2}}
\def\Figrangetworef#1#2{Figures~\ref{#1}~and~\ref{#2}}

\def\Secref#1{Section~\ref{#1}}

\def\Twosecrefs#1#2{Sections~\ref{#1} and~\ref{#2}}

\def\eqref#1{equation~\ref{#1}}

\def\Eqref#1{Equation~\ref{#1}}

\def\Appref#1{Appendix~\ref{#1}}

\def\Twoappref#1#2{Appendices~\ref{#1} and~\ref{#2}}

\def\1{\bm{1}}

\def\rvmu{{\boldsymbol{\mu}}}

\def\rvx{{\mathbf{x}}}
\def\rvy{{\mathbf{y}}}
\def\rvz{{\mathbf{z}}}

\def\vzero{{\bm{0}}}
\def\vone{{\bm{1}}}

\def\vtheta{{\bm{\theta}}}
\def\vphi{{\bm{\phi}}}

\def\vx{{\bm{x}}}

\def\mA{{\bm{A}}}
\def\mB{{\bm{B}}}
\def\mC{{\bm{C}}}
\def\mD{{\bm{D}}}

\def\mI{{\bm{I}}}

\def\mM{{\bm{M}}}

\def\mQ{{\bm{Q}}}

\def\mX{{\bm{X}}}
\def\mY{{\bm{Y}}}

\DeclareMathAlphabet{\mathsfit}{\encodingdefault}{\sfdefault}{m}{sl}
\SetMathAlphabet{\mathsfit}{bold}{\encodingdefault}{\sfdefault}{bx}{n}

\newcommand{\N}{\mathcal{N}}

\newcommand{\E}{\mathbb{E}}

\newcommand{\R}{\mathbb{R}}

\newcommand{\kld}{D_\mathrm{KL}}
\DeclarePairedDelimiterXPP\kl[2]{\kld}(){}{#1\;\delimsize\|\;#2}
\def\KL{\kl*}

\newcommandx{\Dbig}[4][3,4,usedefault]{D^{#4}_{\mathrm{#3}}\big(#1\;\|\;#2\big)}
\newcommandx{\DBig}[4][3,4,usedefault]{D^{#4}_{\mathrm{#3}}\Big(#1\;\big\|\;#2\Big)}
\newcommandx{\Dbigg}[4][3,4,usedefault]{D^{#4}_{\mathrm{#3}}\bigg(#1\;\Big\|\;#2\bigg)}
\newcommandx{\DBigg}[4][3,4,usedefault]{D^{#4}_{\mathrm{#3}}\Bigg(#1\;\bigg\|\;#2\Bigg)}

\DeclareMathOperator{\ELBO}{\mathcal{L}(\vtheta, \vphi; \rvx)}
\DeclarePairedDelimiter\norm{\lVert}{\rVert}

\newcommand{\coo}{\ensuremath{\mathrm{CO_2}}}
\newcommandx{\suggestion}[2][1=]{\todo[linecolor=ProcessBlue,backgroundcolor=ProcessBlue!25,bordercolor=ProcessBlue,#1]{#2}}
\newcommandx{\donelast}[2][1=]{\todo[linecolor=gray,backgroundcolor=gray!25,bordercolor=gray,#1]{#2}}
\renewcommand\labelenumi{(\roman{enumi})}
\renewcommand\theenumi\labelenumi

\title{How good are variational autoencoders at transfer learning?}

\author{Lisa Bonheme \& Marek Grzes\\
    School of Computing\\
    University of Kent\\
    Canterbury, UK\\
    \texttt{\{lb732, m.grzes\}@kent.ac.uk}
}

\begin{document}
    \maketitle

    \begin{abstract}
        Variational autoencoders (VAEs) are used for transfer learning across various research domains such as music
        generation or medical image analysis. However, there is no principled way to assess before transfer which components
        to retrain or whether transfer learning is likely to help on a target task. We propose to explore this question through
        the lens of representational similarity. Specifically, using Centred Kernel Alignment (CKA) to evaluate the similarity
        of VAEs trained on different datasets, we show that encoders' representations are generic but decoders' specific.
        Based on these insights, we discuss the implications for selecting which components of a VAE to retrain and propose
        a method to visually assess whether transfer learning is likely to help on classification tasks.
    \end{abstract}

    \section{Introduction}\label{sec:intro}

Transfer learning using variational autoencoders (VAEs) is popular in various domains, ranging from music generation to
chemistry~\citep{Inoue2018,Hung2019,Akrami2020,Lovric2021}. However, there is no principled way to assess before transfer which components
to retrain or whether transfer learning is likely to help on a target task.\\

\noindent To bridge this gap, we propose to explore the representational similarity of VAEs.
The domain of deep representational similarity is an active area of research
and metrics such as SVCCA~\citep{Raghu2017,Morcos2018}, Procrustes distance~\citep{Schonemann1966}, or Centred Kernel
Alignment (CKA)~\citep{Kornblith2019} have already proven very useful in analysing the learning dynamics of various
models~\citep{Wang2019a,Kudugunta2019,Raghu2019,Neyshabur2020}. Such metrics could help identify common
representations between models, which could in turn indicate which components of VAEs to retrain.\\

\noindent In this paper, our aim is to use such representational similarity techniques to analyse the representations learned by VAEs
on different datasets, and use these results to provide some insight into the transferability of the representations
learned by VAEs for generation and reconstruction on the target domain. Specifically, based on the results of CKA, we will show that the
representations learned by encoders across a range of source and target datasets are generic while decoders' representations are
specific to the dataset on which they were trained. We will further discuss the implications of these findings for
transfer learning using VAEs and provide a simple method which can be used a priori, without further retraining, to assess
the transferability of latent representations for classification tasks.\\

\noindent Our contributions are as follows:
\begin{enumerate}
    \itemsep 0em
    \item We verify the consistency of CKA for measuring the representational similarity of VAEs by demonstrating that the similarity scores agree with several known properties of VAEs.
    \item We show that encoders' representations are generic but decoders' specific.
    \item Based on these insights, we discuss the implications for selecting the components to retrain depending on the target tasks and propose
    a simple method to visually assess whether transfer learning is likely to help on classification tasks.
\end{enumerate}

    \section{Background}\label{sec:background}

\subsection{Variational Autoencoders}\label{subsec:bg-VAEs}
Variational Autoencoders (VAEs)~\citep{Kingma2013,Rezende2015}
are deep probabilistic generative models based on variational inference.~The encoder, $q_\vphi(\rvz|\rvx)$, maps some input $\rvx$ to
a latent representation $\rvz$, which the decoder, $p_\vtheta(\rvx|\rvz)$, uses to attempt to reconstruct $\rvx$.
This can be optimised by maximising $\mathcal{L}$, the evidence lower bound (ELBO)
\begin{equation}
    \label{eq:elbo}
    \ELBO = \underbrace{\E_{q_\vphi(\rvz|\rvx)}[\log p_\vtheta(\rvx|\rvz)]}_{\text{reconstruction term}} -
    \underbrace{\KL{q_\vphi(\rvz|\rvx)}{p(\rvz)}}_{\text{regularisation term}},
\end{equation}
where $p(\rvz)$ is generally modelled as a multivariate Gaussian distribution $\N(\vzero, \mI)$ to permit closed
form computation of the regularisation term~\citep{Doersch2016}. We refer to the regularisation term of \Eqref{eq:elbo} as regularisation in the rest of the paper, and we do not tune any other forms of regularisation (e.g., L1, dropout).
While our goal is not to study disentanglement, our experiments will focus on a range of VAEs designed to disentangle~\citep{Higgins2017,Chen2018,Burgess2018,Kumar2018}
because they permit to easily increase the regularisation and create posterior collapse. This will be useful to assess the consistency of CKA, as discussed below.
We refer the reader to~\Appref{sec:app-disentanglement} for more details on these VAEs.

\subsubsection*{Polarised regime and posterior collapse}
The polarised regime, also known as selective posterior collapse, is the ability of VAEs to ``shut down''
superfluous dimensions of their sampled latent representations while providing a high precision on the remaining
ones~\citep{Rolinek2019,Dai2020}.
The existence of the polarised regime is a necessary condition for
the VAEs to provide a good reconstruction~\citep{Dai2018,Dai2020}.~However, when
the weight on the regularisation term of the ELBO given in~\Eqref{eq:elbo} becomes too large, the
representations collapse to the prior~\citep{Lucas2019,Dai2020}. Because this behaviour is well-studied, we will use
it to verify the consistency of CKA in~\Secref{subsec:cka-check}.

\subsection{Representational similarity metrics}\label{subsec:bg-similarity}
As stated in~\Secref{sec:intro}, our aim is to study the potential of transfer learning of VAEs using representational
similarity techniques. In this section, we will thus present two well-established metrics that will be used in our experiment.
Representational similarity metrics aim to compare the geometric similarity between two representations.
In the context of deep learning, these representations correspond to $\R^{n \times p}$ matrices of activations, where
$n$ is the number of data examples and $p$ the number of neurons in a layer.~Such metrics can provide various information on
deep neural networks (e.g., the training dynamics of neural networks, common and specialised layers between models).

\subsubsection*{Centred Kernel Alignment} Centred Kernel Alignment (CKA)~\citep{Cortes2012,Cristianini2002} is
a normalised version of the Hillbert-Schmit Independence Criterion (HSIC)~\citep{Gretton2005}. As its name suggests,
it measures the alignment between the $n \times n$ kernel matrices of two representations, and works well with linear kernels~\citep{Kornblith2019} for representational similarity of centred layer activations.
We thus focus on the linear CKA, also known as RV-coefficient~\citep{Escouffier1973,Robert1976}.
Given the centered layer activations $\mX \in \R^{n \times m}$ and $\mY \in \R^{n \times p}$ taken over $n$ data examples,
linear CKA is defined as:
\begin{equation}\label{eq:cka}
    CKA(\mX, \mY) = \frac{\norm{\mY^T\mX}_F^2}{\norm{\mX^T\mX}_F\norm{\mY^T\mY}_F},
\end{equation}
where $\norm{\cdot}_F$ is the Frobenius norm.
CKA is a generalisation of Pearson's correlation coefficient to higher dimensional representations~\citep{Escouffier1973,Robert1976} and can be seen as
measuring the cosine between matrices~\citep{Josse2016}. It takes values between 0 (not similar) and 1 ($\mX = \mY$).
For conciseness, we will refer to linear CKA as CKA in the rest of this paper.

\subsubsection*{Orthogonal Procrustes} The aim of orthogonal Procrustes~\citep{Schonemann1966} is to align a matrix $\mY$ to a matrix $\mX$ using orthogonal transformations $\mQ$
such that
\begin{equation}\label{eq:p-mini}
    \min_{\mQ} \norm{\mX - \mY\mQ}^2_F \quad\mathrm{s.t.}\quad \mQ^T\mQ = \mI.
\end{equation}
The Procrustes distance, $P_d$, is the difference remaining between $\mX$ and $\mY$ when $\mQ$ is optimal,
\begin{equation}\label{eq:pd}
    P_d(\mX, \mY) = \norm{\mX}^2_F + \norm{\mY}^2_F - 2\norm{\mY^T\mX}_*,
\end{equation}
where $\norm{\cdot}_*$ is the nuclear norm (see \cite[pp.~327-328]{Golub2013} for the full derivation from \Eqref{eq:p-mini} to \Eqref{eq:pd}).~To easily compare the results of \Eqref{eq:pd} with CKA,
we first bound its results between 0 and 2 using normalised $\dot{\mX}$ and $\dot{\mY}$, as detailed in \Appref{sec:xp-setup}.
Then, we transform the result to a similarity metric ranging from 0 (not similar) to 1 ($\mX = \mY$),
\begin{equation}\label{eq:ps}
    P_s(\mX, \mY) = 1 - \frac{1}{2} \left(\norm{\dot{\mX}}^2_F + \norm{\dot{\mY}}^2_F - 2\norm{\dot{\mY}^T\dot{\mX}}_*\right).
\end{equation}
We will refer to \Eqref{eq:ps} as Procrustes similarity in the following sections.

\subsection{Limitations of CKA and Procrustes similarities}\label{subsec:bg-limitations}
While CKA and Procrustes lead to accurate results in practice, they suffer from some limitations that need to be taken into account in
our study.~Before we discuss these limitations, we should clarify that, in the rest of this paper, $sim(\cdot, \cdot)$ represents a similarity metric in general, while
$CKA(\cdot, \cdot)$ and $P_s(\cdot, \cdot)$ specifically refer to CKA and Procrustes similarities.

\subsubsection*{Sensitivity to architectures}~\cite{Maheswaranathan2019} have shown that similarity metrics comparing the geometry of representations
were overly sensitive to differences in neural architectures.~As CKA and Procrustes belong to this metrics family,
we can expect them to underestimate the similarity between activations coming from layers of different type (e.g., convolutional
and deconvolutional).

\subsubsection*{Procrustes is sensitive to the number of data examples}
\begin{figure}[ht!]
    \centering
    \subcaptionbox{$sim(\mA,\mB)$\label{fig:sim}}{
        \includegraphics[width=0.3\linewidth]{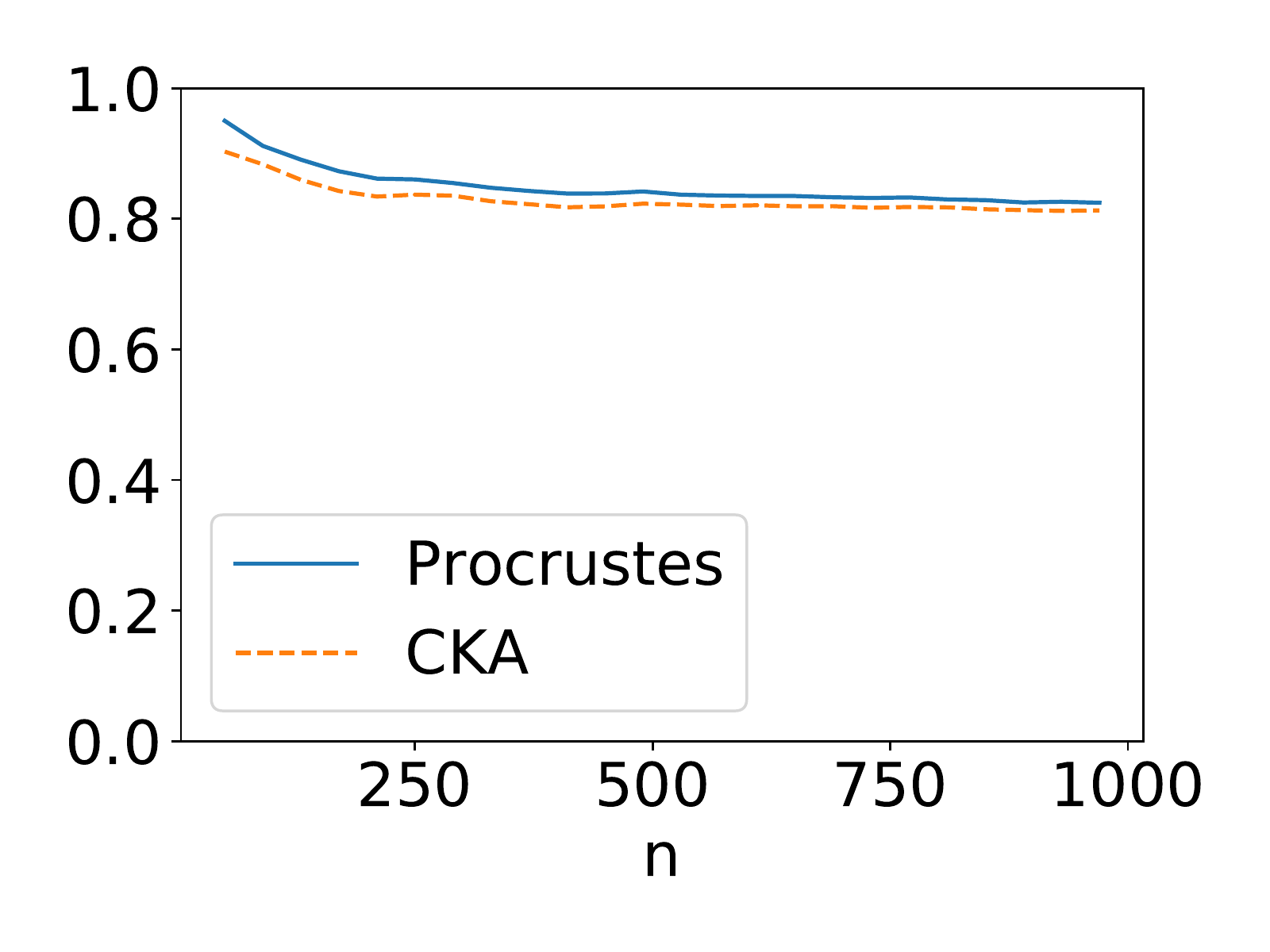}
    }%
    \hfill
    \subcaptionbox{$sim(\mA,\mC)$\label{fig:mid-sim}}{
        \includegraphics[width=0.3\linewidth]{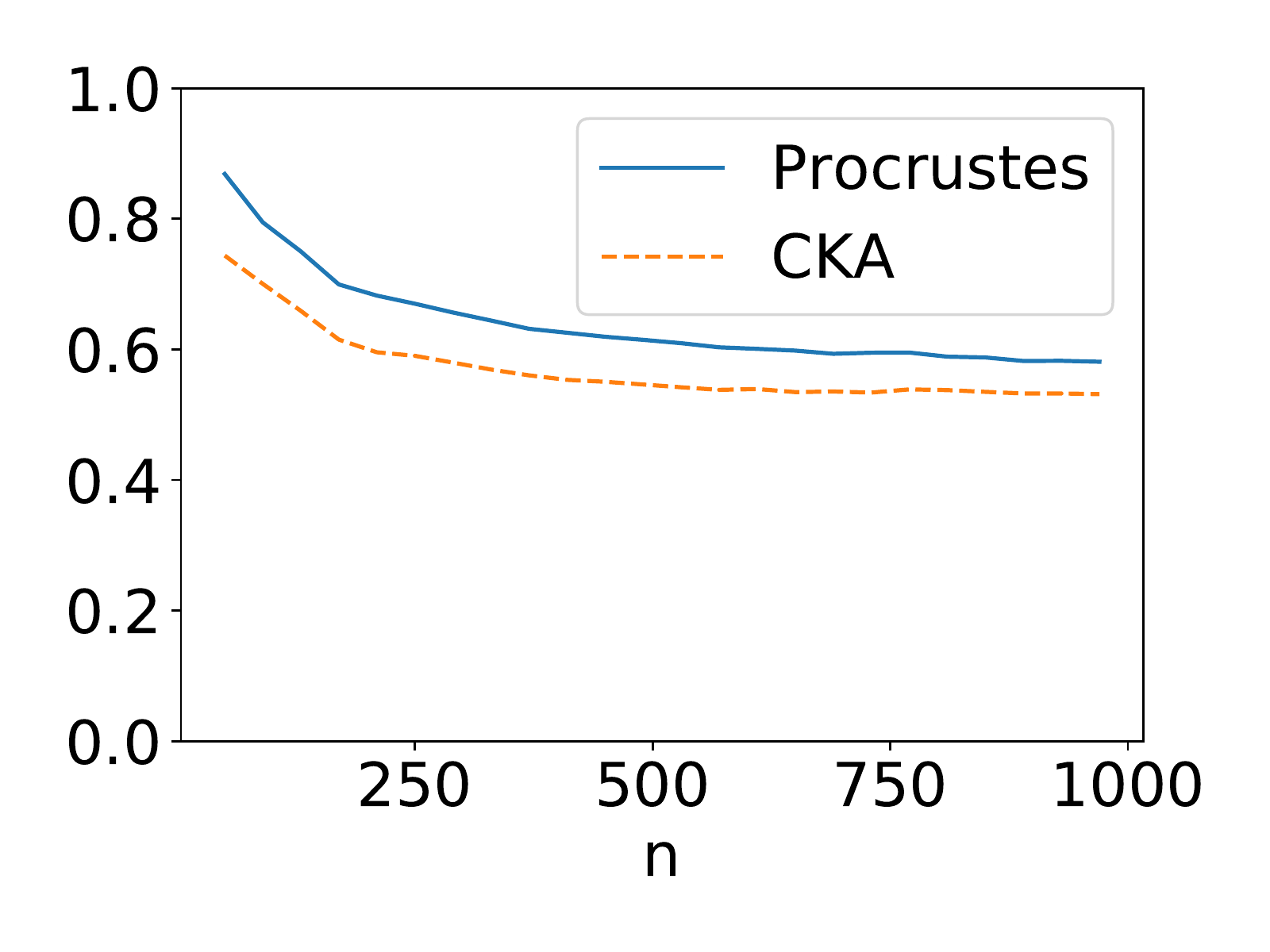}
    }%
    \hfill
    \subcaptionbox{$sim(\mA,\mD)$\label{fig:diff}}{
        \includegraphics[width=0.3\linewidth]{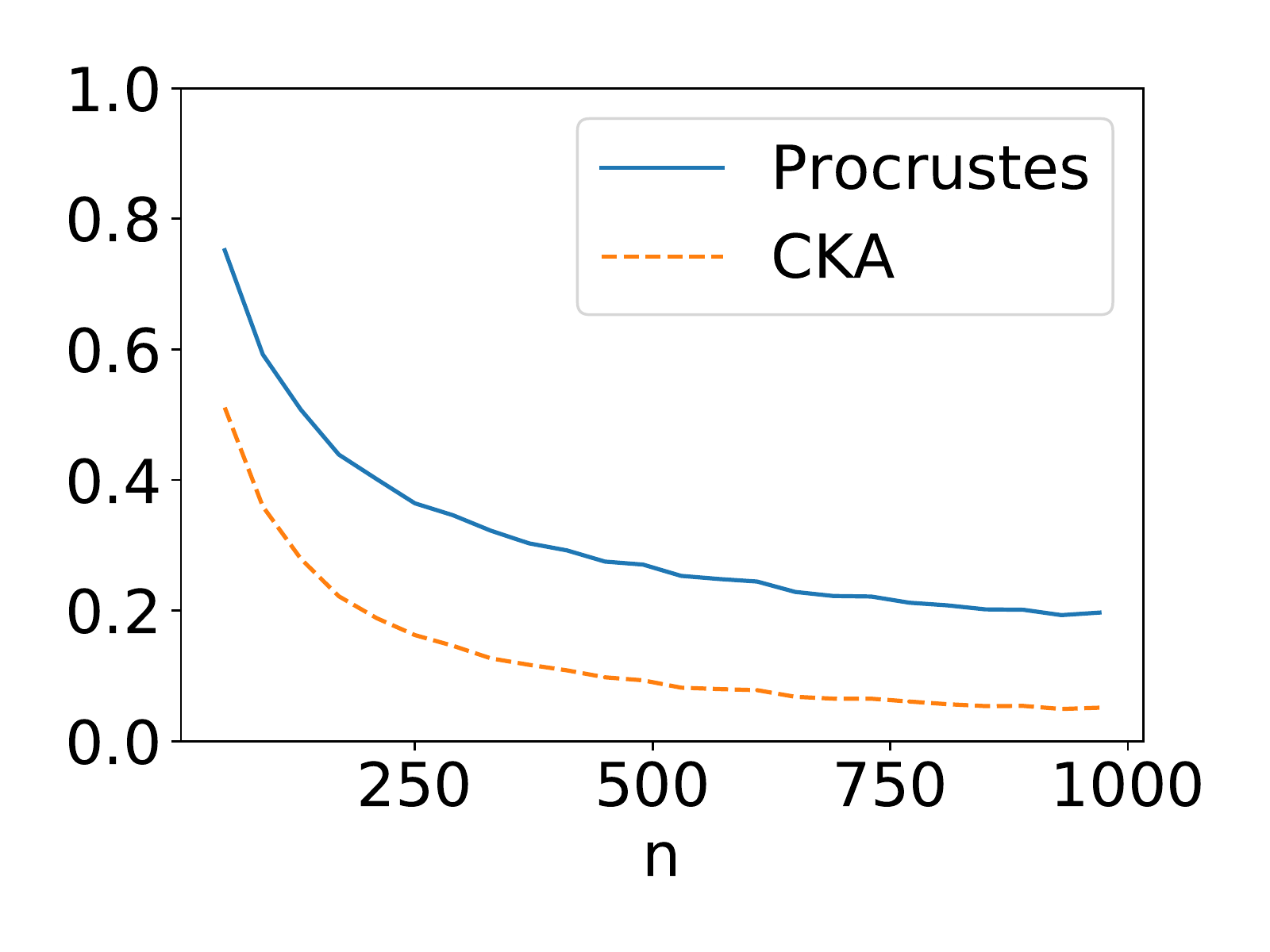}
    }%
   \caption{We compute CKA and Procrustes similarity scores with an increasing number of data examples $n$, and different
    similarity strength: $\mB$ retains 80\% of $\mA$'s features, $\mC$ 50\%, and $\mD$ 0\%.~Both metrics agree in
    (a) and (b), but Procrustes overestimates similarity in (c).}
    \label{fig:cka-procrustes}
\end{figure}

As we may have representations with high dimensional features (e.g., activations of convolutional layers),
we checked the impact of the number of data examples on CKA and Procrustes.
To do so, we created four increasingly different matrices
$\mA, \mB, \mC$, and $\mD$ with 50 features each: $\mB$ retains 80\% of $\mA$'s features, $\mC$ 50\%, and $\mD$ 0\%.
~We then computed the similarity scores given by CKA and Procrustes while varying the number
of data examples.~As shown in \Twofigref{fig:sim}{fig:mid-sim}, both metrics agree for $sim(\mA,\mB)$ and $sim(\mA,\mC)$,
giving scores that are close to the fraction of common features between the two matrices.
However, we can see in~\Figref{fig:diff} that Procrustes highly overestimates
$sim(\mA, \mD)$ while CKA scores rapidly drop.

\subsubsection*{CKA ignores small changes in representations}
When considering a sufficient number of data examples for both Procrustes and CKA, if two representations do not have
dramatic differences (i.e., their 10\% largest principal components are the same),
CKA may overestimate similarity, while Procrustes remains stable, as observed by~\cite{Ding2021}.

\subsubsection*{Ensuring accurate analysis}~Given the limitations previously mentioned, we take three remedial actions
to guarantee that our analysis is as accurate as possible.
Firstly, as both metrics will likely underestimate the similarity between different layer types,
we will only discuss the variation of similarity when analysing such cases.
For example, we will not compare $sim(\mA, \mB)$ and $sim(\mA, \mC)$ if $\mA$ and $\mB$ are convolutional layers but
$\mC$ is deconvolutional.~We will nevertheless analyse the changes of $sim(\mA, \mC)$ at different steps of training.
Secondly, when both metrics disagree, we know that one of them is likely overestimating the similarity: Procrustes if
the number of data examples is not sufficient, CKA if the difference between the two representations is not large enough.
~Thus, we will always use the smallest of the two results for our interpretations.

\subsection{Transfer learning}
Transfer learning is the process of reusing knowledge learned from
one or more source domains to improve the performance of a model on a target domain~\citep{Pan2009}. Each domain
$\mathcal{D}$ is composed of a feature space $\mathcal{X}$ and a marginal probability distribution $p(\mX)$ where $\mX = \{\rvx^{(i)}\}_{i=1}^n$
such that $\mathcal{D} \triangleq \{\mathcal{X},p(\mX)\}$.

\subsubsection*{Types of transfer learning} Depending on the nature of the domains and considered tasks, transfer learning
can be decomposed into multiple subcategories (see~\citep{Pan2009} for a detailed overview). In this study, we are interested
in settings where the source and target domains are different but related and no classification or regression labels are available in the source domain.
Following~\cite{Pan2009}, this corresponds to self-taught learning~\citep{Raina2007} when the target domain contains labeled data and unsupervised
learning otherwise.

\subsubsection*{Transfer learning with VAEs} Most of the research on transfer learning using VAEs focuses on how to efficiently
perform transfer learning on specific applications. However, there is no clearly defined method to decide a priori which components should be retrained in which context.
For example~\cite{Lovric2021} directly reuse the encoder learned on the source dataset in a target task on a different domain,
\cite{Inoue2018} only retrain the encoder on the target domain, and~\cite{Hung2019,Akrami2020} fine-tune the entire model.
In~\Secref{subsec:cka-tl}, we will assess the specificity of the representations learned by VAEs using CKA.
From this analysis, we will provide guidelines on which components should be retrained depending on the type of the target
task. To assess whether transfer learning could be beneficial for a target task whose labels are known,
we will also propose a method to visually identify shared latent variables between source and target domains.

    \section{Experimental setup}\label{sec:experiment}
As stated in \Secref{sec:intro}, the objectives of this experiment are 1) to estimate which components of the model would require retraining for the target task, and
2) to assess the transferability of the learned representations for self-taught transfer learning.
To do so, in \Secref{subsec:cka-check}, we first ensure the chosen representational similarity metric is consistent with known facts about
the learning dynamics of VAEs. Then, in \Secref{subsec:cka-tl}, we assess the representational similarity of models learned
on source and target domains when evaluated on target instances.
Based on these observations, in \Secref{subsec:cka-tl-impl} we provide some guidelines on which components to retrain depending on the target task, fulfilling our first objective.
Then, we propose a method to visually identify shared variables between the source and target domain that are learned by VAEs, adressing our second objective.
Finally, we confirm the validity of the visual analysis by comparing the observations with the results of classification on the target domain using latent representations
learned from the source domain without any fine-tuning.

\subsubsection*{Learning objectives} We generally use vanilla VAEs except in \Secref{subsec:cka-check}, when comparing the representational similarity of
models across different learning objectives and regularisation strength. In this case, we use learning objectives
which allow for easy tuning of the ELBO's regularisation strength, namely $\beta$-VAE~\citep{Higgins2017}, $\beta$-TC VAE~\citep{Chen2018}, Annealed VAE~\citep{Burgess2018}, and DIP-VAE II~\citep{Kumar2018}.
A description of these methods can be found in~\Appref{sec:app-disentanglement}.
~To provide fair and complementary insights into previous observations of such models~\citep{Locatello2019a, Bonheme2021},
we will follow the experimental design of~\cite{Locatello2019a} regarding the architecture, learning objectives, and regularisation used.
Moreover, \texttt{disentanglement lib}\footnote{\url{https://github.com/google-research/disentanglement_lib}} will be used as a codebase for our experiment.
The complete details are available in \Appref{sec:xp-setup}.

\subsubsection*{Datasets} For~\Secref{subsec:cka-check}, we use three datasets which, based on the results of~\cite{Locatello2019a}, are increasingly difficult for VAEs in terms of reconstruction loss:
dSprites\footnote{Licensed under an Apache 2.0 licence.}~\citep{Higgins2017}, Cars3D \citep{Reed2015}, and SmallNorb~\citep{LeCun2004}.
For~\Secref{subsec:cka-tl}, we additionally use Symsol\_reduced~\citep{Bonheme2022b} and Celeba~\citep{Liu2015} to create
increaslingly challenging transfer learning configurations using the two pairs of datasets (dSprites, Symsol) and (Cars3D, Celeba).

\subsubsection*{Training process} For~\Secref{subsec:cka-check}, we trained five models with different initialisations for 300,000 steps for each (learning objective, regularisation strength, dataset)
triplet, and saved intermediate models to compare the similarity within individual models at different epochs.
\Appref{sec:app-epochs} explains our epoch selection methodology. We further trained five classical VAEs with different
initialisations for 300,000 steps on Celeba and Symsol for~\Secref{subsec:cka-tl}.

\subsubsection*{Similarity measurement} For every dataset, we sampled 5,000 data examples, and we used them to compute
all the similarity measurements.
We compute the similarity scores between all pairs of layers of the different models following the different combinations outlined above.
As Procrustes similarity takes significantly longer to compute compared to CKA (see below), we only used
it to validate CKA results, restricting its usage to one dataset: Cars3D.
We obtained similar results for the two metrics on Cars3D, thus we only reported CKA results in the main paper.
Procrustes results can be found in~\Appref{sec:procrustes}.
%

    \section{Results}\label{sec:results}

In this section, we will discuss the similarity between models
through the heatmaps obtained with CKA. The two models being compared will be described in the $x$ and $y$ axis, and
the results will be averaged over 5 runs of each model. Specifically, given the $i^{th}$ run of two models $A_i$
and $B_i$ and their activations $\mM_j^{A_i}$ and $\mM_j^{B_i}$ over $n$ examples at layer $j$,
the value displayed at the cell $(j,k)$ of the heatmap corresponds to the averaged CKA scores between the $j^{th}$
layer of $A$ and the $k^{th}$ layer of $B$:
\begin{equation}\label{eq:cka-avg}
    CKA_{avg}(\mM_j^A, \mM_k^B) = \frac{1}{25}\sum_{i=1}^{5} \sum_{l=1}^{5} CKA(\mM_j^{A_i},\mM_k^{B_l}).
\end{equation}
When describing these figures, we will refer to top-left (resp. bottom-right) quadrants to indicate the similarity
scores between all the representations of the encoder (resp. decoder). This includes the off-diagonal CKA scores between layers of the
same type. Similarly we will refer to the top-right (resp. bottom-left) quadrants when comparing the representations
learned by the encoder and decoder of two models. Note that in this case, we will always dicuss the difference of scores
between both models, (i.e., both quadrants will be compared). Indeed the layers of the encoder and decoder are of
different type and discussing the scores obtained for only one quadrant without contrasting it with the other
may be misleading as explained in~\Secref{subsec:bg-similarity}.

\subsection{Assessing the coherence of CKA scores with known facts about VAEs}\label{subsec:cka-check}
The goal of this section is to verify that CKA can provide accurate information about the learning dynamics of VAEs.
We thus check that the results observed using representational similarity are consistent with known facts about VAEs.
Note that we obtained similar results using Procrustes similarity and fully connected neural network architectures,
as reported in~\Twoappref{sec:procrustes}{sec:app-fc}.

\begin{figure}[ht!]
    \centering
    \subcaptionbox{Trained on Cars3D\label{fig:fact1-1}}{
        \includegraphics[width=0.33\textwidth]{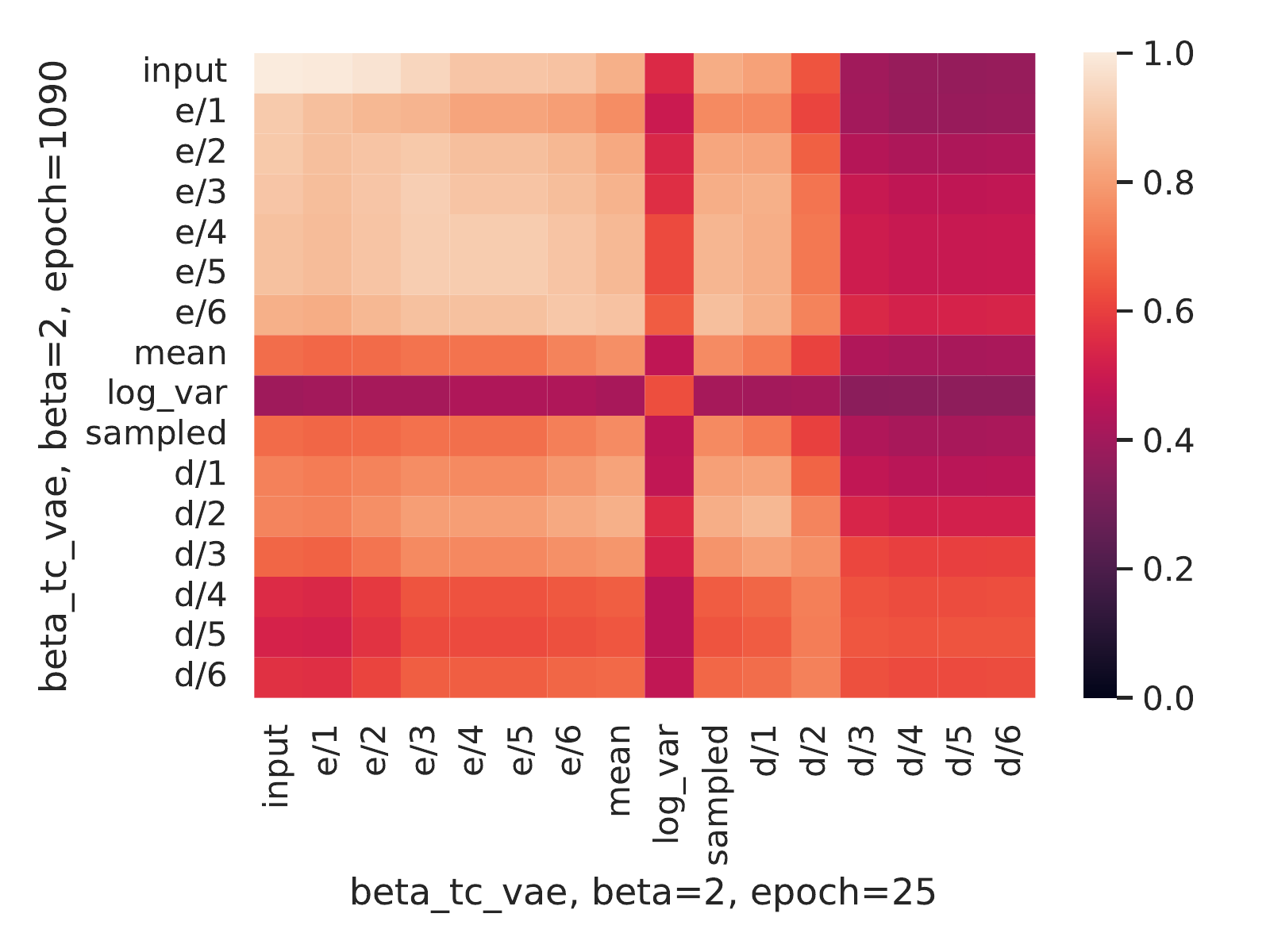}
    }%
    \subcaptionbox{Trained on dSprites\label{fig:fact1-2}}{
        \includegraphics[width=0.33\textwidth]{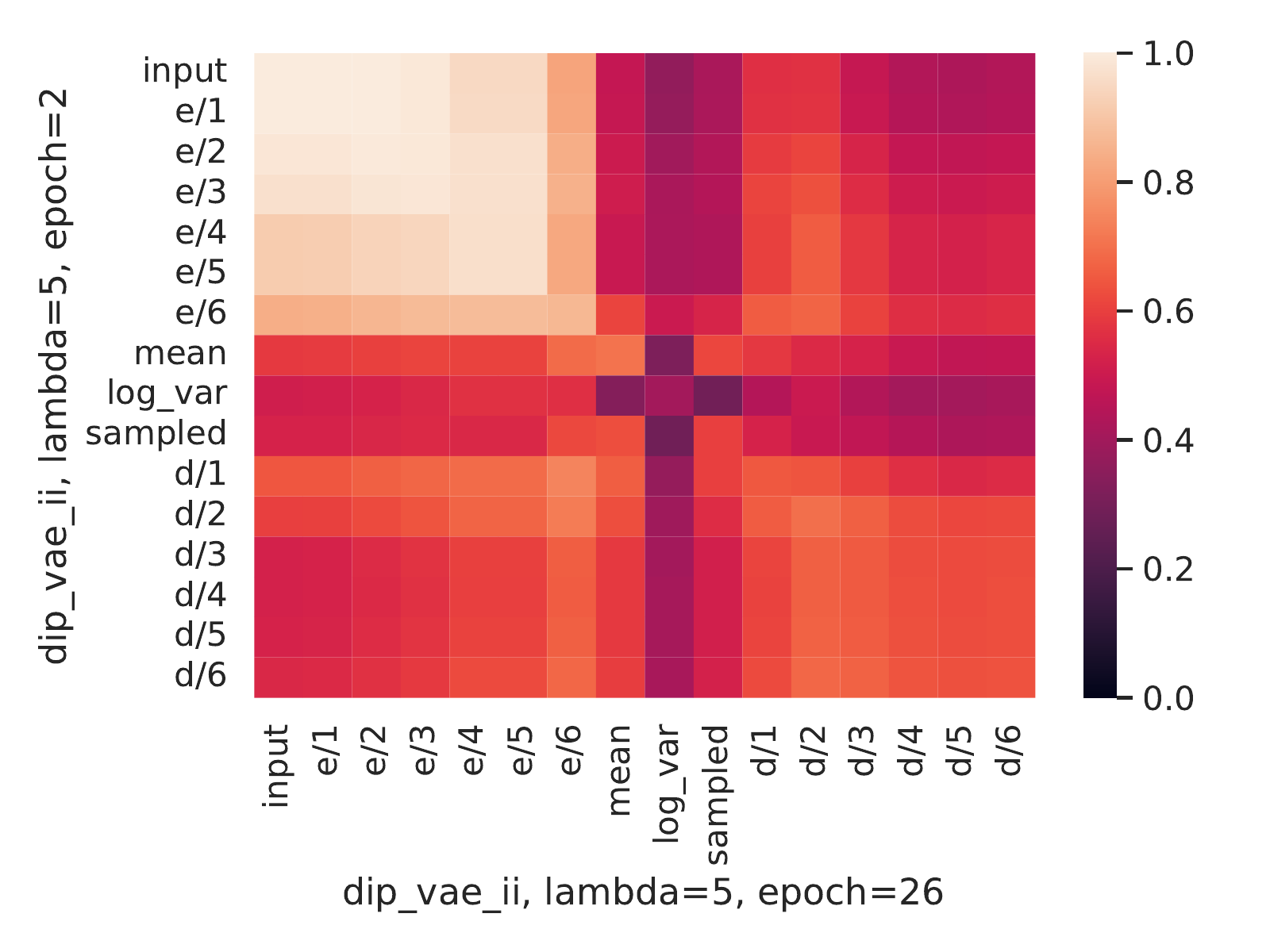}
    }%
    \subcaptionbox{Trained on SmallNorb\label{fig:fact1-3}}{
        \includegraphics[width=0.33\textwidth]{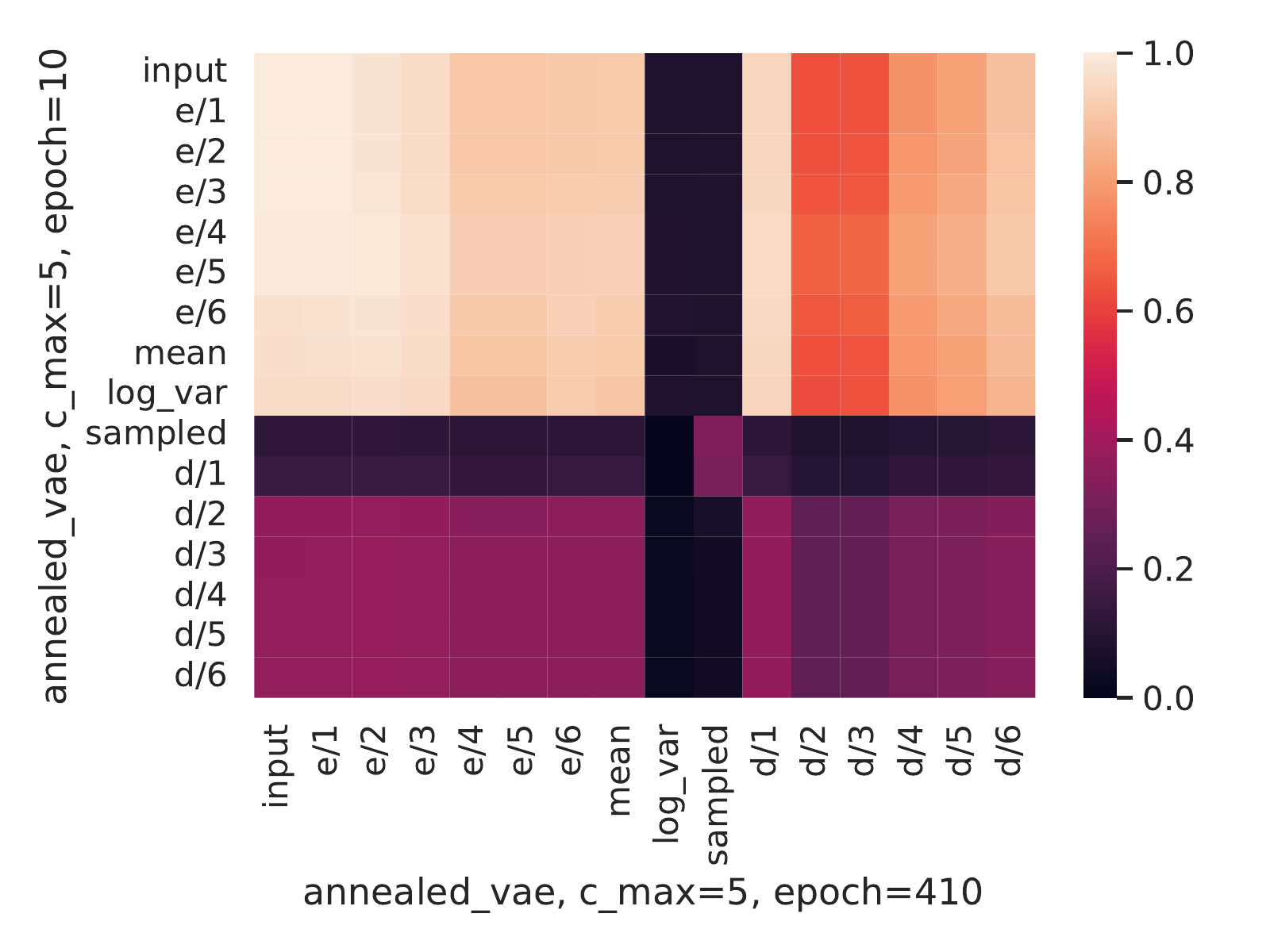}
    }%
    \caption{(a) shows the CKA similarity scores of activations at epochs 25 and 1090 of $\beta$-TC VAE trained on Cars3D with $\beta=2$.
        (b) shows the CKA similarity scores of activations at epochs 2 and 26 of DIP-VAE II trained on dSprites with $\lambda=5$.
        (c) shows the CKA similarity scores of activations at epochs 10 and 410 of Annealed VAE trained on SmallNorb with $c_{max}=5$.}\label{fig:fact1}
\end{figure}

\subsubsection*{Fact 1: the encoder is learned before the decoder}~Using the information bottleneck (IB) theory,
\citep{Lee2021} have shown that in VAEs, the encoder is learned before the decoder.~Moreover, this behaviour seems to be
required for VAEs to learn meaningful representations as decoders which ignore the latent representations (e.g.,
because of posterior collapse or lagging inference) provide suboptimal reconstructions~\citep{Bowman2016,He2019}.
When comparing the representations learned at different epochs in~\Figref{fig:fact1}, we can see that
CKA provides consistent results about this phenomenon: the encoder is learned first, and the representations of its
layers become similar to the input after a few epochs (see the bright cells in the top-left quadrants in
\Threefigref{fig:fact1-1}{fig:fact1-2}{fig:fact1-3}).
The decoder then progressively learns representations that gradually become closer to the input while the mean and
variance representations are refined (see the dark cells in the bottom-right quadrant of
\Threefigref{fig:fact1-1}{fig:fact1-2}{fig:fact1-3}).~Note that our choice of snapshots and snapshot frequency did not
influence the results as verified in~\Twoappref{sec:app-epochs}{sec:app-convergence}.

\begin{figure}[ht!]
    \centering
    \subcaptionbox{$\beta$-VAE\label{fig:fact2-1}}{
        \includegraphics[width=0.35\textwidth]{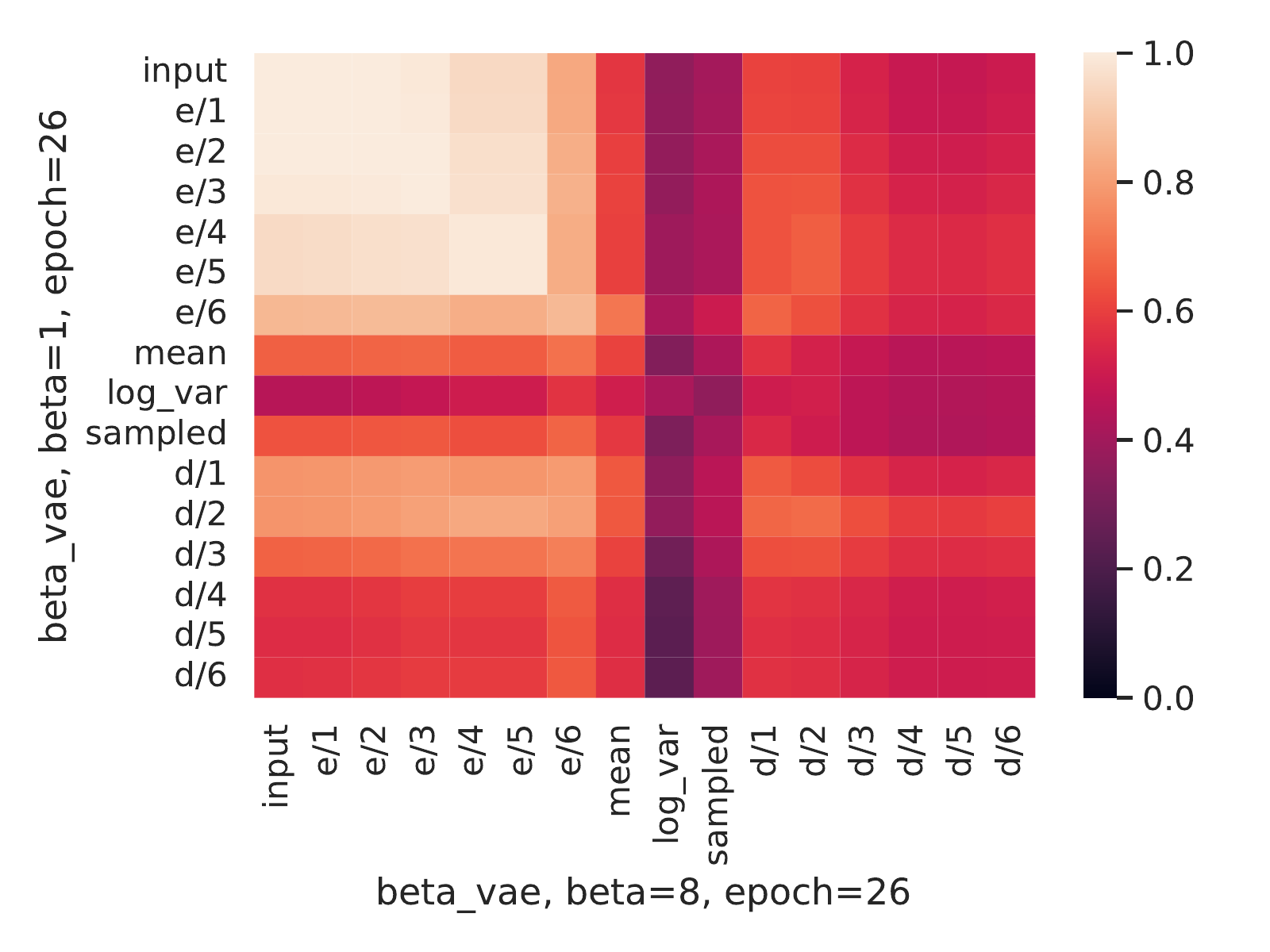}
    }%
    \hfill
    \subcaptionbox{DIP-VAE II\label{fig:fact2-2}}{
        \includegraphics[width=0.35\textwidth]{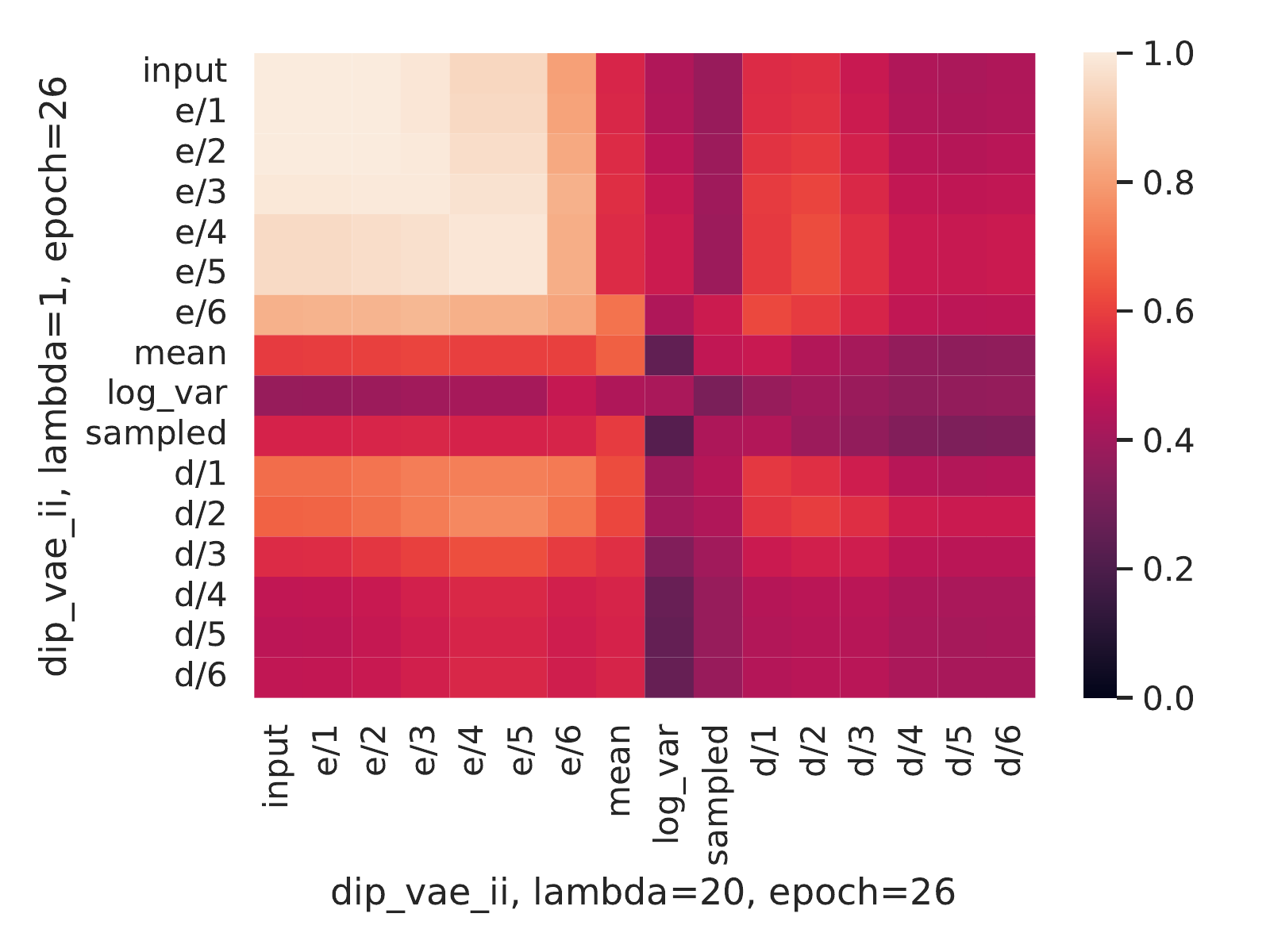}
    }%
    \caption{(a) shows the CKA similarity scores between the activations of two $\beta$-VAEs trained on dSprites with $\beta=1$, and $\beta=8$.
        (b) shows the CKA similarity scores between the activations of two DIP-VAE II trained on dSprites with $\lambda=1$, and $\lambda=20$.
        For both figures, the activations are taken after complete training.}
    \label{fig:fact2}
\end{figure}

\subsubsection*{Fact 2: very high regularisation leads to posterior collapse} It is well known that an excessively high
pressure on the regularisation term of the ELBO in~\Eqref{eq:elbo} leads to posterior
collapse~\citep{Dai2018,Lucas2019,Lucas2019b,Dai2020}.~When this happens, the sampled representation collapses to the prior
--- generally $\N(\vzero, \mI)$ --- and the decoder has a poor reconstruction quality.~This phenomenon is clearly visible with CKA
in~\Figref{fig:fact2}.~Indeed, the sampled representations of the collapsed model (dark line at the ``sampled''
column of~\Twofigref{fig:fact2-1}{fig:fact2-2}) have a very low similarity with the representations learned by the
encoder and decoder of a well-behaved model, in opposition to the sampled representation of a well-behaved model
(lighter line at the ``sampled'' row of~\Twofigref{fig:fact2-1}{fig:fact2-2}).
This indicates that the collapsed sampled representations do not retain any information about the input,
in opposition to any layer of a well-behaved model.

\begin{figure}[ht!]
    \centering
    \subcaptionbox{Trained on Cars3D\label{fig:fact3-1}}{
        \includegraphics[width=0.33\textwidth]{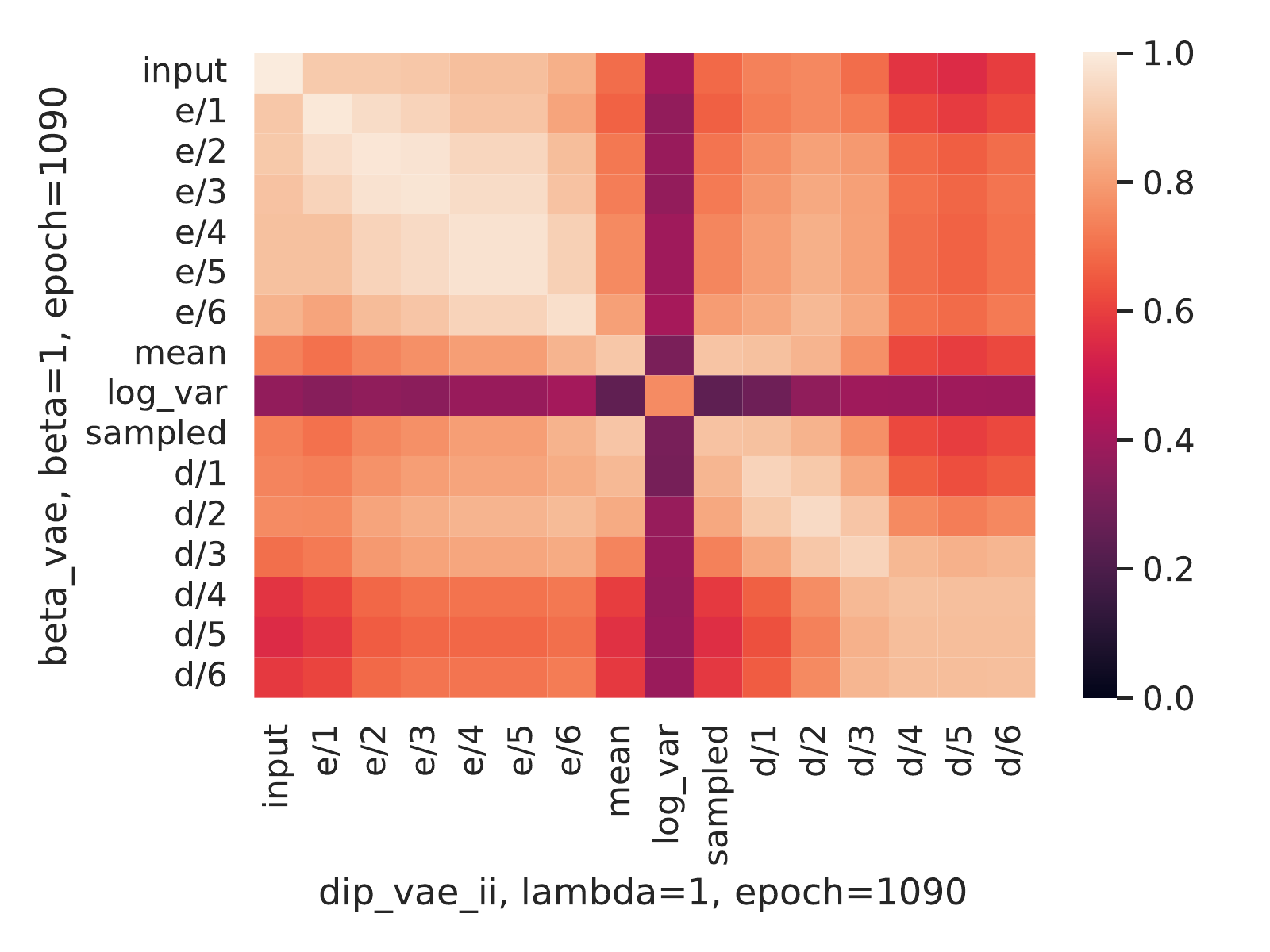}
    }%
    \subcaptionbox{Trained on dSprites\label{fig:fact3-2}}{
        \includegraphics[width=0.33\textwidth]{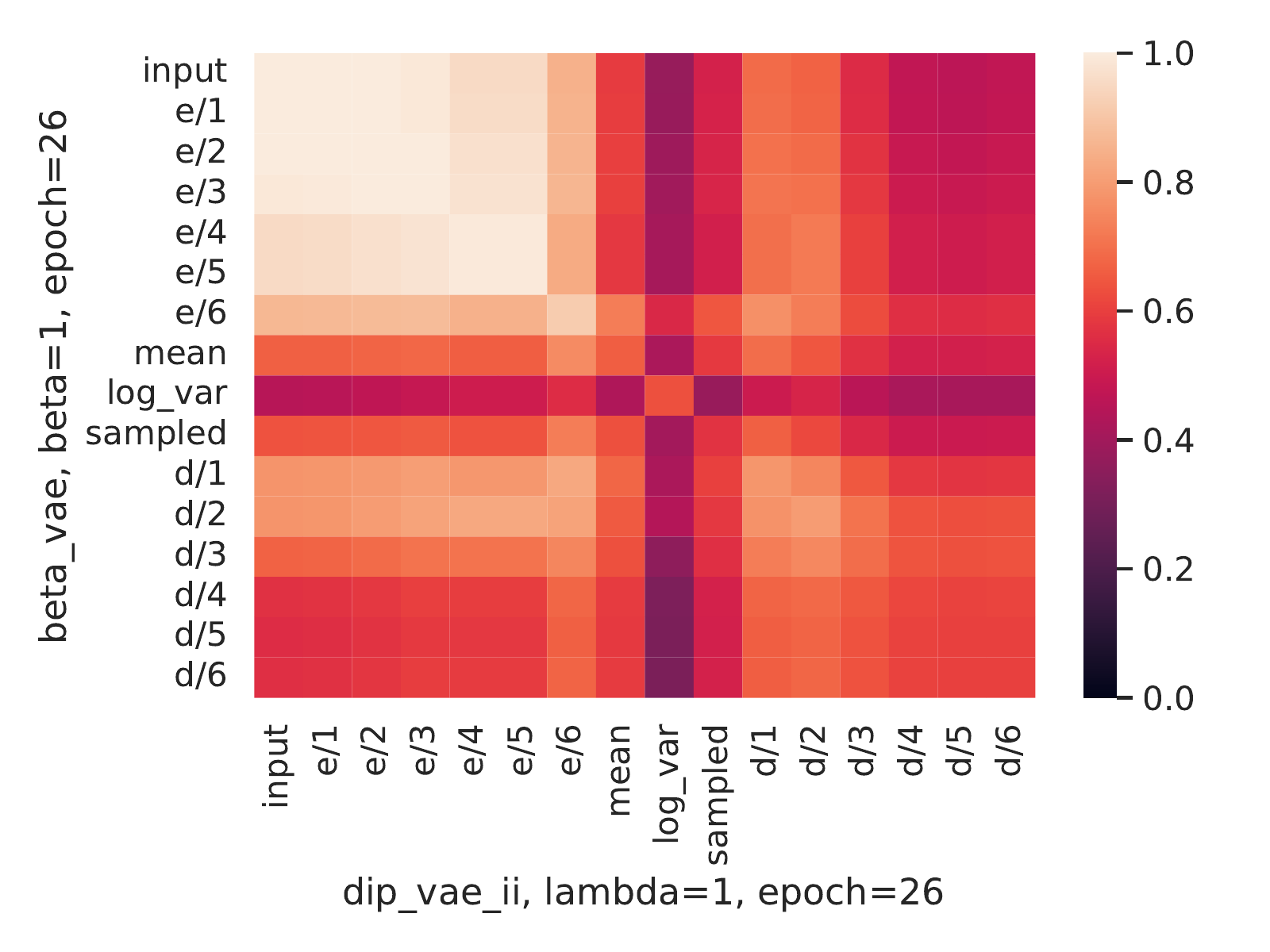}
    }%
    \subcaptionbox{Trained on SmallNorb\label{fig:fact3-3}}{
        \includegraphics[width=0.33\textwidth]{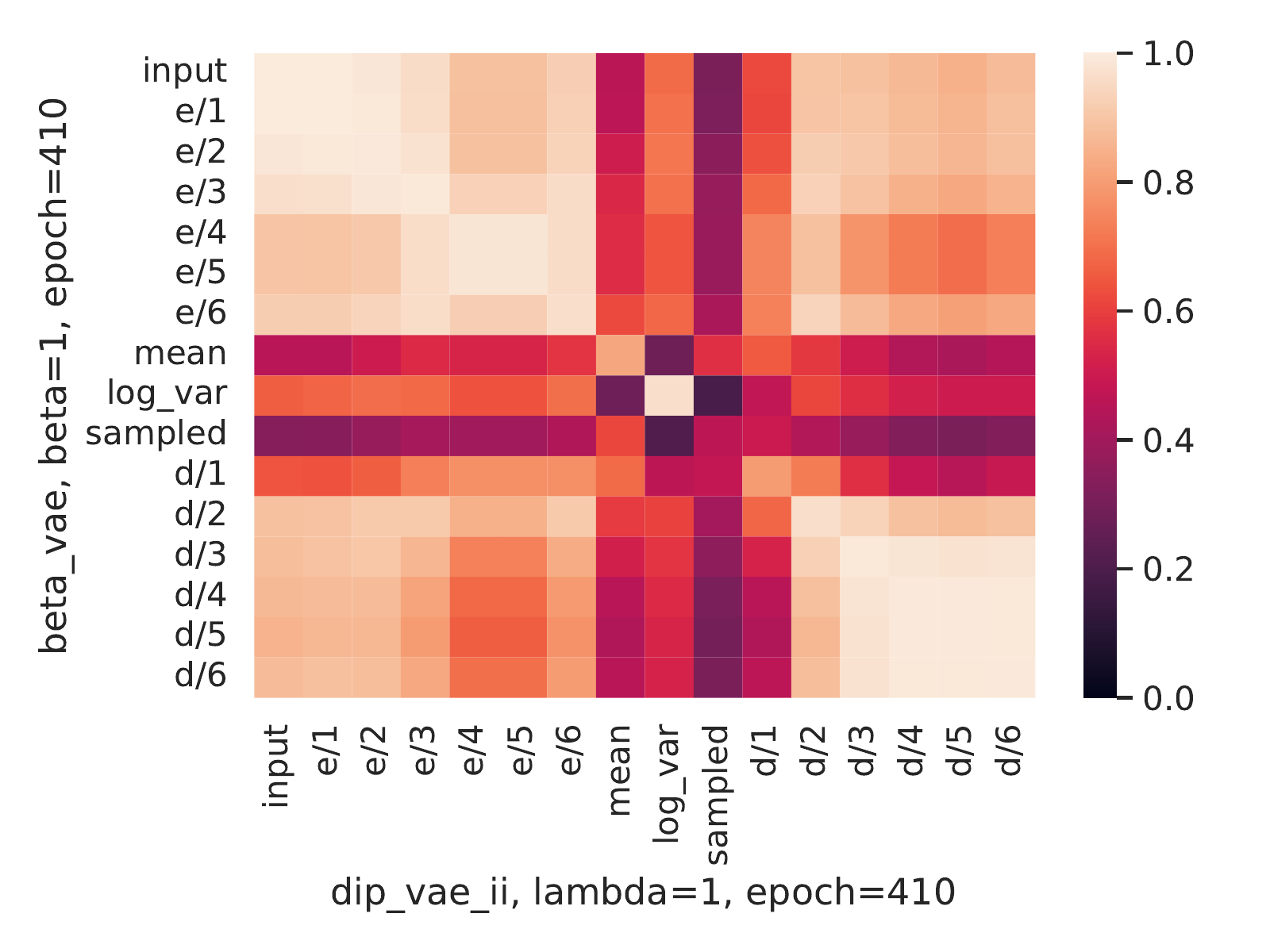}
    }%
    \caption{(a) shows the CKA similarity scores of activations of $\beta$-VAE and DIP-VAE II trained on Cars3D with $\beta=1$, and $\lambda=1$, respectively.
             (b) and (c) show the CKA similarity scores of the same learning objectives and regularisation strengths but trained on dSprites and SmallNorb.
    }
    \label{fig:fact3}
\end{figure}

\subsubsection*{Fact 3: encoders learn abstract features}
It is not uncommon amongst practitioners to apply transfer learning to VAEs by using a pre-trained classifier
as an encoder.~The last layers (i.e., closest to the output and used for classification)
are removed and replaced by the mean and variance layers.~The assumption, which underlies transfer learning,
is that encoders learn abstract features which are shared with other types of network~\citep{Yosinski2014,Bansal2021,Csiszarik2021}.
The results of CKA are also in line with this fact.~Indeed, the bright top-left quadrants of~\Threefigref{fig:fact3-1}{fig:fact3-2}{fig:fact3-3}, indicate
that encoders of VAEs trained using different learning objectives are highly similar.~This observation
holds when comparing encoders and classifiers with equivalent architectures (see~\Appref{sec:app-clf}).

\subsection{The representational similarity of VAEs trained on different domains}\label{subsec:cka-tl}

We have seen in~\Secref{subsec:cka-check} that the representations learned by the encoder of VAEs with different learning
objectives and by classifiers with the same architecture were similar except for the mean and variance representations.
Thus, encoders can be thought of as generic feature extractors.~In this section, we are going one step further and study how
transferable the representations learned by VAEs are between domains. Specifically, how the representations learned on
different but related domains differ and how this informs the potential of VAEs for transfer learning.

\begin{figure}[ht!]
    \centering
    \subcaptionbox{No fine-tuning.\label{fig:cka-td-raw}}{
        \includegraphics[width=0.45\textwidth]{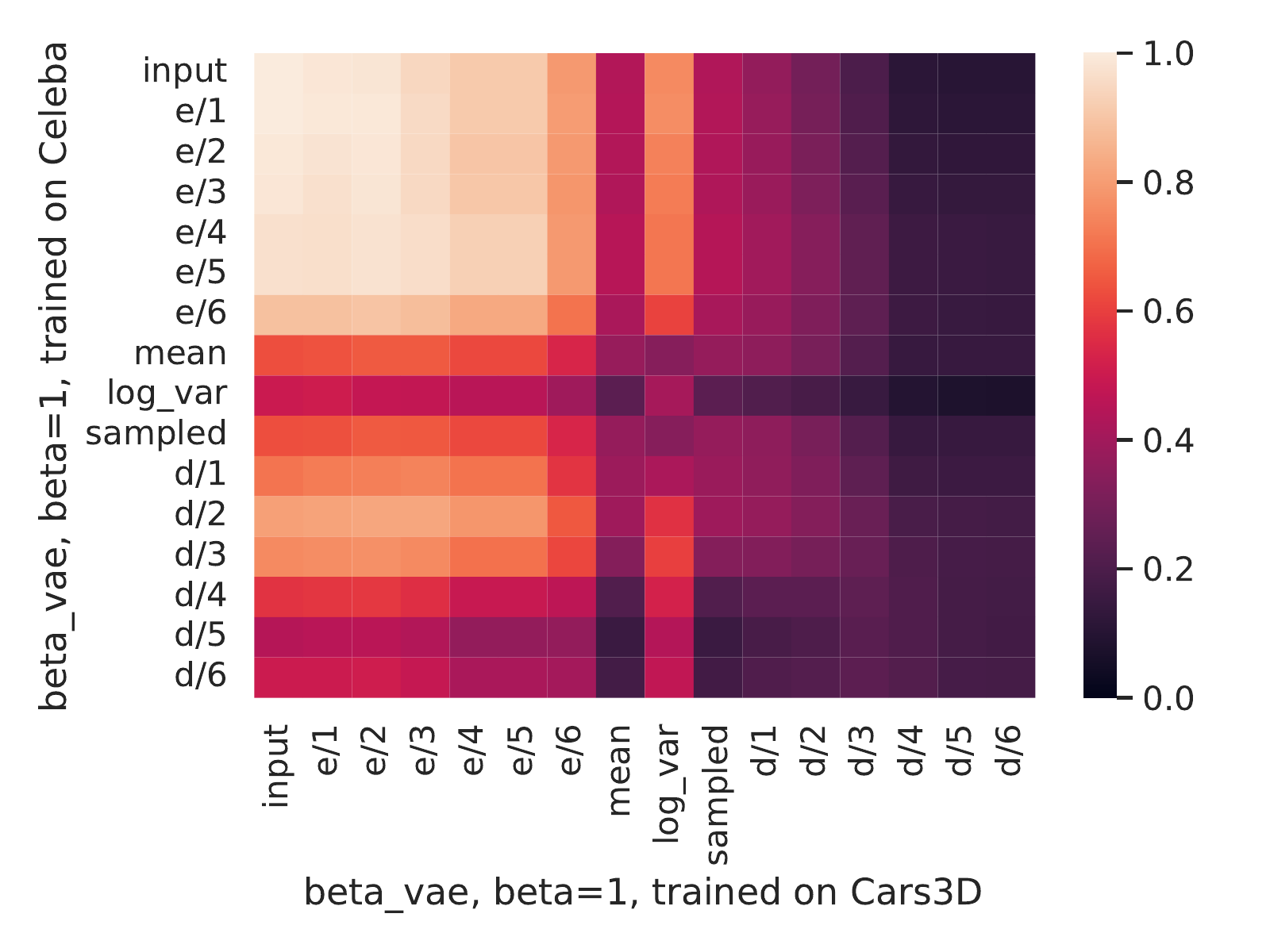}
    }%
    \hfill
    \subcaptionbox{Mean and variance are retrained.\label{fig:cka-td-ft}}{
        \includegraphics[width=0.45\textwidth]{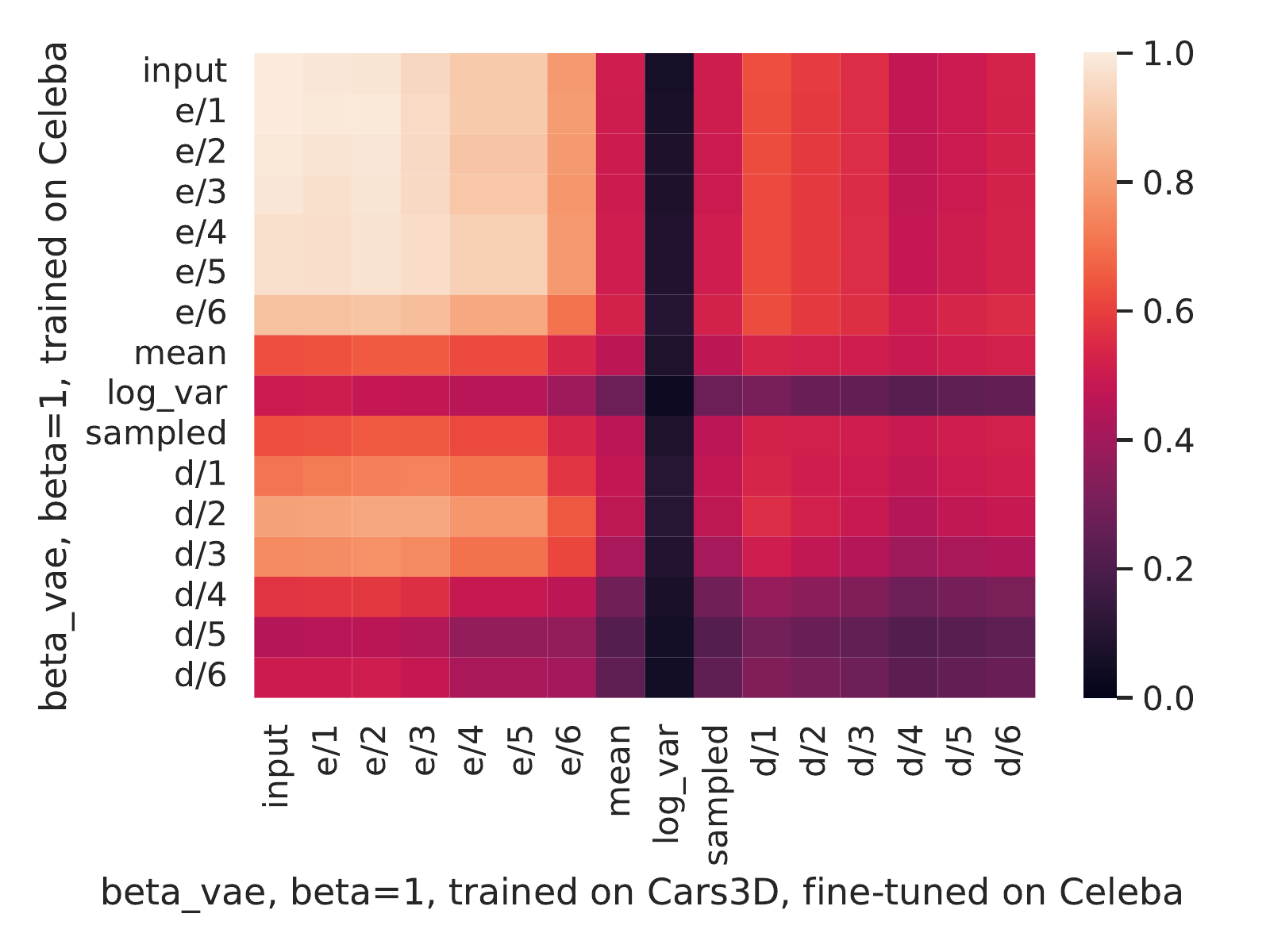}
    }%
    \caption{Both figures show CKA similarity scores between the activations of two VAEs evaluated on the target dataset $\mX_t=$ Celeba but trained on different datasets.
    The models on the x-axis are trained on $\mX_s=$ Cars3D while the models on the y-axis are trained on the target dataset $\mX_t=$ Celeba.
    In (a) the model on the x-axis is evaluated on Celeba without any fine-tuning, in (b) its mean and variance are retrained on Celeba beforehand.}
    \label{fig:cka-td}
\end{figure}

\subsubsection*{Encoders learn generic representations} When assessed on a target dataset $\mX_t$,
the representations learned by an encoder trained on a source dataset $\mX_s$
retain a high similarity with those obtained from a VAE directly trained on $\mX_t$, as demonstrated by the bright
top-left quadrants of~\Figref{fig:cka-td-raw}. In fact the similarity is as high as what is observed on encoders trained on
the same dataset in~\Figref{fig:fact3}. This shows that before the mean and variance layers, the encoders extract features
that are sufficiently abstract to be shared between domains (e.g., curves, edges, frequencies)~\citep{Yosinski2014,Razavian2014}.
Indeed, the source and target images are very different and a high specificity of the features encoded would lower the similarity
score.

\subsubsection*{Decoders learn specific representations} In opposition to the general domain\hyp{}independent representations of encoders, the
dark bottom-right quadrant of~\Figref{fig:cka-td-raw} shows that the decoder's representations are very different if they
were initially trained on the source or target dataset. One could argue that decoders also learn generic features
but have different activations because the input $\rvz$ is different between the models.
However, after retraining the mean and sampled representations of a VAE trained on the source domain,
in~\Figref{fig:cka-td-ft} we still obtain a bottom-right quadrant noticeably darker than for decoders trained on the same dataset in~\Figref{fig:fact3},
indicating that in opposition to encoders' representations, decoders' representations are far more specific to the dataset on which they were trained.

\begin{figure}[ht]
    \centering
    \subcaptionbox{Fixed encoder\label{fig:gen-dec-1}}{
        \includegraphics[width=0.48\textwidth]{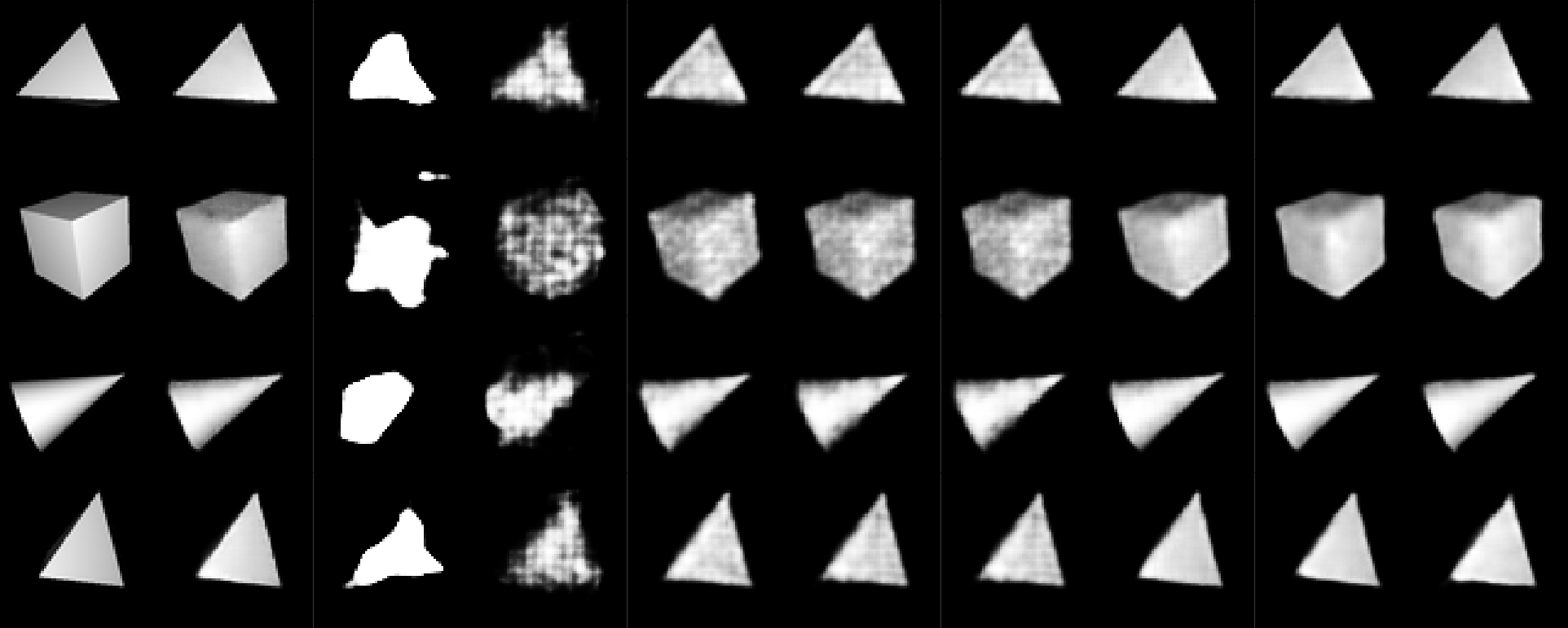}
    }%
    \hfill
    \subcaptionbox{Retrained mean and variance\label{fig:gen-dec-2}}{
        \includegraphics[width=0.48\textwidth]{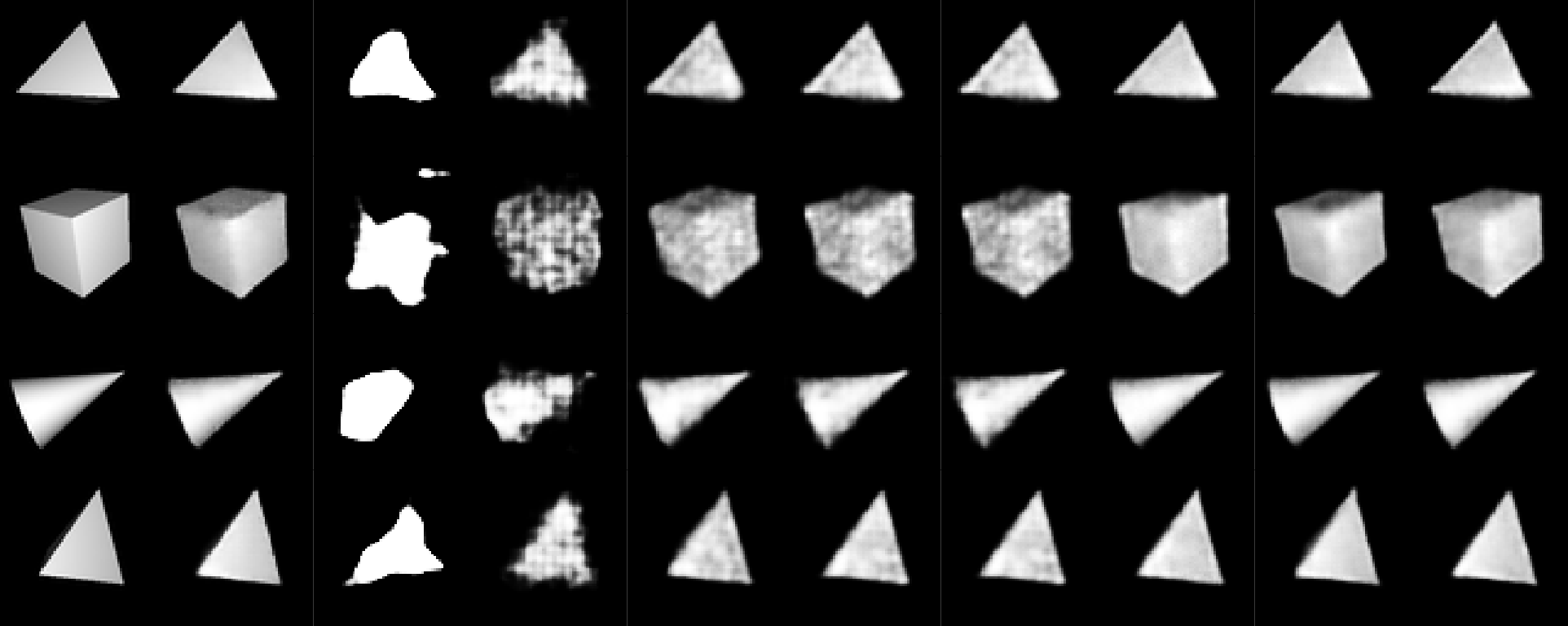}
    }%
    \caption{
        (a) and (b) show the reconstruction of images from $\mX_t = $ Symsol by a VAE trained on $\mX_s = $ dSprites.
        The first two columns on the left are the original images and reconstruction from a model trained on $\mX_t$,
        respectively. From left to right, columns 3 to 10 show the reconstruction as we progressively
        unfreeze and retrain the decoder layers, starting from layers closer to the input.
        The mean and variance layers of the encoder are fixed in (a) but retrained in (b).
    }
    \label{fig:gen-dec}
\end{figure}

\subsection{Implications for transfer learning}\label{subsec:cka-tl-impl}

\subsubsection*{Implications for unsupervised transfer learning}
As we have seen with CKA, decoders' representations are specific. Thus the decoder of a VAE trained on a source domain will
need to be retrained to generate images on the target domain. We show in~\Figref{fig:gen-dec} that the more layers of the
decoder are retrained, the better the reconstruction\footnote{Note that our training strategy does not impact our results as
this observation also holds when we unfreeze the outermost layers of the decoder first as reported in~\Appref{sec:app-fig-gen-dec}.}.
Thus, the specificity of the representations learned by the decoder holds for all the layers, not only a subset of them. Furthermore, the similar outputs obtained in~\Twofigref{fig:gen-dec-1}{fig:gen-dec-2}
show that retraining the latent representation has no impact on the reconstruction quality compared to the number of
retrained layers of the decoder. We can thus conclude that for image generation on the target domain, one can freeze the encoder entirely during transfer
but needs to retrain most of the layers of the decoder.

\subsubsection*{Implications for self-taught transfer learning}
Assessing the implications of our analysis of representational similarity for supervised target tasks is less
straightforward than for reconstruction. Indeed, while we can directly use the similarity between the representations
learned by a decoder trained on the source domain and its counterpart trained on the target domain in the context of
image generation, a similar approach would be unreliable for supervised target tasks.
Let us consider the mean representation of a model trained on the source domain, $\rvmu(\cdot; \vphi_s)$,
and the mean representation of a model trained on the target domain, $\rvmu(\cdot; \vphi_t)$.
The transferability of $\rvmu(\cdot; \vphi_s)$ to the target domain does not depend on how similar $\rvmu(\mX_t; \vphi_s)$ is to
$\rvmu(\mX_t; \vphi_t)$, but on how informative the shared latent representations are about the task labels $\rvy_t$.
For example, let us consider a simple classification task with only one binary label (i.e., $\rvy_t \in \{0,1\}$).
One could have a very high similarity between $\rvmu(\mX_t; \vphi_s)$ and $\rvmu(\mX_t; \vphi_t)$ because both representations share many
variables which are useful for reconstruction. However, $\rvmu(\mX_t; \vphi_s)$ may not encode the most informative
variable about $\rvy_t$, which would lead to poor results on the target task.
Inversely, one could have a very low similarity between $\rvmu(\mX_t; \vphi_s)$ and $\rvmu(\mX_t; \vphi_t)$
with $\rvmu(\mX_t; \vphi_s)$ encoding only the most informative variable about $\rvy_t$, which would lead to good results on the target
task. Despite this, one can still use knowledge from representational similarity to assess the transferability of latent representations
to a supervised target task, albeit in a different way.
We know that encoders' representations are generic. Thus, if a common variable is shared between the source and target
domains, it should be encoded in a similar way regardless of the domain of the input. For example, if colours are encoded on a VAE trained
on the source domain, they should also be encoded by $\rvmu(\mX_t; \vphi_s)$.
Furthermore, we know from representational similarity that the decoder learns specific representations. It means that given
any $\rvmu(\cdot; \vphi_s)$, it should provide a likely reconstruction in the target domain. Hence, akin to paired inputs
in multimodal datasets, the decoder should provide an approximation of any target example $\rvx_t^{(i)}$
in the source domain. One could thus identify the shared latent variables between the source and target domain by visually comparing
the target examples and their reconstruction in the source domain.

\begin{figure}[ht]
    \centering
    \includegraphics[width=.8\textwidth]{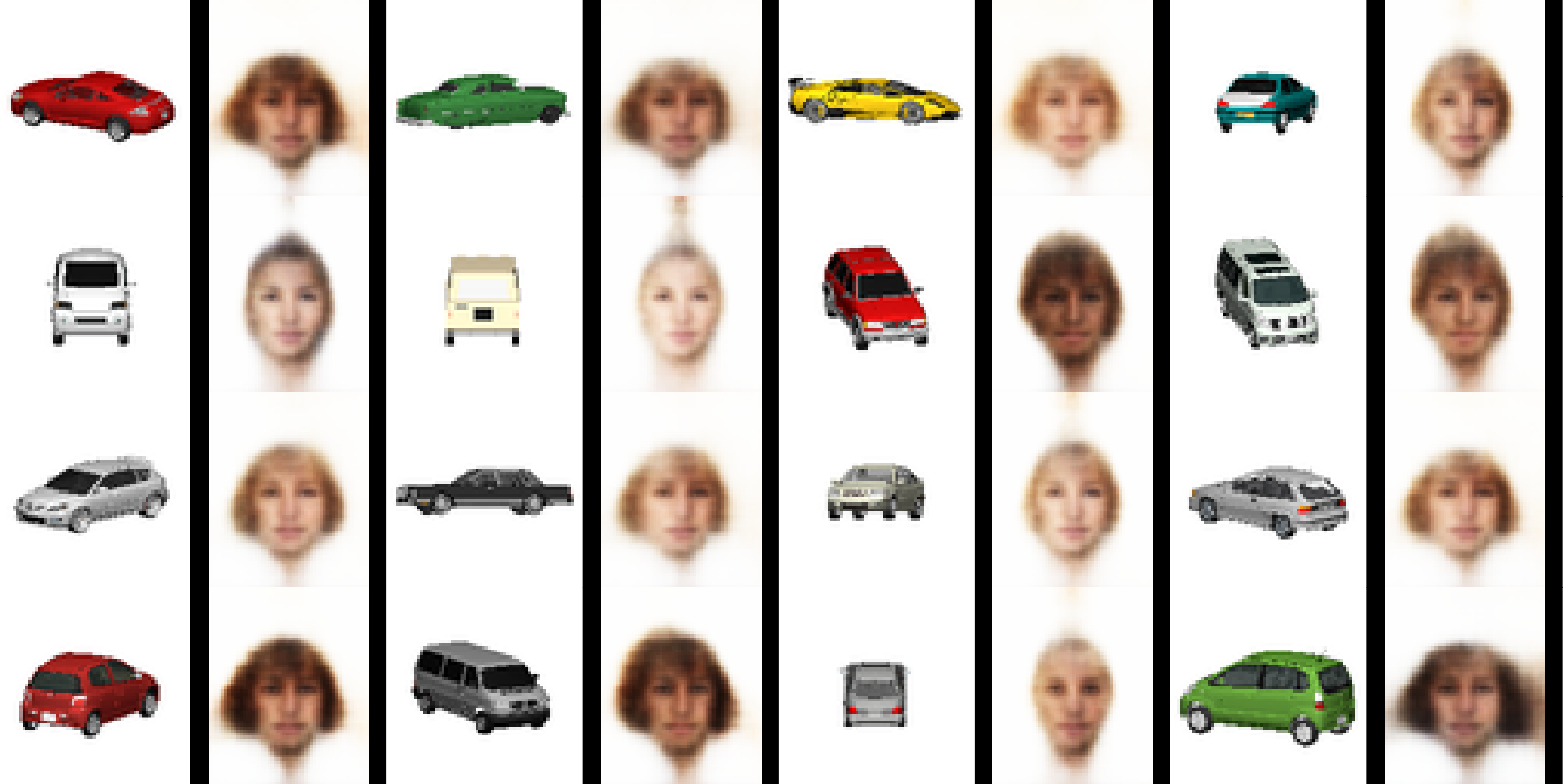}
    \caption{Reconstruction of Cars3D images using a VAE trained on Celeba. The black borders on the reconstructed images
    are due to the presence of similar borders in Celeba images.}
    \label{fig:celeba-to-cars}
\end{figure}

\subsubsection*{Case study 1: Celeba to cars3D} To validate our hypothesis, we will use~\Figref{fig:celeba-to-cars} to
compare the samples from Cars3D with their reconstruction using a VAE trained on Celeba. One can clearly see that
Celeba and Cars3d display a common variable \textbf{colour}: cars with light colours lead to faces with light hair and skin in the source domain while cars of
darker colour are linked with faces with darker hair and skin. In the same way, we can identify a shared variable \textbf{width}:
the profile views of cars lead to faces with wider jaw and haircut than front views of cars which results in faces with fine jaw and tight haircuts.
The labels of the target classification task are object type, elevation, and azimuth which are only loosely related to the shared variables identified.
We can thus hypothesise that despite the observed shared variables, the performances on the target classification task will likely drop when using the representations learned on Cars3D.

\begin{figure}[ht]
    \centering
    \includegraphics[width=.8\textwidth]{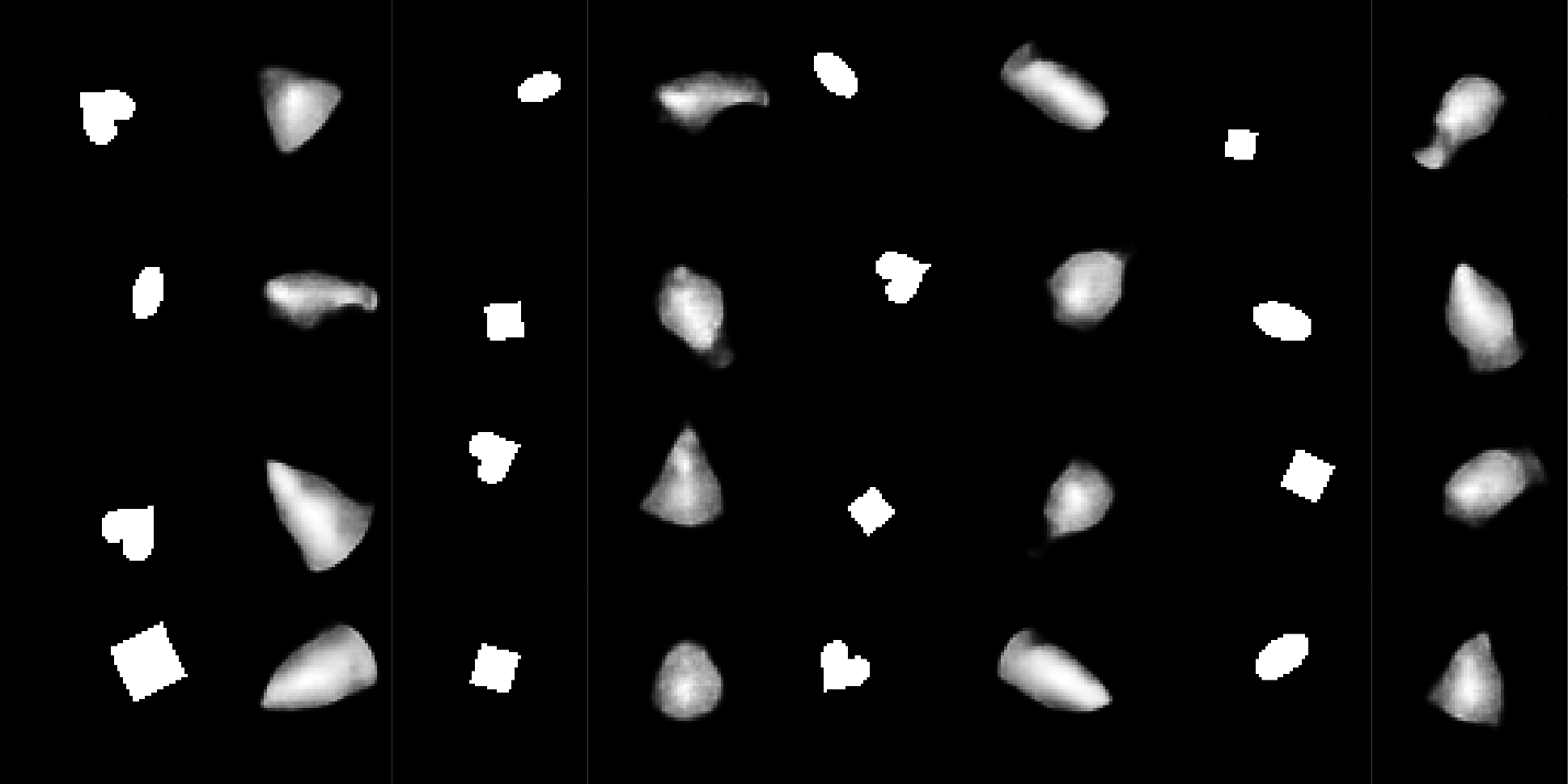}
    \caption{Reconstruction of dSprites images using a VAE trained on Symsol}
    \label{fig:symsol-to-dsprites}
\end{figure}

\subsubsection*{Case study 2: Symsol to dSprites} Taking a second example where we transfer representations from Symsol
to dSprites in~\Figref{fig:symsol-to-dsprites}, we can identify several shared variables: \textbf{position x and y}, as the
reconstructed output is displayed in the same area as the input for all the examples; \textbf{size}, albeit less consistently, as larger sprites
generally result in larger reconstructed outputs (e.g. see the two squares on the last row); \textbf{rotation}, to a lower extent, as the output tends to be slanted
when the input is rotated but the angle of rotation between input and reconstruction generally do not coincide. We can
also see that there is no shared variable for the shape. Indeed, different sprites have sometimes a more similar reconstruction
shape than the sprites of different types (e.g., see the heart and square in rows 3-4 and columns 1-2,
and the square in row 4, columns 3-4).
The labels of the target classification task are shape, scale, rotation, x and y position. We have seen that x and y positions
were clearly shared variables so we can expect good accuracy on the classification of these labels. Scale and rotation were
also identified but in a less consistent fashion. We can thus hypothesise a small drop in accuracy when using representations
learned on Symsol for these classifications. Finally, shape was definitely not a shared variable and we should expect a more
important drop in accuracy for the classification of this label.

\begin{table}
    \centering
    \caption{Averaged classification accuracy of a Gradient Boosted Tree on the target domain using
    the mean representations of VAEs trained on the source and target domain.
    All the results are averaged over 5 seeds.\\}\label{tab:cka-dt}

    \begin{tabular}{ |c c c| }
        \hline
        \textbf{Target domain} & \textbf{Source domain} & \textbf{Accuracy (avg.)}\\
        \hline\hline
        dSprites & dSprites & 0.40\\
        dSprites & Symsol & 0.39\\
        \hline
        Symsol & Symsol & 1.0\\
        Symsol & dSprites & 0.99\\
        \hline
        Celeba & Celeba & 0.83\\
        Celeba & Cars3D & 0.82\\
        \hline
        Cars3D & Cars3D & 0.59\\
        Cars3D & Celeba & 0.42\\
        \hline
    \end{tabular}
\end{table}

\subsubsection*{How well do these observations correlate with target task performances?}
Overall we can see in~\Tableref{tab:cka-dt} that the performances using representations learned on the source domains are on par with the results
obtained with representations learned from the target domain with the exception of $\mX_s=$ Celeba and $\mX_t=$ Cars3D.
To avoid clutter, the accuracy scores in~\Tableref{tab:cka-dt} are averaged over all the labels and the accuracy by label is detailed in~\Appref{app:clf-scores}.
Looking into the results obtained for individual labels of the classification on dSprites using representations learned on Symsol, the second
case study in this section, we observe the largest drop for shape classification (0.19), followed by smaller drops in rotation and scaling (around 0.05 for both),
which is consistent with our preliminary observations. Interestingly, this is compensated by a gain in accuracy of 0.15 for x and y positions which
were the most noticeable shared variables.
While not studied visually, classification on Celeba using representations learned on Cars3D shows similar results to those obtained with representations
learned on Celeba on most labels. The accuracy decreases mainly for attributes which intuitively cannot be inferred from car representations such as
smiling, wearing makeup, or being male.
The lower performances of $\mX_s=$ Celeba and $\mX_t=$ Cars3D are coherent with our analysis of the corresponding first case study in this section.
The shared variables were not very informative about the labels and the classification accuracy evenly dropped by 0.13 to 0.19
across all the labels.

    \section{Conclusion}\label{sec:conclusion}
After ensuring that CKA was consistent with known behaviours of VAEs in~\Secref{subsec:cka-check}, we used this metric
in~\Secref{subsec:cka-tl} to show that encoders' representations are generic but decoders' specific.
In~\Secref{subsec:cka-tl-impl} we further studied the implications of this analysis for transfer learning.

\subsubsection*{Impact on unsupervised transfer learning} When the target task is image generation, because the decoder
learns specific representations, one needs to retrain most of its layers to obtain a good reconstruction. However, as
the encoder learns generic representations, it does not need to be retrained.

\subsubsection*{Impact on self-taught transfer learning} Using the fact that encoders' representations are generic,
we hypothesised that when provided with an input from the target domain, a VAE trained on a source domain will retain
the shared variables between both domains in its latent representations. Furthermore, due to the specificity of
decoders' representations, any target example will be reconstructed in the source domain using these shared variables.
Thus, by visually comparing inputs from the target domain and outputs from a VAE trained on a different source domain,
one can identify the shared variables learned by this model. We have seen that such analysis is very informative on
the transferability of the latent representations to target classification tasks. Indeed, clearly identified shared variables,
when related to the classification labels, lead to equivalent or better results when using latents obtained from an
encoder learned on a different source domain. This finding can have implications in practical applications where one
can only obtain a few samples from the target domain but a large dataset is available from a source domain.
\clearpage

\subsubsection*{Ethical statement}
While the methods proposed in this paper open avenues for greener transfer learning of VAEs,
this experiment still required extensive computations.
We trained more than 300 VAEs using 4 learning objectives, 5 different initialisations,
5 regularisation strengths, and 3 datasets, which took around 6,000 hours on an NVIDIA A100 GPU.
We then computed the CKA scores for the 15 layer activations (plus the input) of each model combinations
considered above at 5 different epochs, resulting in 470 million similarity scores and approximately 7,000 hours of computation
on an Intel Xeon Gold 6136 CPU.~As Procrustes is slowed down by the computation of the nuclear norm for high dimensional activations, the same number of similarity
scores would have been prohibitively long to compute, requiring 30,000 hours on an NVIDIA A100 GPU.~We thus only computed the Procrustes
similarity for one dataset, reducing the computation time to 10,000 hours.
Overall, based on the estimations of~\cite{Lacoste2019}, the computations done for this experiment amount to
2,200 Kg of \coo, which corresponds to the \coo~produced by one person over 5 months.~To mitigate the negative environmental impact of our work,
we released \ifanonymous\else all our trained models and metric scores at \url{https://data.kent.ac.uk/428/}, and \url{https://data.kent.ac.uk/444/}, respectively\fi.
We hope that this will help to prevent unnecessary recomputation should others wish to reuse our results.
Moreover, we believe that our findings could help practitioners to pre-select likely candidate models for transfer learning
and avoid unnecessary retraining, reducing \coo~emissions in the future.
    \ifanonymous
    \else
        \subsubsection*{Acknowledgments}
The authors thank Frances Ding for an insightful discussion on the Procrustes distance, as well as Th\'{e}ophile Champion
and Declan Collins for their helpful comments on the paper.
    \fi
    \clearpage
    \bibliographystyle{arxiv}
    \bibliography{main}


    \clearpage
    \appendix
    \section{Accuracy of classification tasks detailed per label}\label{app:clf-scores}

This section details the accuracy per label of the classification tasks presented in~\Tableref{tab:cka-dt} of \Secref{subsec:cka-tl-impl}.
Note that Symsol is omitted because the classification task has only one label.

\begin{table}
    \centering
    \caption{Averaged classification accuracy of a Gradient Boosted Tree on the Cars3D using
    the mean representations of VAEs trained on different source domains.
    All the results are averaged over 5 seeds.\\}\label{tab:dt-cars}

    \begin{tabular}{ |c c c c| }
        \hline
        \textbf{Source domain} & \textbf{Elevation} & \textbf{Azimuth} & \textbf{Object type}\\
        \hline\hline
        Cars3d & 0.69 & 0.73 & 0.35\\
        Celeba & 0.50 & 0.60 & 0.16\\
        \hline
    \end{tabular}
\end{table}

\begin{table}
    \centering
    \caption{Averaged classification accuracy of a Gradient Boosted Tree on the dSprites using
    the mean representations of VAEs trained on different source domains.
    All the results are averaged over 5 seeds.\\}\label{tab:dt-dsprites}

    \begin{tabular}{ |c c c c c c| }
        \hline
        \textbf{Source domain} & \textbf{Shape} & \textbf{Scale} & \textbf{Orientation} & \textbf{x position} & \textbf{y position} \\
        \hline\hline
        dSprites & 0.72 & 0.53 & 0.11 & 0.31 & 0.31\\
        Symsol & 0.52 & 0.48 & 0.05 & 0.46 & 0.45\\
        \hline
    \end{tabular}
\end{table}

\begin{table}
    \centering
    \caption{Averaged classification accuracy of a Gradient Boosted Tree on the Celeba using
    the mean representations of VAEs trained on different source domains. As CelebA has 40 labels,
    only those with a difference in accuracy higher than $0.01$ between source domains are reported for readability.
    All the results are averaged over 5 seeds.\\}\label{tab:dt-celeba}

    \begin{tabular}{ |c c c| }
        \hline
        \textbf{Label} & \textbf{Source domain} & \textbf{Accuracy}\\
        \hline\hline
        \textbf{Attractive} & Celeba & 0.68\\
        \textbf{Attractive} & Cars3D & 0.64\\
        \hline
        \textbf{Blond Hair} & Celeba & 0.89\\
        \textbf{Blond Hair} & Cars3D & 0.87\\
        \hline
        \textbf{Heavy Makeup} & Celeba & 0.74\\
        \textbf{Heavy Makeup} & Cars3D & 0.68\\
        \hline
        \textbf{High Cheekbones} & Celeba & 0.62\\
        \textbf{High Cheekbones} & Cars3D & 0.59\\
        \hline
        \textbf{Male} & Celeba & 0.75\\
        \textbf{Male} & Cars3D & 0.67\\
        \hline
        \textbf{Smiling} & Celeba & 0.63\\
        \textbf{Smiling} & Cars3D & 0.60\\
        \hline
        \textbf{Wavy Hair} & Celeba & 0.72\\
        \textbf{Wavy Hair} & Cars3D & 0.70\\
        \hline
        \textbf{Wearing Lipstick} & Celeba & 0.75\\
        \textbf{Wearing Lipstick} & Cars3D & 0.67\\
        \hline
    \end{tabular}
\end{table}
    \clearpage
    \section{Additional figures obtained when retraining the outermost layers of the decoder first}\label{sec:app-fig-gen-dec}
\begin{figure}[ht]
    \centering
    \subcaptionbox{Fixed encoder\label{fig:app-gen-dec-1}}{
        \includegraphics[width=0.48\textwidth]{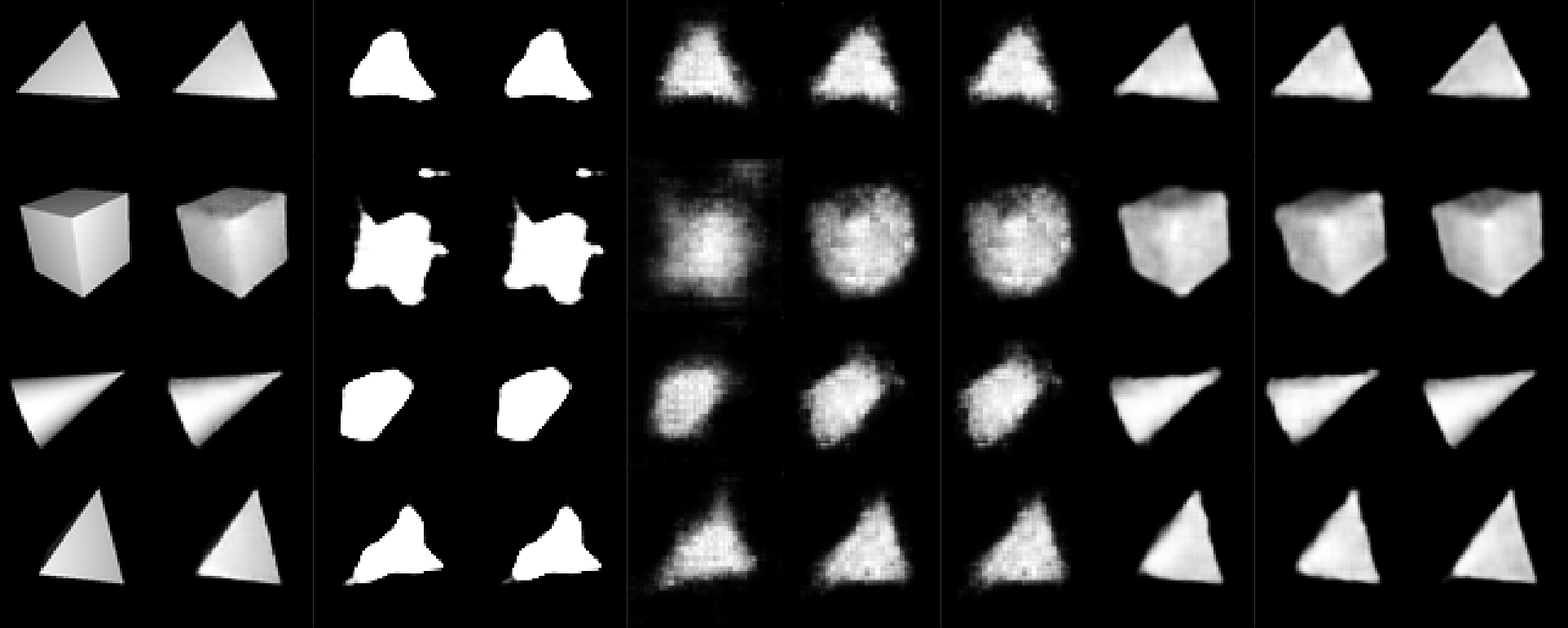}
    }%
    \hfill
    \subcaptionbox{Retrained mean and variance\label{fig:app-gen-dec-2}}{
        \includegraphics[width=0.48\textwidth]{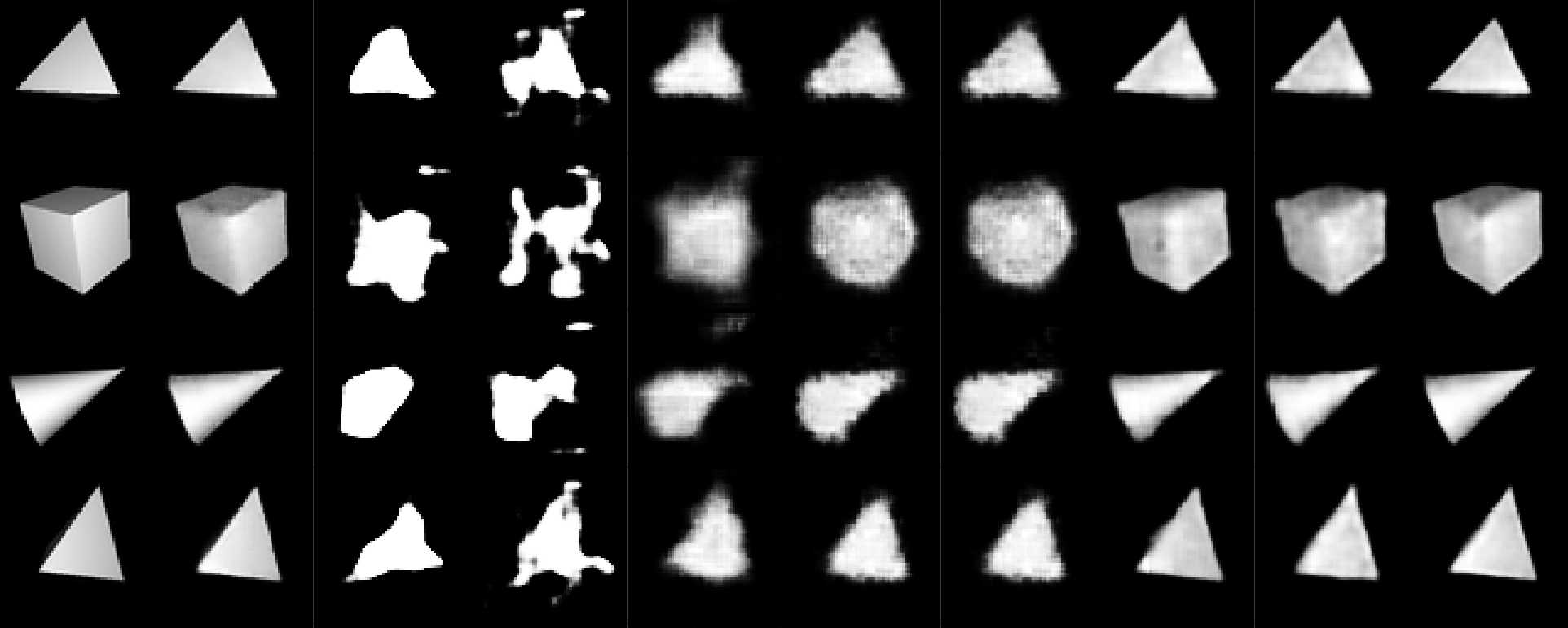}
    }%
    \caption{
        (a) and (b) show the reconstruction of Symsol images by a VAE trained on dSprites.
        The first two columns on the left are the original images and reconstruction from a model trained on the target
        dataset, respectively. From left to right, columns 3 to 10 show the reconstruction as we progressively
        unfreeze and retrain the decoder layers, starting from layers closer to the output.
        The mean and variance layers of the encoder are fixed in (a) but retrained in (b).
    }
    \label{fig:app-gen-dec}
\end{figure}

This section provides complementary observations to~\Figref{fig:gen-dec} when we unfreeze the outermost layers of the decoder first.
As in~\Figref{fig:gen-dec} the more layers of the decoder are retrained, the better the reconstruction.
The similar outputs obtained in~\Twofigref{fig:app-gen-dec-1}{fig:app-gen-dec-2} show that retraining the latent
representation has no impact on the reconstruction quality compared to the number of retrained layers of the decoder.

    \section{Experimental setup}\label{sec:xp-setup}
To facilitate the reproducibility of our experiment, we detail below the Procrustes normalisation process and the configuration used for model training.
\subsubsection*{Procrustes normalisation} Similarly to~\cite{Ding2021}, given an activation matrix $\mX \in \R^{n \times m}$
containing $n$ samples and $m$ features, we compute the vector $\bar{\vx} \in \R^{m}$ containing the mean values of the columns of $\mX$.
Using the outer product $\otimes$, we get $\bar{\mX} = \vone_n \otimes \bar{\vx}$, where $\vone_n \in \R^{n}$ is a vector of ones
and $\bar{\mX} \in \R^{n \times m}$. We then normalise $\mX$ such that
\begin{equation}
        \dot{\mX} = \frac{\mX - \bar{\mX}}{\norm{\mX - \bar{\mX}}_F}.
\end{equation}
As the Frobenius norm of $\dot{\mX}$ and $\dot{\mY}$ is 1, and $\norm{\dot{\mY}^T\dot{\mX}}_*$ is always positive (1 when
$\dot{\mX} = \dot{\mY}$, smaller otherwise), \Eqref{eq:pd} lies in $[0,2]$, and \Eqref{eq:ps} in $[0, 1]$.

\subsubsection*{VAE training} Our implementation uses the same hyperparameters as~\cite{Locatello2019a}, and the details are listed in~\Twotablerefs{table:global-hyperparam}{table:model-hyperparam}.
We reimplemented~\cite{Locatello2019a} code base, designed for Tensorflow 1, in Tensorflow 2 using Keras.
The model architecture used is also identical, as described in~\Tableref{table:architecture}.
Each model is trained 5 times, on seeded runs with seed values from 0 to 4.
Intermediate models are saved every 1,000 steps for SmallNorb, 6,000 steps for Cars3D and 11,520 steps for dSprites.
Every image input is normalised to have pixel values between 0 and 1.

For the fully-connected models presented in~\Appref{sec:app-fc}, we used the same architecture and hyperparameters as those implemented in \texttt{disentanglement lib} of~\cite{Locatello2019a}, and the details are presented in~\Twotablerefs{table:linear-architecture}{table:linear-model-hyperparam}.

\begin{table}[h!]
    \centering
    \caption{Shared hyperparameters}
    \label{table:global-hyperparam}
    \begin{tabular}{ l l }
        \hline
        Parameter & Value \\
        \hline
        Batch size & 64  \\
        Latent space dimension & 10  \\
        Optimizer & Adam \\
        Adam: $\beta_1$ & 0.9 \\
        Adam: $\beta_2$ & 0.999 \\
        Adam: $\epsilon$ & 1e-8 \\
        Adam: learning rate & 0.0001 \\
        Reconstruction loss & Bernoulli \\
        Training steps & 300,000 \\
        Intermediate model saving & every 6K steps\\
        Train/test split & 90/10\\
        \hline
    \end{tabular}
\end{table}

\begin{table}[h!]
    \centering
    \caption{Model-specific hyperparameters}
    \label{table:model-hyperparam}
    \begin{tabular}{ l l l }
        \hline
        Model & Parameter & Value \\
        \hline
        $\beta$-VAE & $\beta$ & [1, 2, 4, 6, 8] \\
        $\beta$-TC VAE & $\beta$ & [1, 2, 4, 6, 8] \\
        DIP-VAE II & $\lambda_{od}$ & [1, 2, 5, 10, 20] \\
        & $\lambda_{d}$ & $\lambda_{od}$ \\
        Annealed VAE & $C_{max}$ & [5, 10, 25, 50, 75] \\
        & $\gamma$ & 1,000 \\
        & iteration threshold & 100,000 \\
        \hline
    \end{tabular}
\end{table}

\begin{table}[h!]
    \centering
    \caption{Shared architecture}
    \label{table:architecture}
    \begin{tabularx}{\linewidth}{ X X }
        \hline
        Encoder & Decoder \\
        \hline
        Input: $\R^{64 \times 63 \times channels}$ & $\R^{10}$ \\
        Conv, kernel=4×4, filters=32, activation=ReLU, strides=2 & FC, output shape=256, activation=ReLU \\
        Conv, kernel=4×4, filters=32, activation=ReLU, strides=2 & FC, output shape=4x4x64, activation=ReLU \\
        Conv, kernel=4×4, filters=64, activation=ReLU, strides=2 & Deconv, kernel=4×4, filters=64, activation=ReLU, strides=2 \\
        Conv, kernel=4×4, filters=64, activation=ReLU, strides=2 & Deconv, kernel=4×4, filters=32, activation=ReLU, strides=2 \\
        FC, output shape=256, activation=ReLU, strides=2 & Deconv, kernel=4×4, filters=32, activation=ReLU, strides=2 \\
        FC, output shape=2x10 & Deconv, kernel=4×4, filters=channels, activation=ReLU, strides=2 \\
        \hline
    \end{tabularx}
\end{table}

\begin{table}[h!]
    \centering
    \caption{Fully-connected architecture}
    \label{table:linear-architecture}
    \begin{tabularx}{\linewidth}{ X X }
        \hline
        Encoder & Decoder \\
        \hline
        Input: $\R^{64 \times 63 \times channels}$ & $\R^{10}$ \\
        FC, output shape=1200, activation=ReLU & FC, output shape=256, activation=tanh \\
        FC, output shape=1200, activation=ReLU & FC, output shape=1200, activation=tanh \\
        FC, output shape=2x10 & FC, output shape=1200, activation=tanh \\
        \hline
    \end{tabularx}
\end{table}

\begin{table}[h!]
    \centering
    \caption{Hyperparameters of fully-connected models}
    \label{table:linear-model-hyperparam}
    \begin{tabular}{ l l l }
        \hline
        Model & Parameter & Value \\
        \hline
        $\beta$-VAE & $\beta$ & [1, 8, 16] \\
        $\beta$-TC VAE & $\beta$ & [2] \\
        DIP-VAE II & $\lambda_{od}$ & [1, 20, 50] \\
        & $\lambda_{d}$ & $\lambda_{od}$ \\
        Annealed VAE & $C_{max}$ & [5] \\
        & $\gamma$ & 1,000 \\
        & iteration threshold & 100,000 \\
        \hline
    \end{tabular}
\end{table}
    \section{Resources}\label{sec:app-ressources}
As mentioned in~\Twosecrefs{sec:intro}{sec:experiment}, we released the code of our experiment, the pre-trained models
and similarity scores of~\Secref{subsec:cka-check}:
\ifanonymous
\else
\begin{itemize}
    \item The similarity scores can be downloaded at~\url{https://data.kent.ac.uk/444/}
    \item The pre-trained models can be downloaded at~\url{https://data.kent.ac.uk/428/}
    \item The code is available at~\url{https://github.com/bonheml/VAE_learning_dynamics}
\end{itemize}
\fi

\noindent Note that the 300 VAE models released correspond to models trained with:
\begin{itemize}
    \item 4 different learning objectives,
    \item 5 initialisations,
    \item 3 datasets,
    \item 5 regularisation strengths.
\end{itemize}
    \section{Disentangled representation learning}\label{sec:app-disentanglement}
As mentioned in~\Secref{sec:background}, we are interested in the family of methods modifying the weight on the
regularisation term of~\Eqref{eq:elbo} to encourage disentanglement. In our paper, the term regularisation refers to the moderation of this parameter only.~To achieve this, our experiment will focus on the models described below.

\subsubsection*{$\beta$-VAE} The goal of this method~\citep{Higgins2017} is to penalise the regularisation
term of~\Eqref{eq:elbo} by a factor $\beta > 1$, such that
\begin{equation}
    \label{eq:Higgins2017}
    \ELBO = \E_{q_{\vphi}(\rvz|\rvx)}\left[\log p_{\vtheta}(\rvx|\rvz)\right]-\beta\KL{q_{\vphi}(\rvz|\rvx)}{p(\rvz)}.
\end{equation}

\subsubsection*{Annealed VAE}~\cite{Burgess2018} proposed to gradually increase
the encoding capacity of the network during the training process. The goal is to progressively learn latent variables by
decreasing order of importance.
This leads to the following objective, where \textrm{C} is a parameter that can be understood as a channel capacity and $\gamma$ is a hyper-parameter
penalising the divergence, similarly to $\beta$ in $\beta$-VAE:
\begin{equation}
    \ELBO = \E_{q_{\vphi}(\rvz|\rvx)}\left[\log p_{\vtheta}(\rvx|\rvz)\right]-\gamma\left|\KL{q_{\vphi}(\rvz|\rvx)}{p(\rvz)}-\mathrm{C}\right|\label{eq:Burgess2018}.
\end{equation}
As the training progresses, the channel capacity $\mathrm{C}$ is increased, going from zero to its maximum channel capacity $\mathrm{C_{max}}$ and allowing a higher value of the KL divergence term.
VAEs that use~\Eqref{eq:Burgess2018} as a learning objective are referred to as Annealed VAEs in this paper.

\subsubsection*{$\beta$-TC VAE}~\cite{Chen2018} argued that only the distance between the
estimated latent factors and the prior should be penalised to encourage disentanglement, such that
\begin{multline}
    \label{eq:kl-pz-qz}
    \ELBO = \E_{p(\rvx)}\left[\E_{q_{\vphi}(\rvz|\rvx)}\left[\log p_{\vtheta}(\rvx|\rvz)\right]-\KL{q_{\vphi}(\rvz|\rvx)}{p(\rvz)}\right]\\
    -\lambda\KL{q_{\vphi}(\rvz)}{p(\rvz)}.
\end{multline}
Here, $\KL{q_{\vphi}(\rvz)}{p(\rvz)}$ is approximated by penalising the dependencies between the dimensions of $q_{\vphi}(\rvz)$:
\begin{multline}
    \ELBO \approx \frac{1}{n}\sum_{i=1}^{n}\left[\E_{q_{\vphi}(\rvz|\rvx^{(i)})}\left[\log p_{\vtheta}(\rvx^{(i)}|\rvz)\right]-\KL{q_{\vphi}(\rvz|\rvx^{(i)})}{p(\rvz)}\right]\\
    -\underbrace{\lambda\KL{q_{\vphi}(\rvz)}{\prod_{j=1}^{\mathrm{D}}q_{\vphi}(\rvz_{j})}}_{\textrm{total correlation}}\label{eq:Kim2018}.
\end{multline}
The total correlation of \Eqref{eq:Kim2018} is then approximated over a mini-batch of samples $\{\rvx^{(i)}\}_{i=1}^{m}$ as follows:
\begin{equation}\label{eq:beta-tc-vae}
\E_{q_{\vphi}(\rvz)}[\log{q_{\vphi}(\rvz)}]\approx\frac{1}{m}\sum_{i=1}^{m}\left(\log \frac{1}{nm}\sum_{k=1}^{m}q_{\vphi}(\rvz^{(i)}|\rvx^{(k)})\right),
\end{equation}
where $m$ is the number of samples in the mini-batch, and $n$ total number of input examples.
$\E_{q_{\vphi}(\rvz_j)}[\log{q_{\vphi}(\rvz_j)}]$ can be computed in a similar way.
We refer the reader to~\cite[Appendix C.1]{Chen2018} for the detailed derivation of~\Eqref{eq:beta-tc-vae}.

\subsubsection*{DIP-VAE} Similarly to~\cite{Chen2018},~\cite{Kumar2018} proposed to regularise the
distance between $q_\vphi(\rvz)$ and $p(\rvz)$ using~\Eqref{eq:kl-pz-qz}.
The main difference is that here $\KL{q_{\vphi}(\rvz)}{p(\rvz)}$ is measured by matching the moments of the learned
distribution $q_{\vphi}(\rvz)$ and its prior $p(\rvz)$.
The second moment of the learned distribution is given by
\begin{equation}
    \label{eq:Kumar2018Cov}
    \mathrm{Cov}_{q_{\vphi}(\rvz)}[\rvz] = \mathrm{Cov}_{p(\rvx)}\left[\mu_{\vphi}(\rvx)\right] + \E_{p(\rvx)}\left[\Sigma_{\vphi}(\rvx)\right].
\end{equation}
DIP-VAE II penalises both terms of~\Eqref{eq:Kumar2018Cov} such that
\begin{equation*}
    \lambda\KL{q_{\vphi}(\rvz)}{p(\rvz)} = \lambda_{od} \sum_{i\neq j}\left(\mathrm{Cov}_{q_{\vphi}(\rvz)}\left[\rvz\right]\right)_{ij}^{2}
    + \lambda_{d} \sum_{i} \left(\mathrm{Cov}_{q_{\vphi}(\rvz)}\left[\rvz\right]_{ii} -1\right)^{2},
\end{equation*}
where $\lambda_{d}$ and $\lambda_{od}$ are the penalisation terms for the diagonal and off-diagonal values respectively.
    \section{Consistency of the results with Procrustes Similarity}\label{sec:procrustes}
\begin{figure}[ht!]
    \centering
    \subcaptionbox{CKA\label{fig:heatmap-cka}}{
        \includegraphics[width=0.5\textwidth]{heatmaps/cars3d/beta_tc_vae_2_epoch_25_beta_tc_vae_2_epoch_1090}
    }%
    \subcaptionbox{Procrustes\label{fig:heatmap-procrustes}}{
        \includegraphics[width=0.5\textwidth]{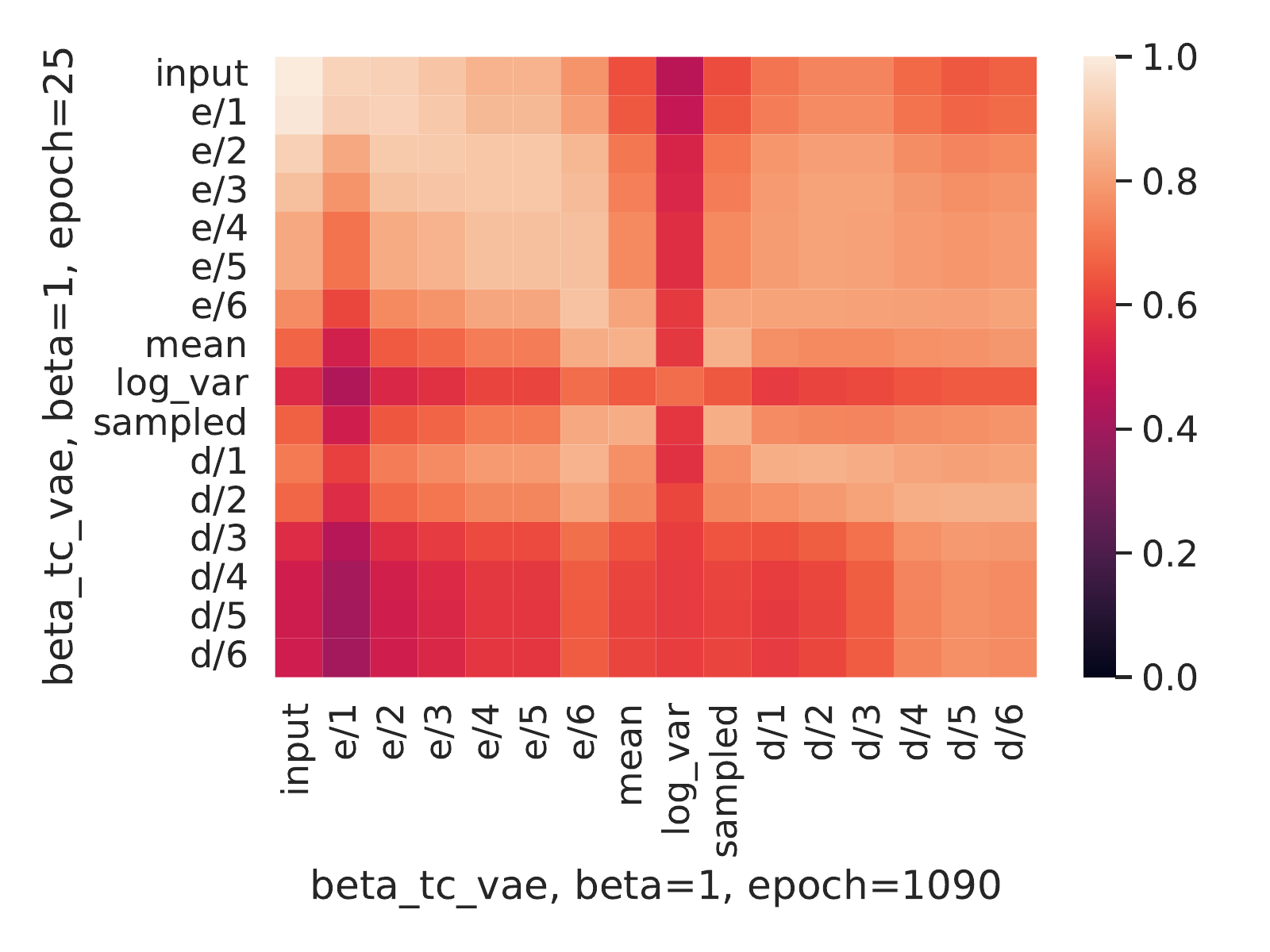}
    }%
    \caption{(a) shows the CKA similarity scores of activations
        at epochs 25 and 1090 of $\beta$-TC VAE trained on Cars3D with $\beta=2$.
        (b) shows the Procrustes similarity scores of the same configuration.~We observe the same trend with both metrics
        with Procrustes slightly overestimating the similarity between high dimensional activations (bottom-right quadrants), which agrees with the properties of the Procrustes similarity reported in \Secref{subsec:bg-limitations}.
        }
    \label{fig:training-sim-comp}
\end{figure}

\begin{figure}[ht!]
    \centering
    \subcaptionbox{CKA\label{fig:methods-cka}}{
        \includegraphics[width=0.5\textwidth]{heatmaps/cars3d/beta_vae_1_epoch_1090_dip_vae_ii_1_epoch_1090}
    }%
     \subcaptionbox{Procrustes\label{fig:methods-procrustes}}{
        \includegraphics[width=0.5\textwidth]{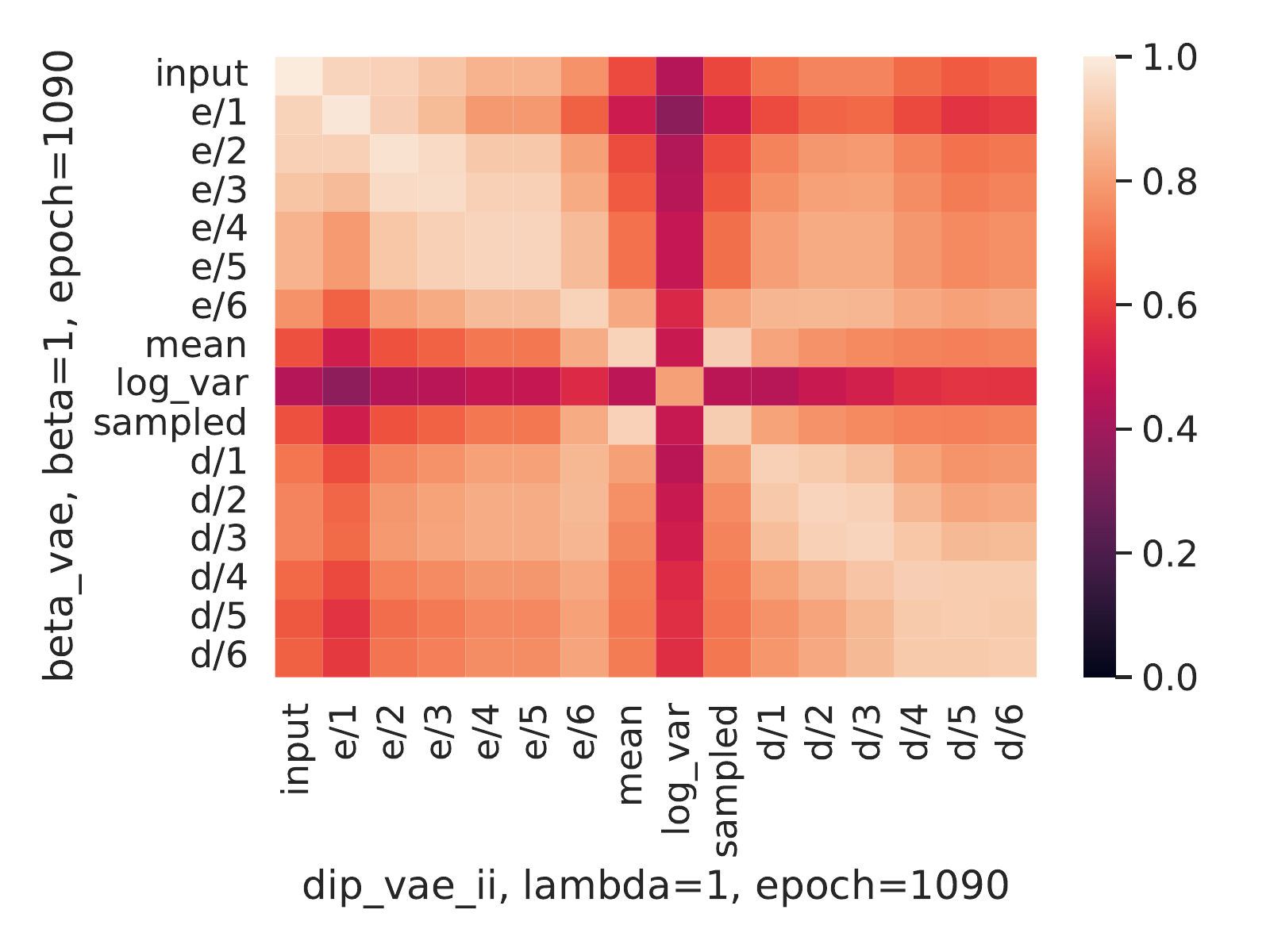}
    }%
    \caption{(a) shows the CKA similarity scores of activations of $\beta$-VAE and DIP-VAE II trained on Cars3D with $\beta=1$, and $\lambda=1$, respectively.
        (b) shows the Procrustes similarity scores using the same configuration.~We observe the same trend with both metrics
        with Procrustes slightly overestimating the similarity between high dimensional activations (bottom-right quadrants) (c.f.~\Secref{subsec:bg-limitations}).
    }
    \label{fig:methods-comp}
\end{figure}

\begin{figure}[ht!]
    \centering
    \subcaptionbox{CKA for $\beta=1$\label{fig:enc-1-cka}}{
        \includegraphics[width=0.5\textwidth]{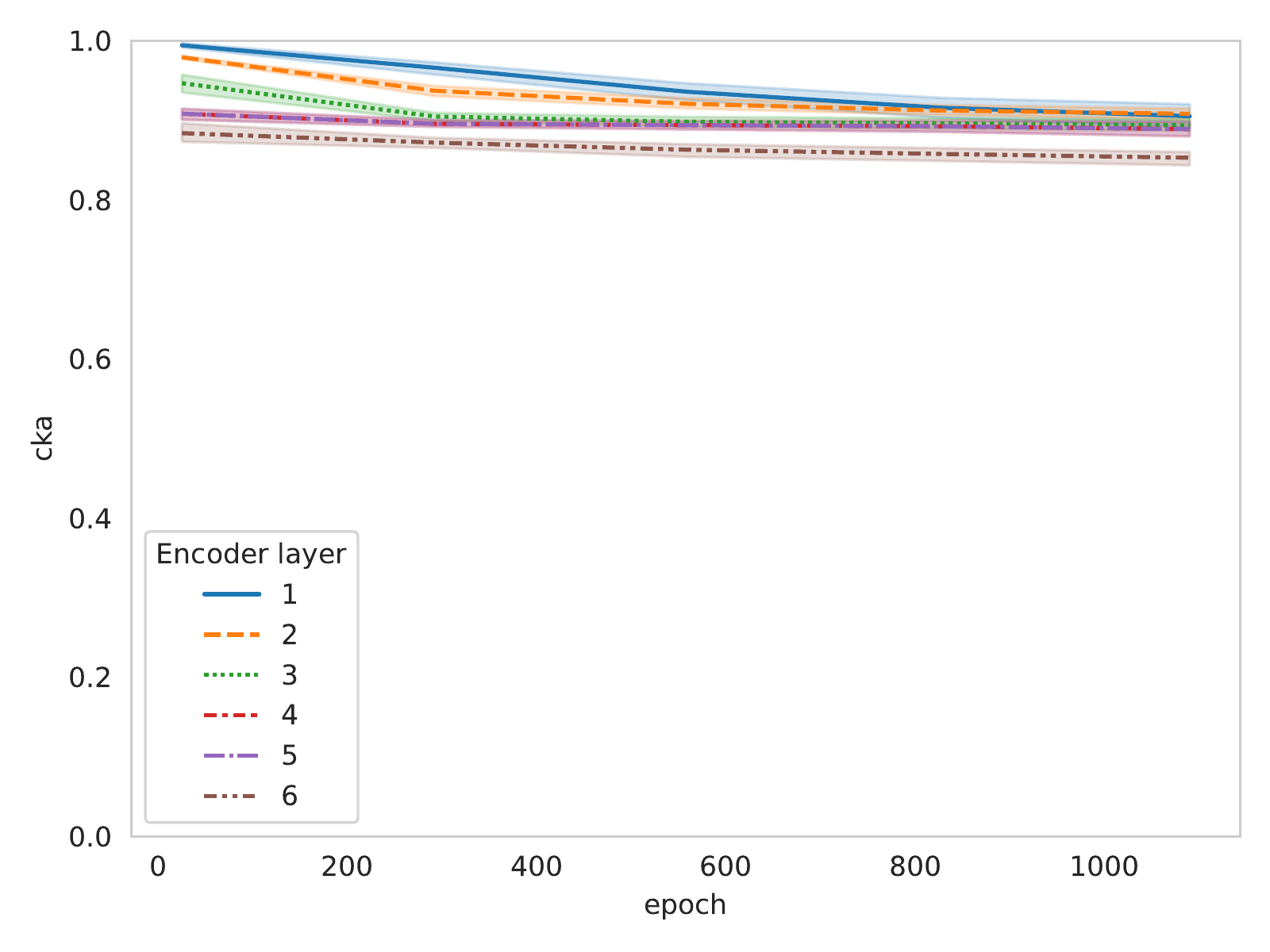}
    }%
    \subcaptionbox{CKA for $\beta=8$\label{fig:enc-2-cka}}{
        \includegraphics[width=0.5\textwidth]{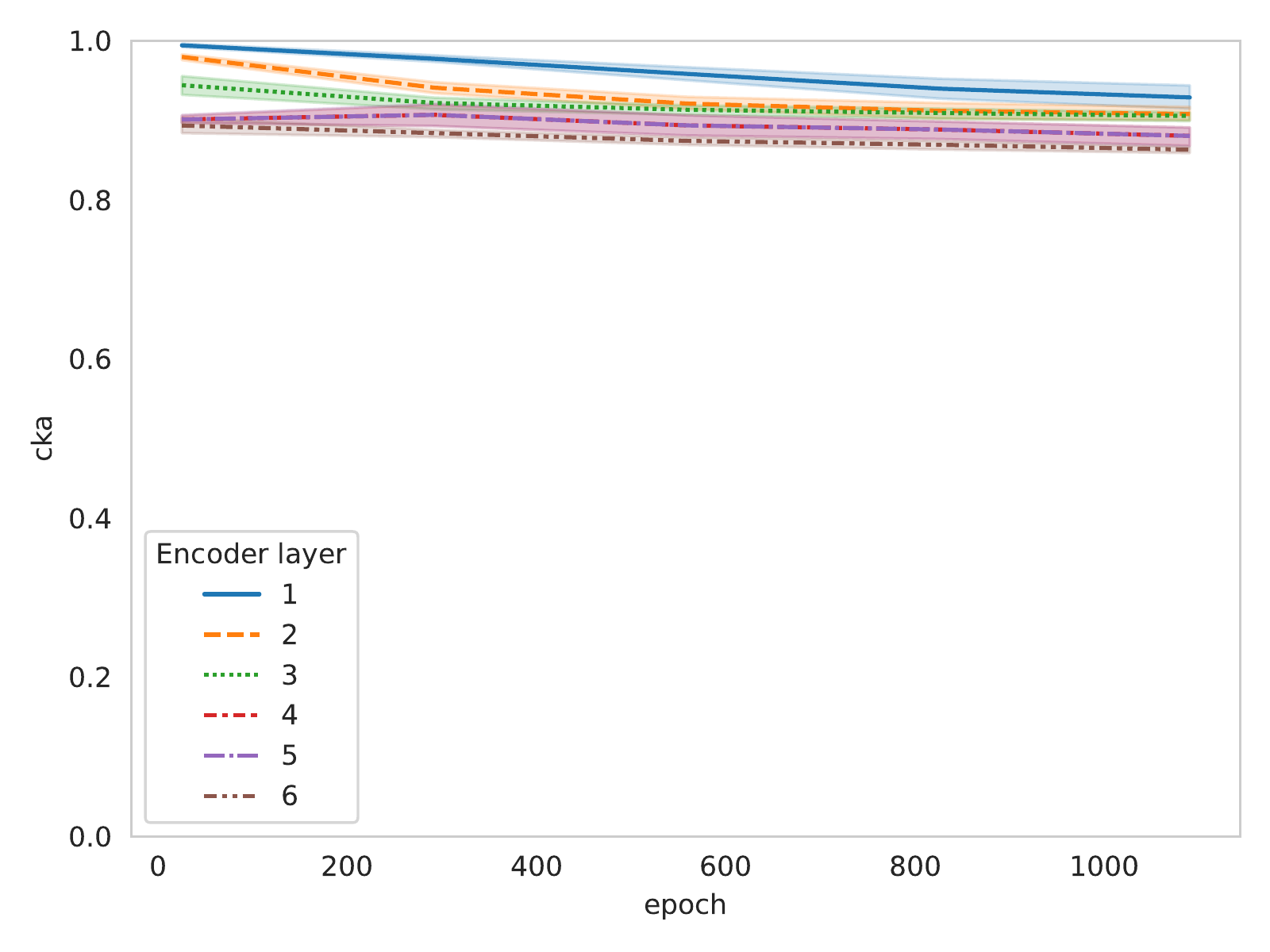}
    }%
    \\
    \subcaptionbox{Procrustes for $\beta=1$\label{fig:enc-1-procrustes}}{
        \includegraphics[width=0.5\textwidth]{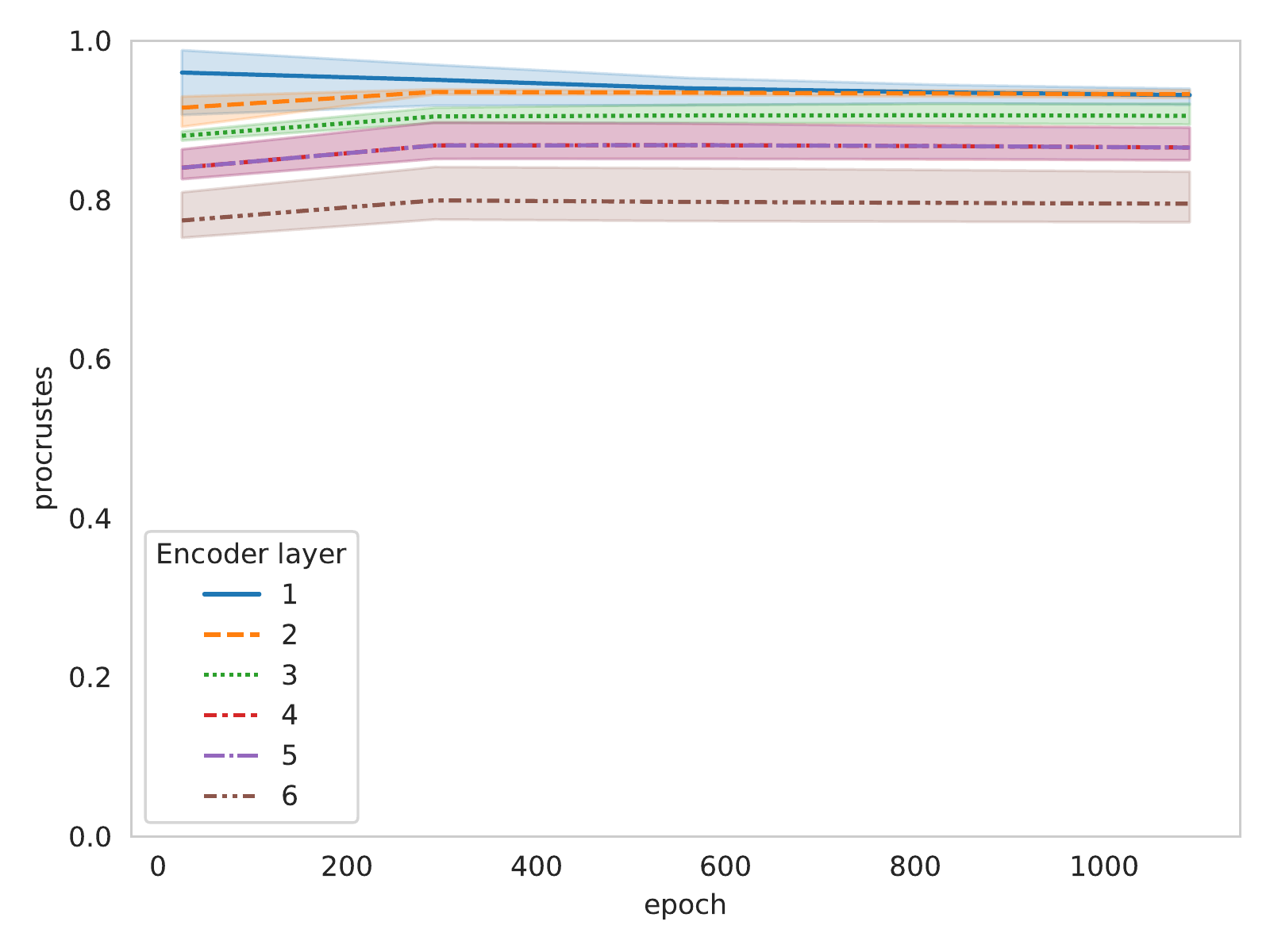}
    }%
    \subcaptionbox{Procrustes for $\beta=8$\label{fig:enc-2-procrustes}}{
        \includegraphics[width=0.5\textwidth]{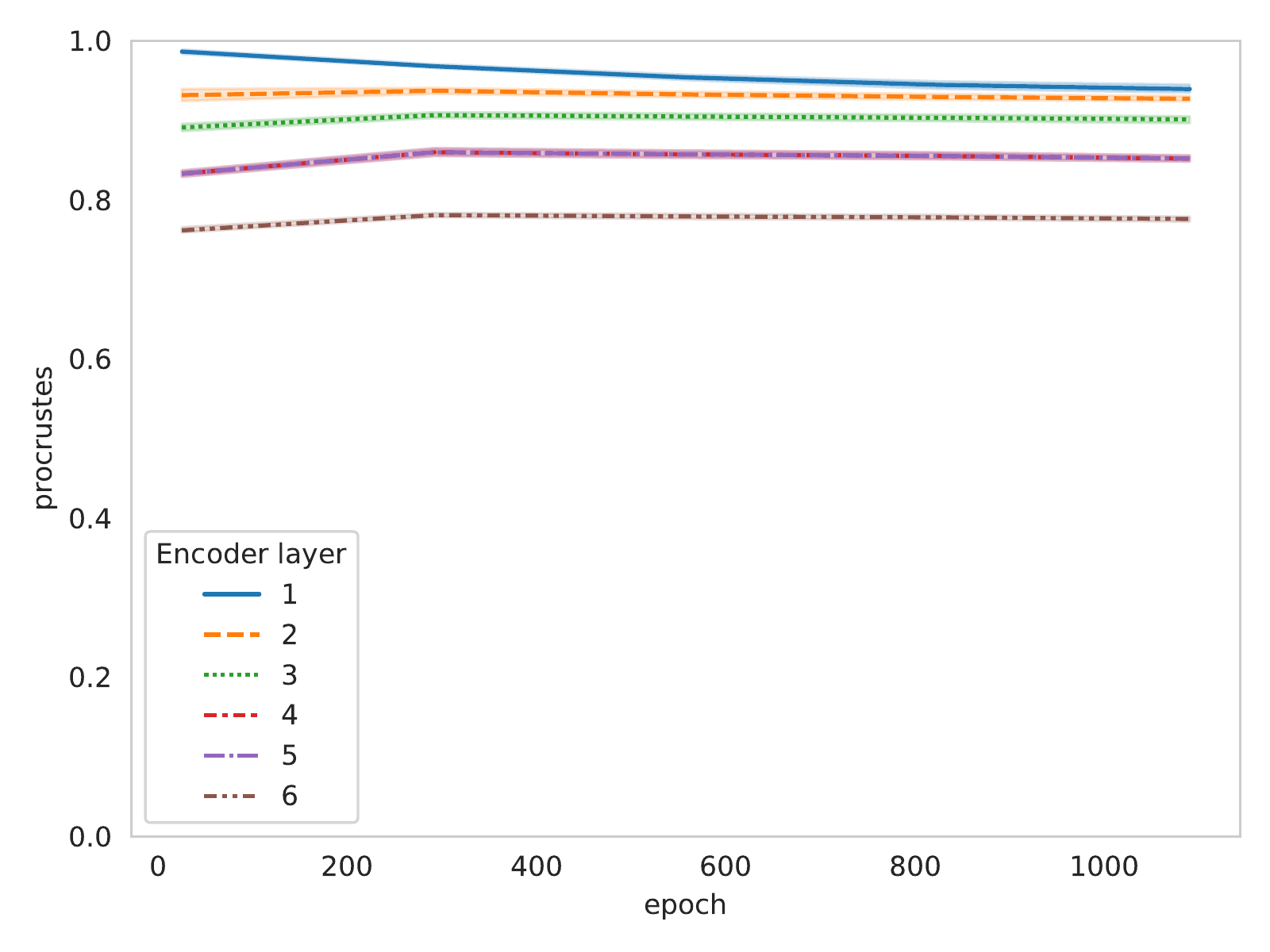}
    }%
     \caption{(a) shows the CKA scores between the inputs and the activations of the first 6 layers of the encoder of a
        $\beta$-VAE trained on Cars3D with $\beta=1$.~(b) shows the scores between the same representations with $\beta=8$.
         (c) and (d) are the Procrustes scores of the same configurations.~We observe the same trend for both metrics
        with more variance in (c) for Procrustes with $\beta=1$.~Procrustes also displays a slightly lower similarity for layers 4 to 6 of the encoder,
        possibly due to changes in the representation that are underestimated by CKA (c.f.~\Secref{subsec:bg-limitations}).}
    \label{fig:reg-encoder-comp}
\end{figure}

As mentioned in~\Secref{sec:experiment}, in this section we provide a comparison between the CKA scores reported in the
main paper, and the Procrustes scores for the Cars3D dataset.
We can see in \Figrangeref{fig:training-sim-comp}{fig:reg-encoder-comp} that Procrustes and CKA provide similar results.
~\Figrangetworef{fig:training-sim-comp}{fig:methods-comp} show that Procrustes tends to overestimate the similarity between
high-dimensional inputs, as mentioned in~\Secref{subsec:bg-limitations} (recall the example given in~\Figref{fig:cka-procrustes}).
In~\Figref{fig:reg-encoder-comp}, we observe a slightly lower similarity with Procrustes than CKA on the $5^{th}$ and $6^{th}$ layers
of the encoder, indicating that some small changes in the representations may have been underestimated by CKA,
as discussed in~\Secref{subsec:bg-limitations} and by~\cite{Ding2021}.~Note that the difference between the CKA and
Procrustes similarity scores in~\Figref{fig:reg-encoder-comp} remains very small (around 0.1) indicating consistent
results between both metrics.


    \section{CKA on fully-connected architectures}\label{sec:app-fc}
In order to assess the generalisability of our findings, we have repeated our observations on the fully-connected VAEs that are described in~\Appref{sec:xp-setup}.
We can see in~\Threefigref{fig:lep}{fig:lreg}{fig:methods-linear} that the same general trend as for the convolutional architectures can be identified.

\subsubsection*{Learning in fully-connected VAEs is also bottom-up}
\begin{figure}
    \centering
    \subcaptionbox{Epoch 1\label{fig:lep-1}}{
        \includegraphics[width=0.33\textwidth]{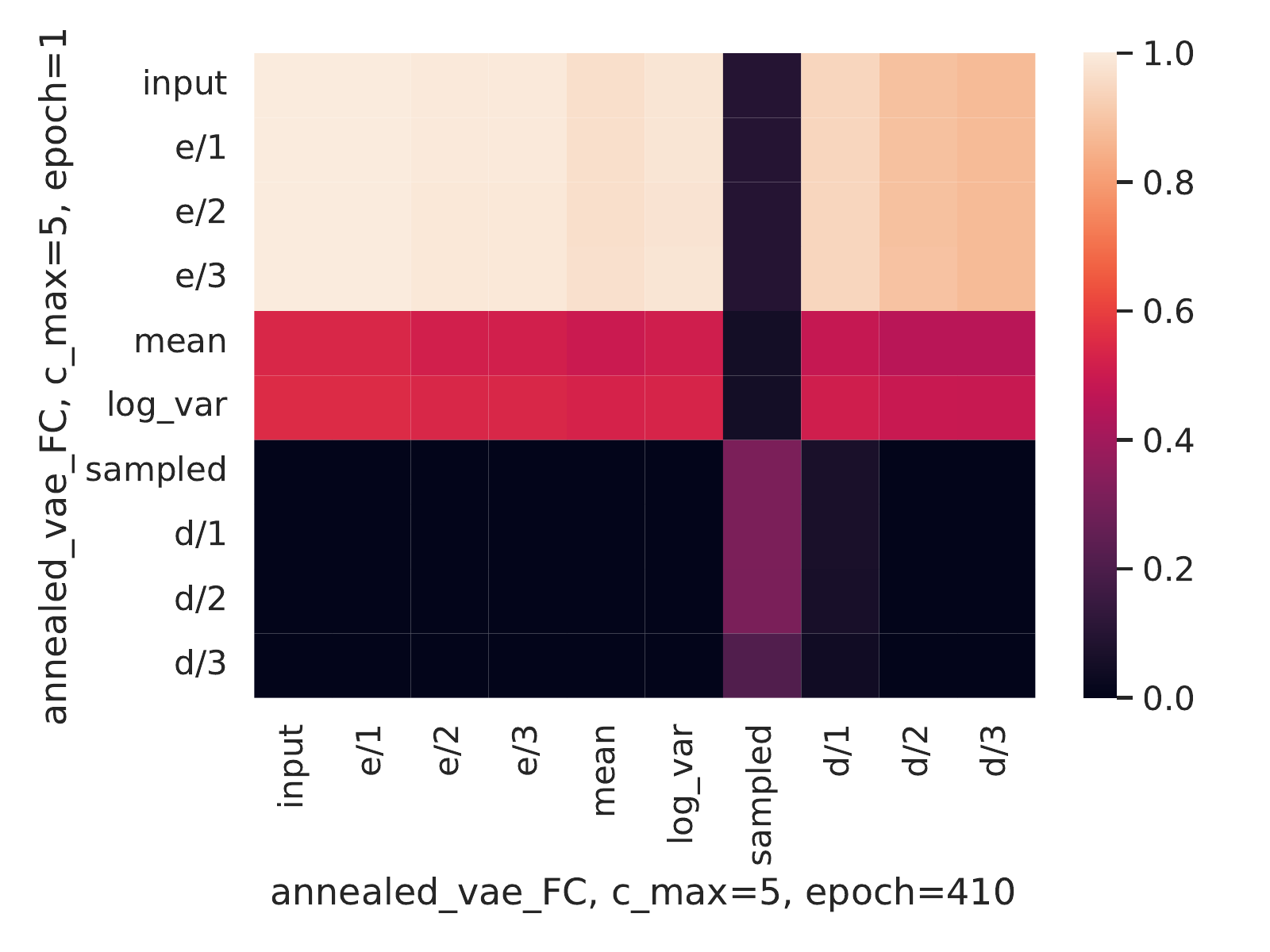}
    }%
    \subcaptionbox{Epoch 10\label{fig:lep-2}}{
        \includegraphics[width=0.33\textwidth]{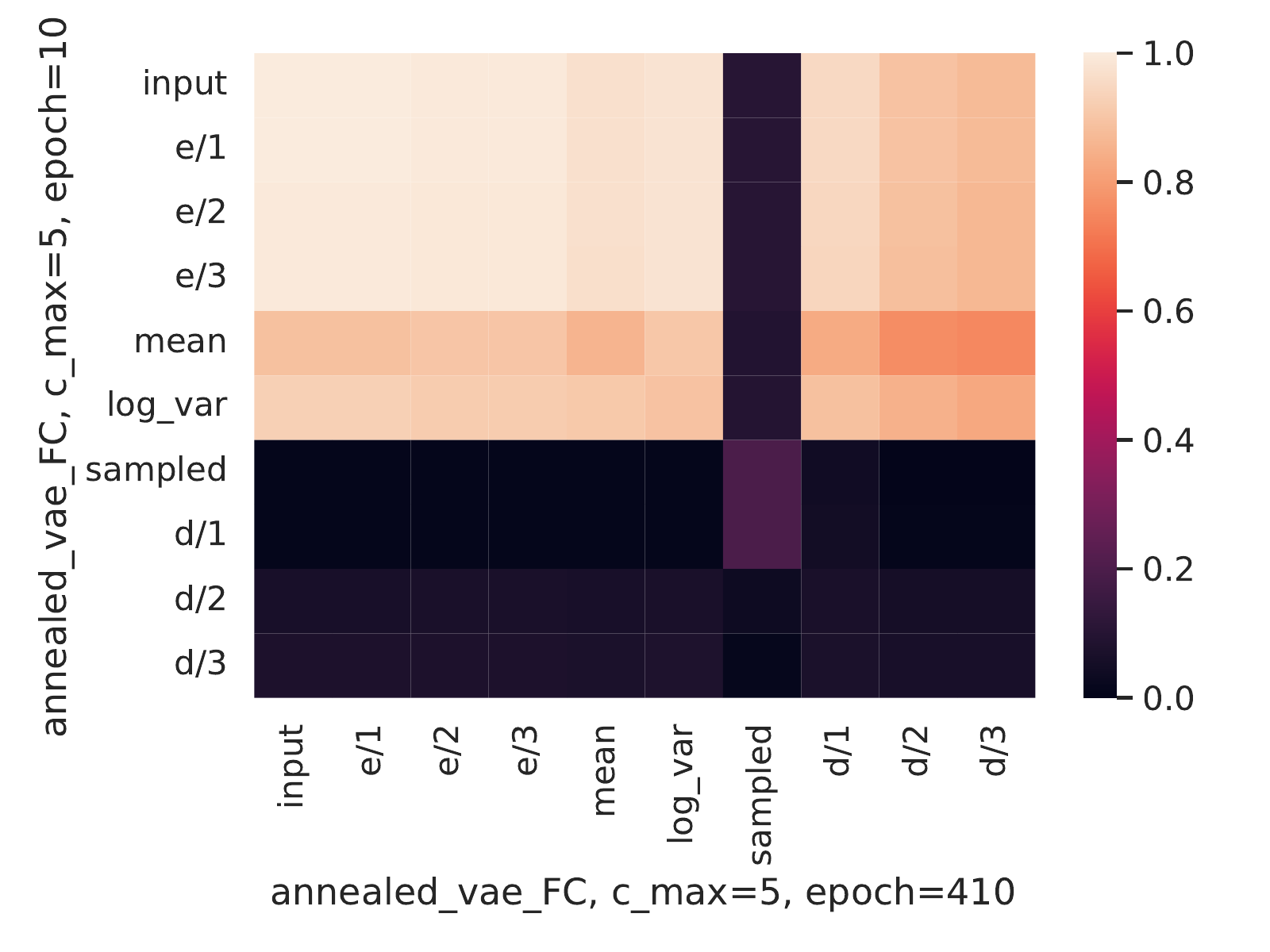}
    }%
    \subcaptionbox{Epoch 410\label{fig:lep-3}}{
        \includegraphics[width=0.33\textwidth]{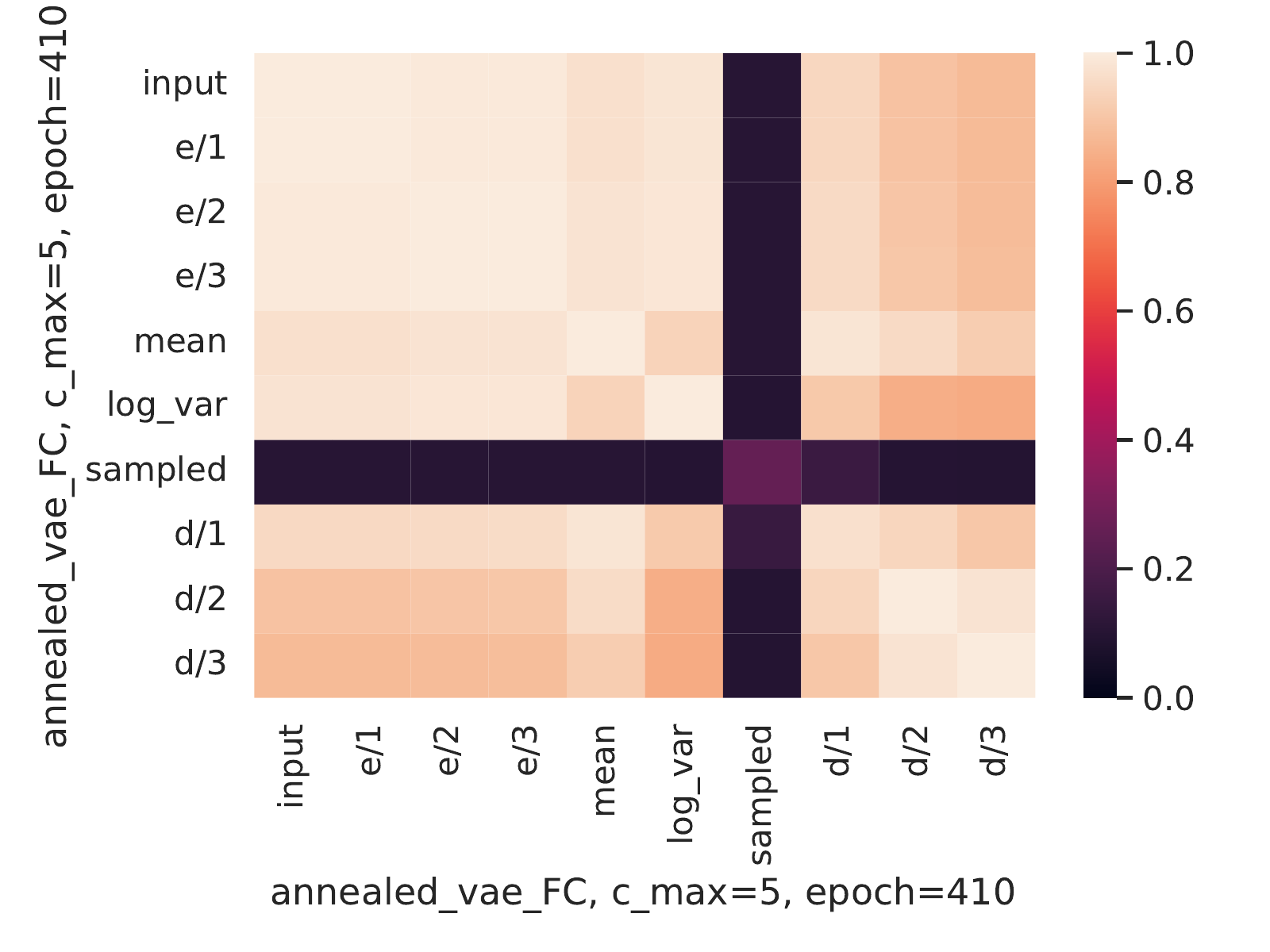}
    }%
    \caption{(a), (b), and (c) show CKA scores between a fully-trained fully-connected Annealed VAE and a fully-connected Annealed VAE trained
    for 1, 10 and 410 epochs, respectively. All the models are trained on SmallNorb and the results are averaged over 5 seeds.
     As in~\Figref{fig:fact1} of~\Secref{subsec:cka-check}, we can see that there is
        a high similarity between the representations learned by the encoder early in the training and after
        complete training (see the bright cells in the top-left quadrants in (a), (b), and (c)). The
        mean and variance representations similarity with a fully trained model increase after a few more epochs (the purple line in the middle disappear between (a) and (b)),
        and finally the decoder is learned (see bright cells in bottom-right quadrant of (c)).}
    \label{fig:lep}
\end{figure}

We can see in~\Figref{fig:lep} that, similarly to the convolutional architectures shown in \Figref{fig:fact1}, the encoder is learned early in the
training process.~Indeed between epochs 1 and 10, the encoder representations become highly similar to the representations of the
fully trained model (see~\Twofigref{fig:lep-1}{fig:lep-2}).~The decoder is then learned with its representational similarity
with the fully trained decoder raising after epoch 10 (see~\Figref{fig:lep-3}).

\subsubsection*{Impact of regularisation}
\begin{figure}
    \centering
    \subcaptionbox{$\beta = 8$\label{fig:lreg-1}}{
        \includegraphics[width=0.33\textwidth]{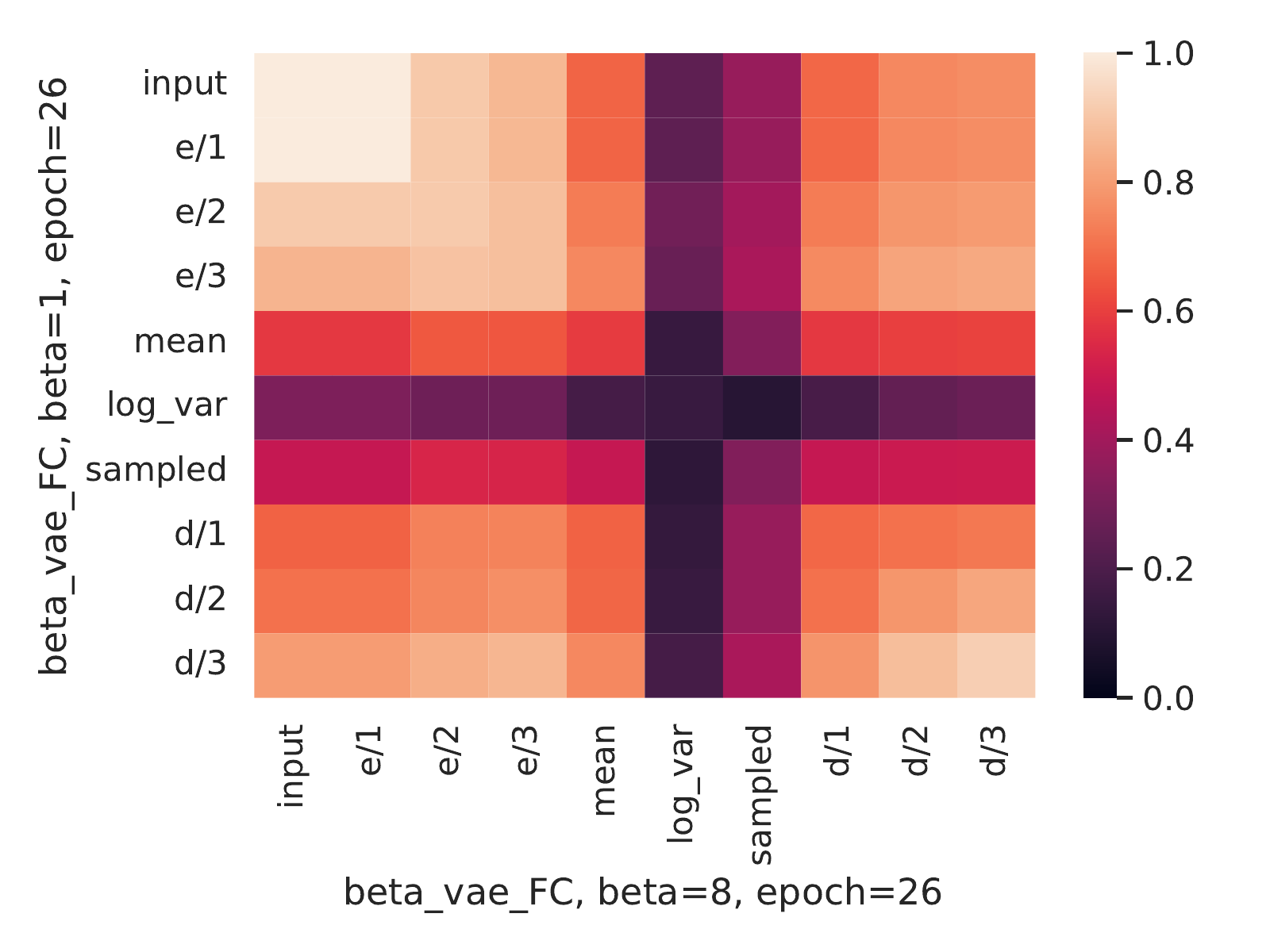}
    }%
    \subcaptionbox{$\beta = 16$\label{fig:lreg-2}}{
        \includegraphics[width=0.33\textwidth]{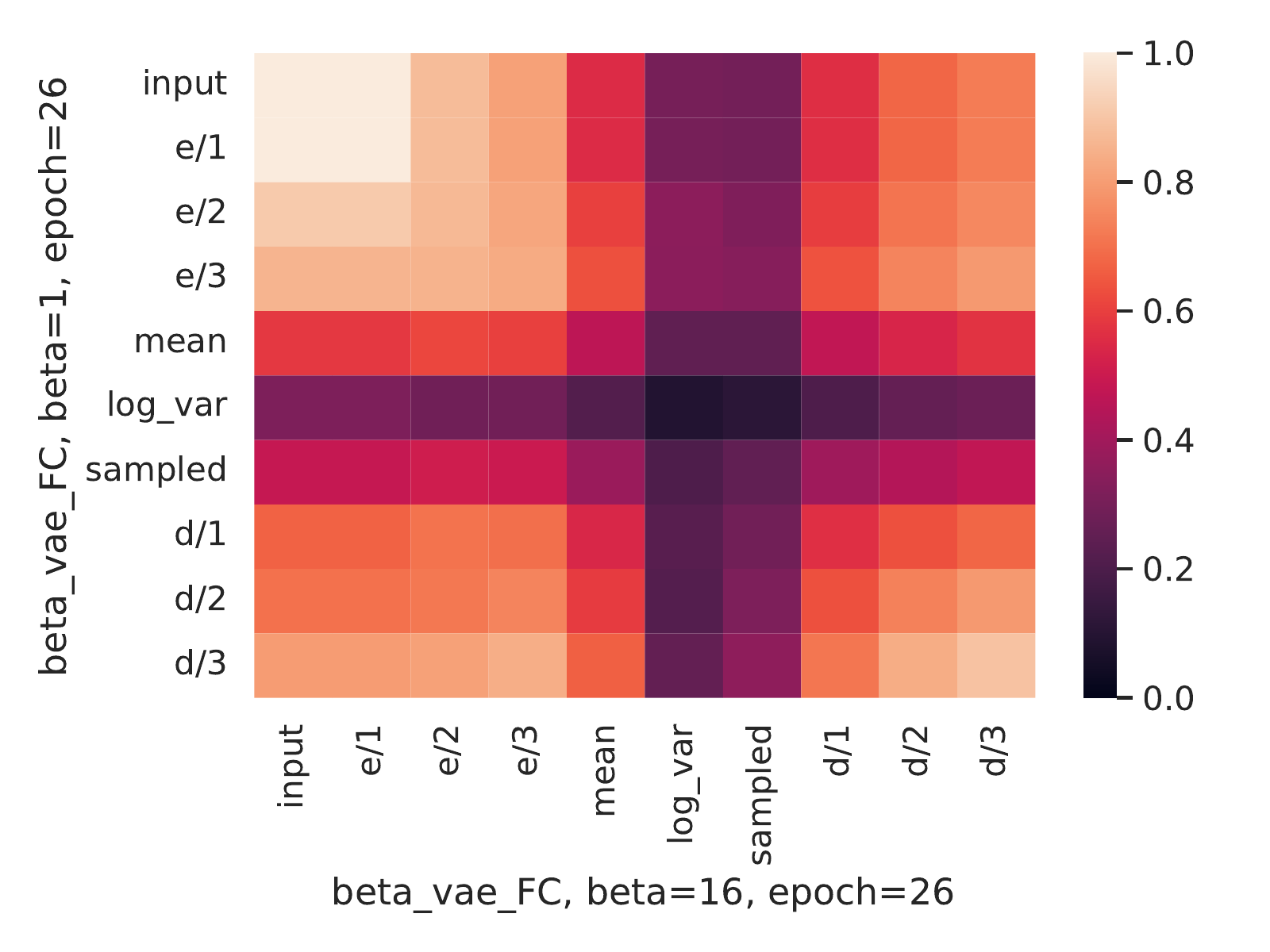}
    }
    \caption{(a) and (b) show the representational similarity between fully-connected $\beta$-VAEs trained with $\beta = 1$,
        and fully-connected $\beta$-VAEs trained with $\beta = 8$ and $\beta = 16$, respectively. All models are trained on dSprites and
        the scores are averaged over 5 seeds. In both figures, the encoder representations stay very similar (bright cells in the top-left quadrants),
        except for the mean, variance and sampled representations. While the variance representation is increasingly different as we increase $\beta$, the
        decoder does not show the dramatic dissimilarity observed in convolutional architectures in~\Figref{fig:fact2}.}
    \label{fig:lreg}
\end{figure}

As in convolutional architectures shown in~\Figref{fig:fact2}, the variance and sampled representations retain little similarity with the encoder representations
in the case of posterior collapse, as shown in~\Figref{fig:lreg}.~Interestingly, in fully-connected architectures the decoder retains more
similarity with its less regularised version than in convolutional architectures, despite suffering from poor reconstruction when heavily regularised.
Thus, CKA of the representations of fully-connected decoders may not be a good predictor of reconstruction quality.

\subsubsection*{Impact of learning objective}
\begin{figure}[ht!]
    \centering
    \subcaptionbox{Trained on Cars3D\label{fig:lmethods-cars}}{
        \includegraphics[width=0.33\textwidth]{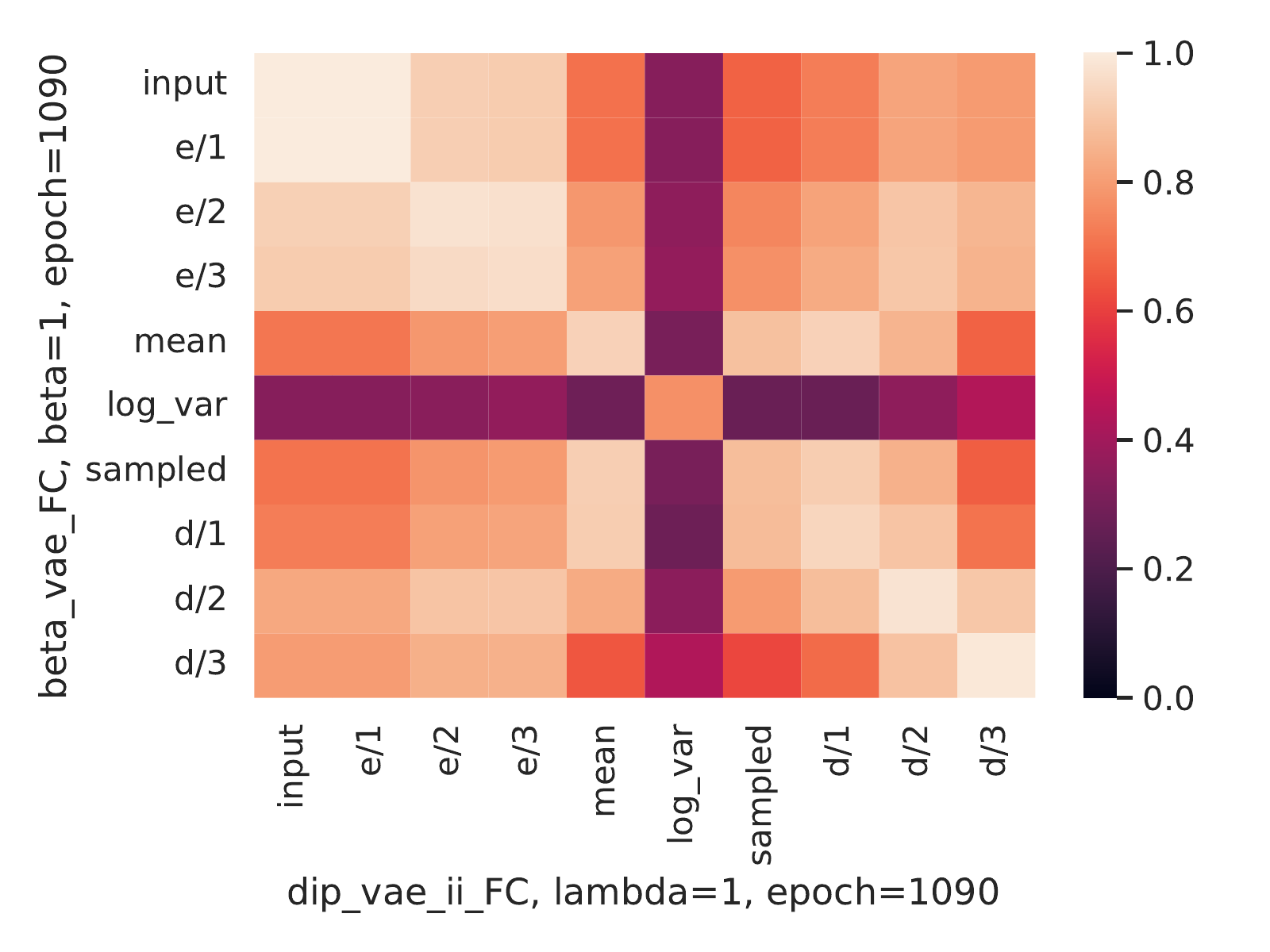}
    }%
    \subcaptionbox{Trained on dSprites\label{fig:lmethods-dsprites}}{
        \includegraphics[width=0.33\textwidth]{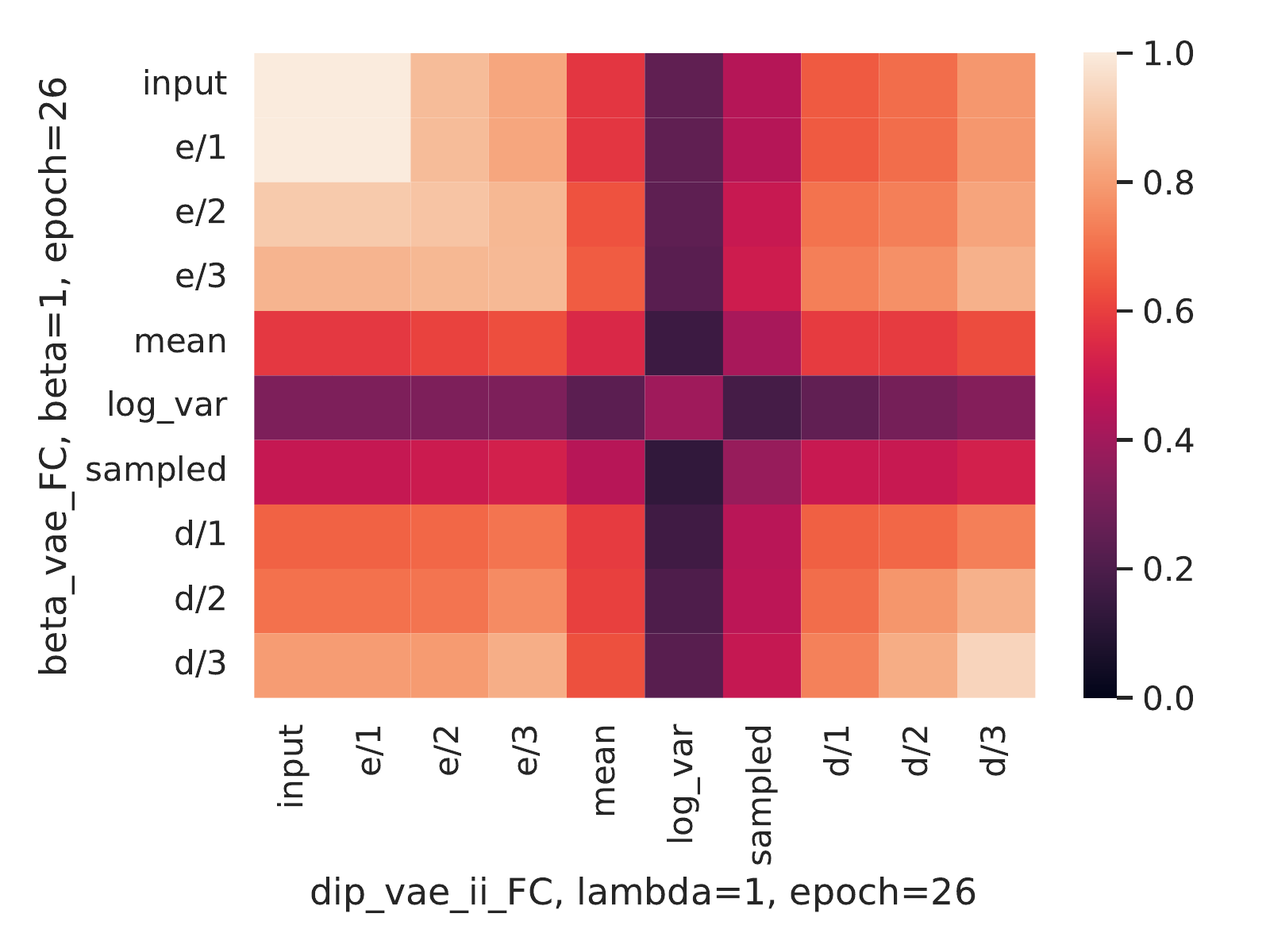}
    }%
    \subcaptionbox{Trained on SmallNorb\label{fig:lmethods-smallnorb}}{
        \includegraphics[width=0.33\textwidth]{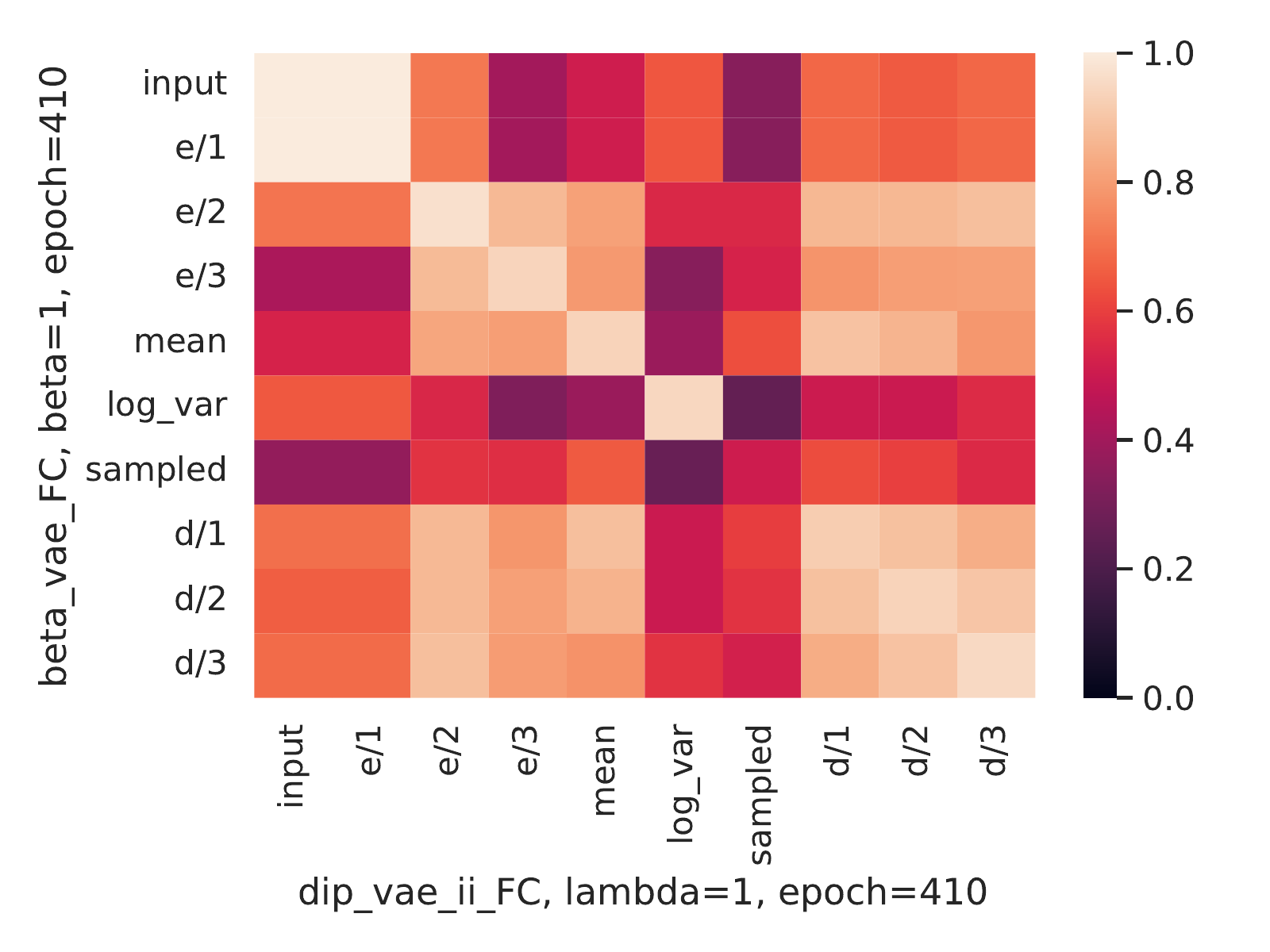}
    }%
    \caption{(a) shows the CKA similarity scores of activations of $\beta$-VAE and DIP-VAE II with fully-connected architectures trained on Cars3D with $\beta=1$, and $\lambda=1$, respectively.
             (b) and (c) show the CKA similarity scores of the same learning objectives and regularisation strengths but trained on dSprites and SmallNorb.
             All the results are averaged over 5 seeds.
             We can see that the representational similarity of all the layers of the encoder (top-left quadrant) except mean and variance is high (CKA $\geqslant 0.8$).
             However, as in~\Figref{fig:fact3}, the mean, variance, sampled (center diagonal values), and decoder (bottom-right quadrants) representational similarity of different learning objectives seems to vary depending on the dataset.
             In (a) and (c) they have high similarity, while in (b) the similarity is lower. Moreover, in (c) the input and first layer of the encoder are quite distinct from the
             other representations, which was not the case in convolutional architectures.
    }
    \label{fig:methods-linear}
\end{figure}

\Figref{fig:methods-linear} provides results similar to the convolutional VAEs observed in~\Figref{fig:fact3}, with a very high similarity between encoder layers
learned from different learning objectives (see diagonal values of the upper-left quadrant).~Here again, the representational similarity
of the decoder seems to vary depending on the dataset, even though this is less marked than for convolutional architectures.
We can also see that the representational similarity between different layers of the encoder vary depending on the dataset,
which was less visible in convolutional architectures.~For example, the similarity between the first and subsequent
layers of the encoder in SmallNorb is much lower in fully-connected VAEs. Given that SmallNorb is a hard dataset to learn for VAEs~\citep{Locatello2019a},
one could hypothesise that the encoder of fully-connected VAE, being less powerful, is unable to retain as much information as its convolutional counterpart,
leading to lower similarity scores with the representations of the first encoder layer.

    \section{How similar are the representations learned by encoders and classifiers?}\label{sec:app-clf}
\begin{figure}[ht!]
    \centering
    \subcaptionbox{Trained on Cars3D\label{fig:clf-cars}}{
        \includegraphics[width=0.33\textwidth]{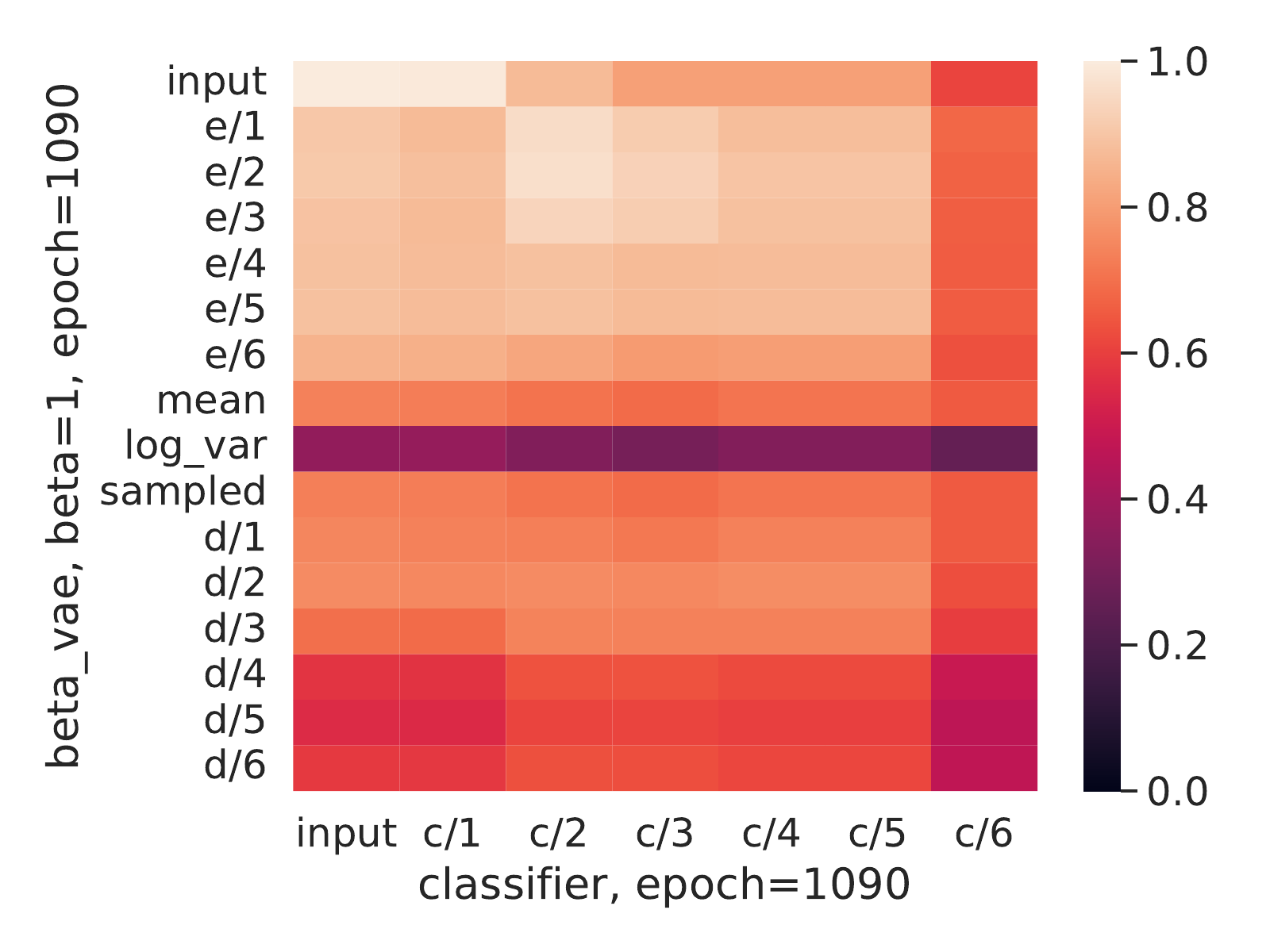}
    }%
    \subcaptionbox{Trained on dSprites\label{fig:clf-dsprites}}{
        \includegraphics[width=0.33\textwidth]{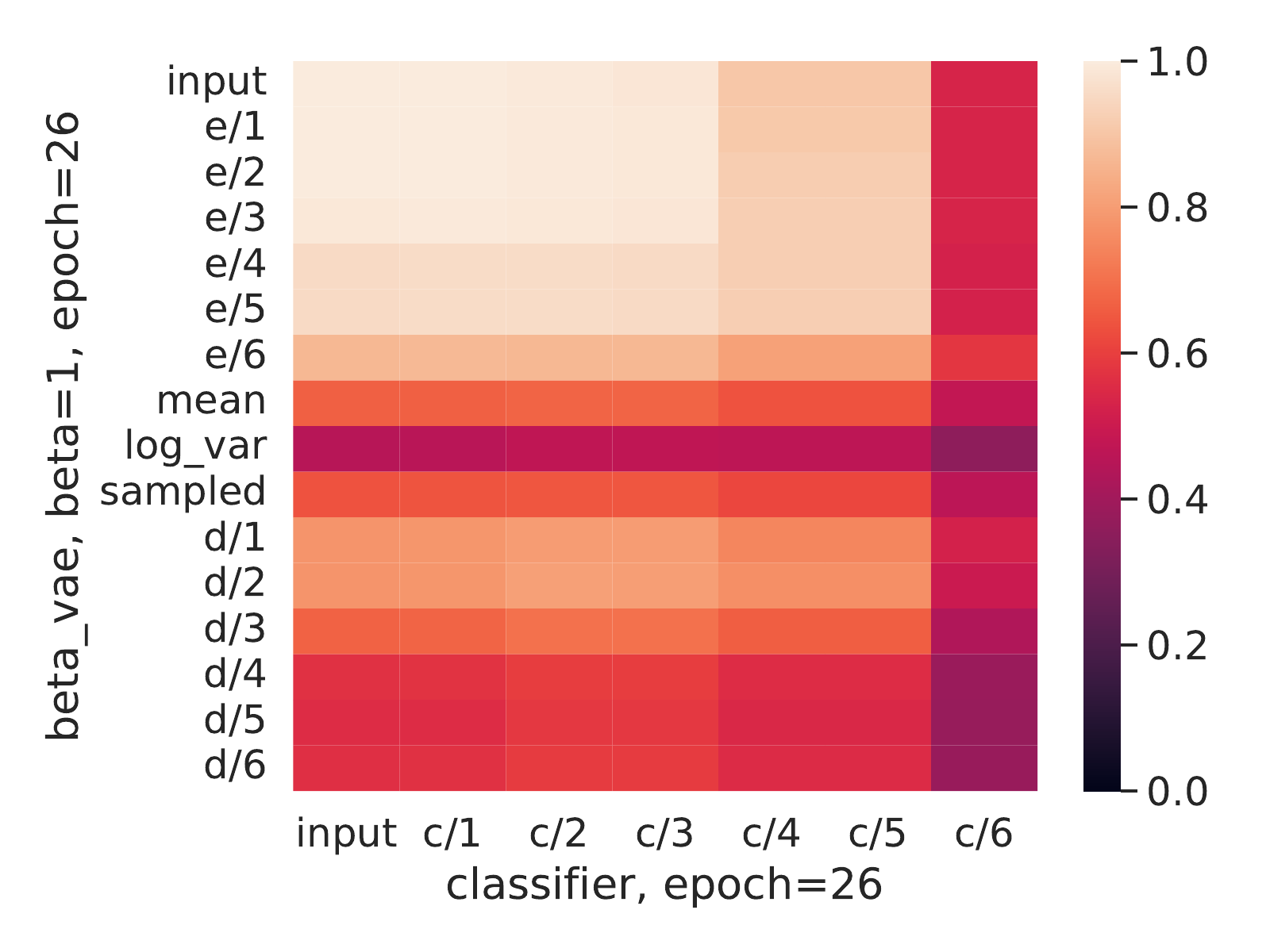}
    }%
    \subcaptionbox{Trained on SmallNorb\label{fig:clf-smallnorb}}{
        \includegraphics[width=0.33\textwidth]{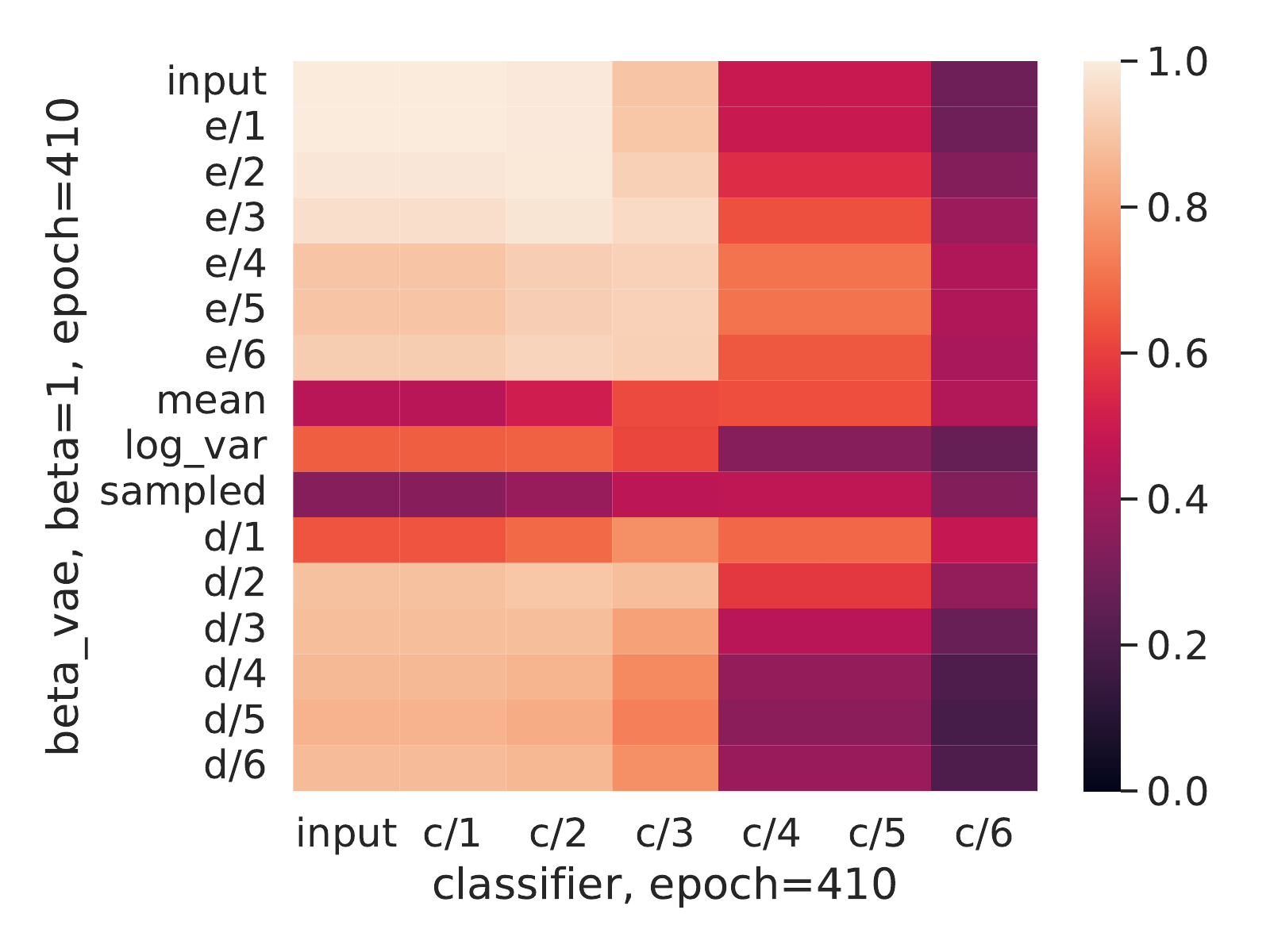}
    }%
    \caption{(a) shows the CKA similarity scores of activations of a classifier and a $\beta$-VAE trained on Cars3D with $\beta=1$.
             (b) and (c) show the CKA similarity scores of the same learning objectives and regularisation strengths but trained on dSprites and SmallNorb.
             All the results are averaged over 5 seeds.
             We can see that the representational similarity between the layers of the classifiers and of the encoder (top-left quadrant) except mean and variance is very high (CKA stays close to 1).
             However, the mean, variance, and sampled representations (central diagonal values) are different from the representations learned by the classifier.
    }
    \label{fig:clf}
\end{figure}

To compare VAEs with classifiers, we used the convolutional architecture of an encoder for classification, replacing
the mean and variance layers by the final classifier layers.
As shown in~\Figref{fig:clf}, we obtain a high representational similarity when comparing VAEs and classifiers indicating,
consistently with the observations of~\cite{Yosinski2015}, that classifiers seem to learn generative features.
This explains why encoders based on pre-trained classifier architectures such as VGG have empirically demonstrated
good performances~\citep{Liu2021} and also suggests that the weights of the pre-trained architecture could be used as-is
without further updates. This also indicates that using pre-trained encoders may be beneficial in the context of transfer
learning, domain adaptation~\citep{Pan2009}, or simply reconstruction quality~\citep{Liu2021} which is consistent with
our results in~\Secref{subsec:cka-tl}.

    \section{Representational similarity of VAEs at different epochs}\label{sec:app-epochs}

The results obtained in~\Secref{subsec:cka-check} have shown a high similarity between the encoders at an early stage of training and fully trained.
One can wonder whether these results are influenced by the choice of epochs used in~\Figref{fig:fact1}.
After explaining our epoch selection process, we show below that it does not influence our results, which are consistent over snapshots taken at
different stages of training.

\subsubsection*{Epoch selection} For dSprites, we took snapshots of the models at each epoch, but for Cars3D and SmallNorb, which both train for a
higher number of epochs, it was not feasible computationally to calculate the CKA between every epoch.~We thus
saved models trained on SmallNorb every 10 epochs, and models trained on Cars3D every 25 epochs.
Consequently, the epochs chosen to represent the early training stage in~\Secref{subsec:cka-check} is always the first snapshot taken for each model. Below, we preform additional experiments with a broader range of epoch numbers to show that the results are consistent with our findings in the main paper, and they do not depend on specific epochs.

\subsubsection*{Similarity changes over multiple epochs}
\begin{figure}[ht!]
    \centering
    \subcaptionbox{Epoch 2\label{fig:dsprites-epoch-1}}{
        \includegraphics[width=0.5\textwidth]{heatmaps/dsprites/dip_vae_ii_5_epoch_2_dip_vae_ii_5_epoch_26}
    }%
    \subcaptionbox{Epoch 7\label{fig:dsprites-epoch-2}}{
        \includegraphics[width=0.5\textwidth]{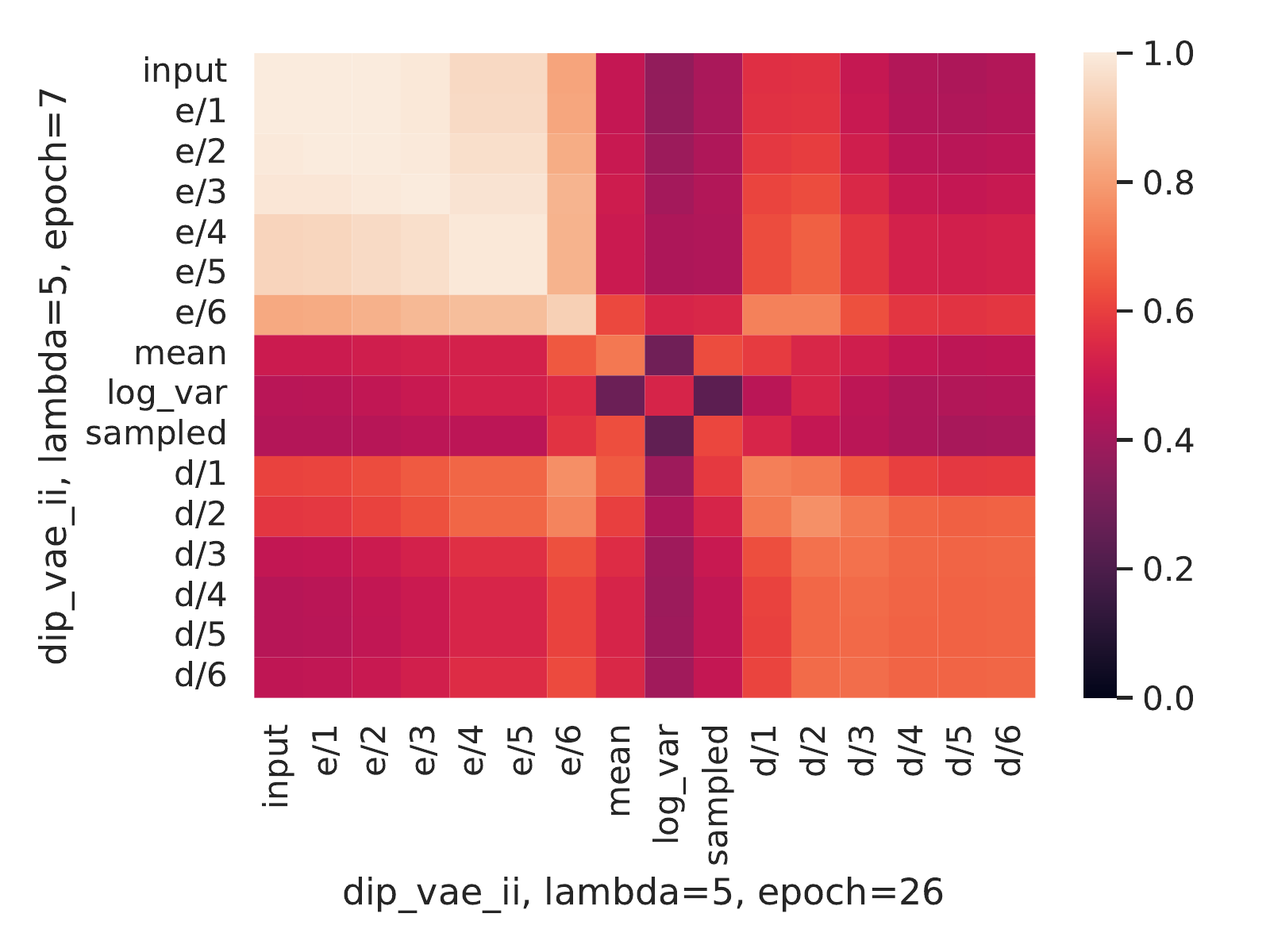}
    }\\
    \subcaptionbox{Epoch 14\label{fig:dsprites-epoch-3}}{
        \includegraphics[width=0.5\textwidth]{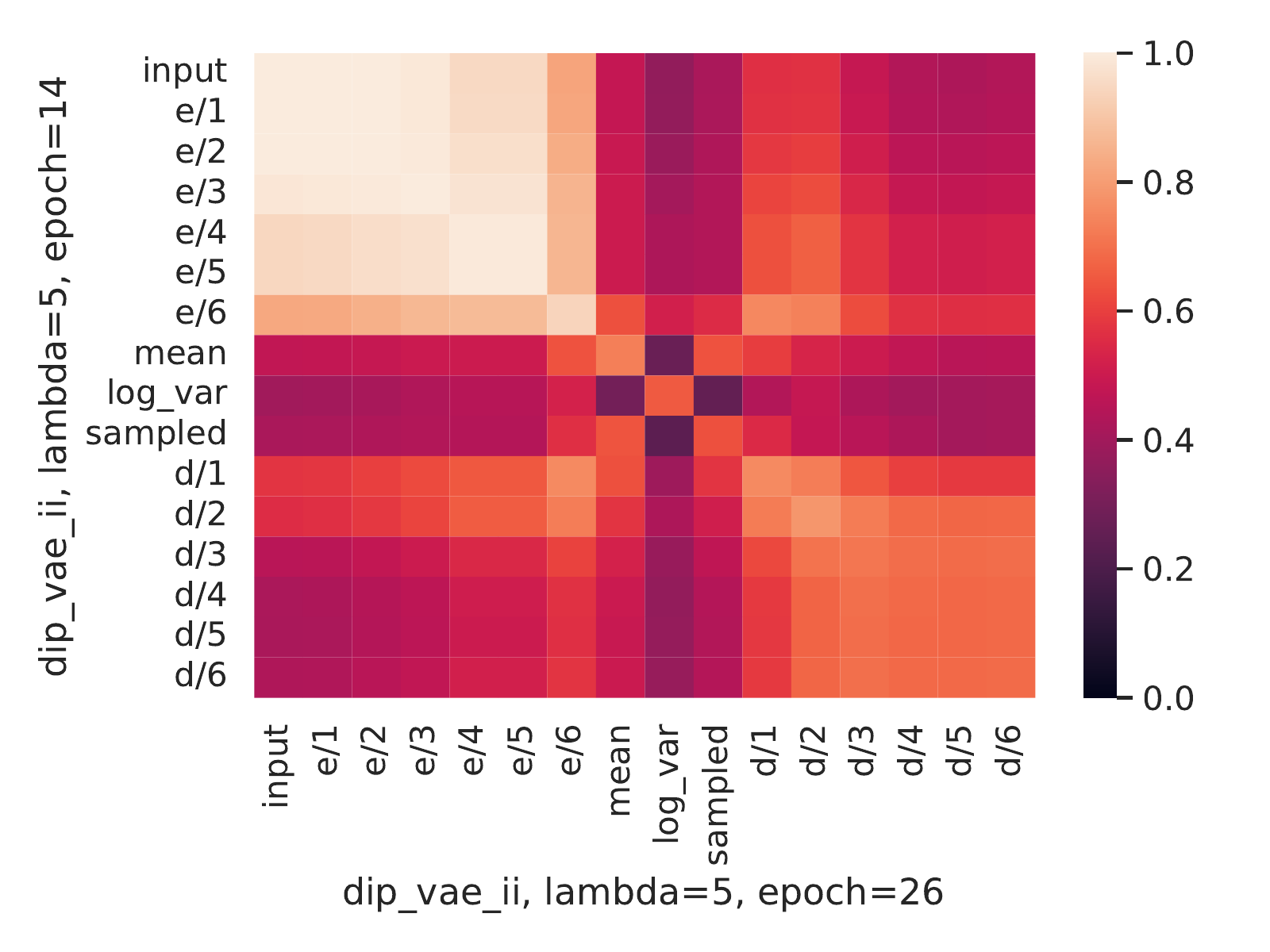}
    }%
    \subcaptionbox{Epoch 19\label{fig:dsprites-epoch-4}}{
        \includegraphics[width=0.5\textwidth]{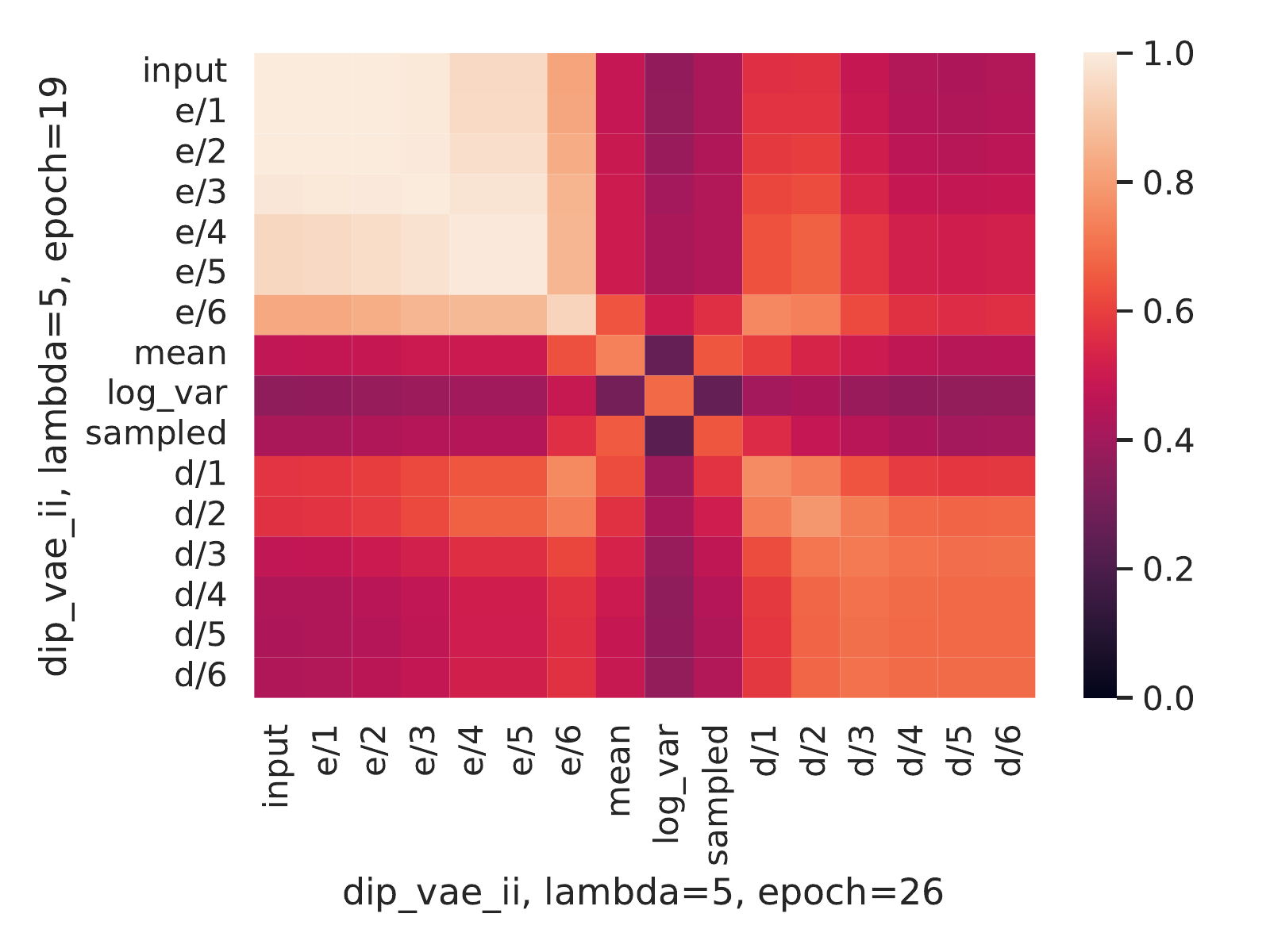}
    }\\
    \subcaptionbox{Epoch 26\label{fig:dsprites-epoch-5}}{
        \includegraphics[width=0.5\textwidth]{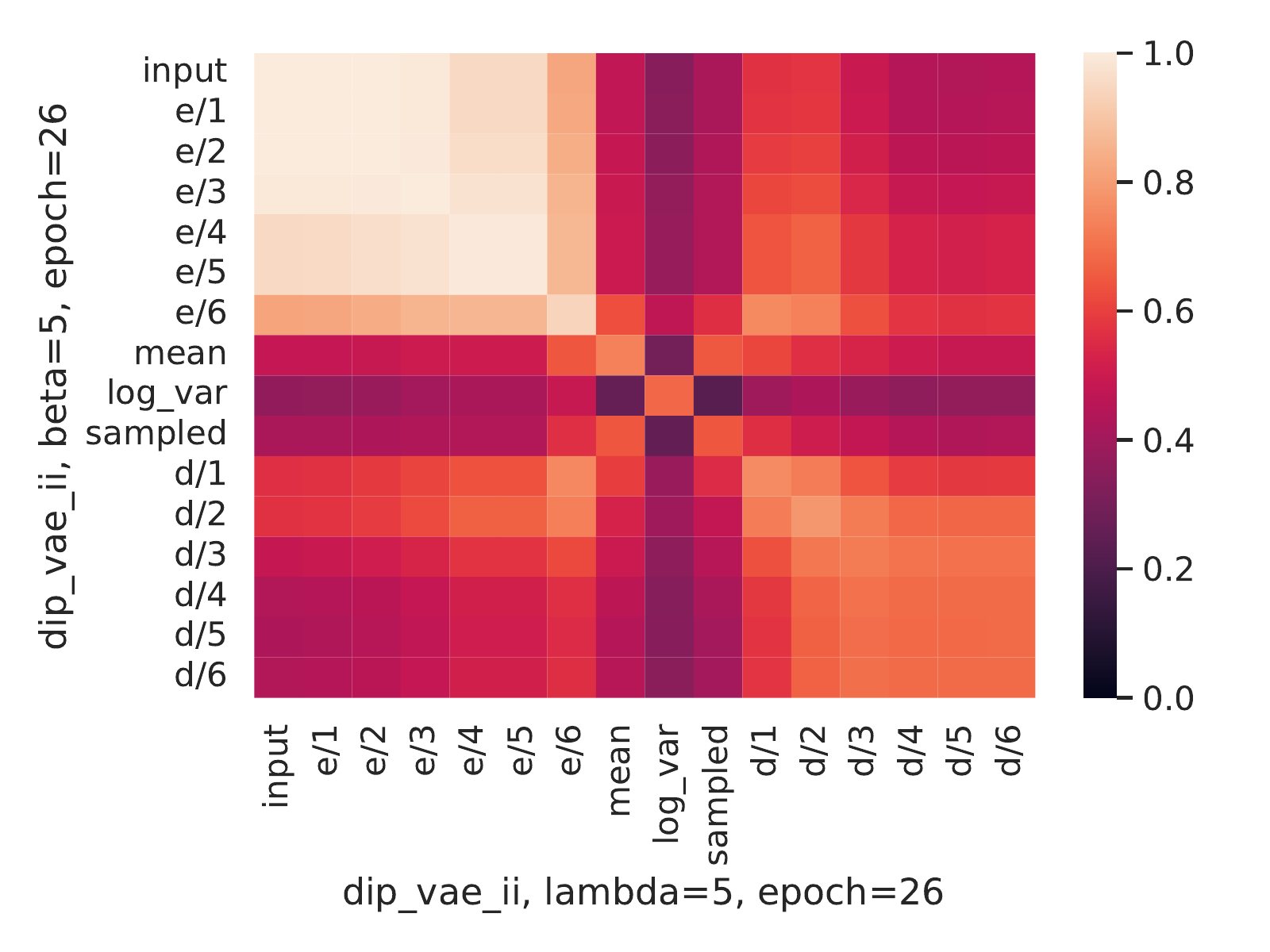}
    }\\
    \caption{(a), (b), (c), (d), and (e) show the representational similarity between DIP-VAE II after full training, and at epochs 2, 7, 14, 19, and 26, respectively.
    All models are trained on dSprites and the results are averaged over 5 runs.}
    \label{fig:dsprites-epochs}
\end{figure}

In~\Threefigref{fig:dsprites-epochs}{fig:cars-epochs}{fig:norb-epochs}, we can observe the same trend of learning
phases as in~\Secref{subsec:cka-check}.~First, the encoder is learned, as shown by the high representational similarity of the upper-left quadrant of
~\Threefigref{fig:dsprites-epoch-1}{fig:cars-epoch-1}{fig:norb-epoch-1}.
Then, the decoder is learned, as shown by the increased representational similarity of the bottom-right quadrant
of~\Threefigref{fig:dsprites-epoch-2}{fig:cars-epoch-2}{fig:norb-epoch-2}.
Finally, further small refinements of the encoder and decoder representations take place in the remaining training time,
as shown by the slight increase of representational similarity in~\Threefigref{fig:dsprites-epoch-3}{fig:cars-epoch-3}{fig:norb-epoch-3},
and~\Threefigref{fig:dsprites-epoch-4}{fig:cars-epoch-4}{fig:norb-epoch-4}.

\begin{figure}[ht!]
    \centering
    \subcaptionbox{Epoch 25\label{fig:cars-epoch-1}}{
        \includegraphics[width=0.5\textwidth]{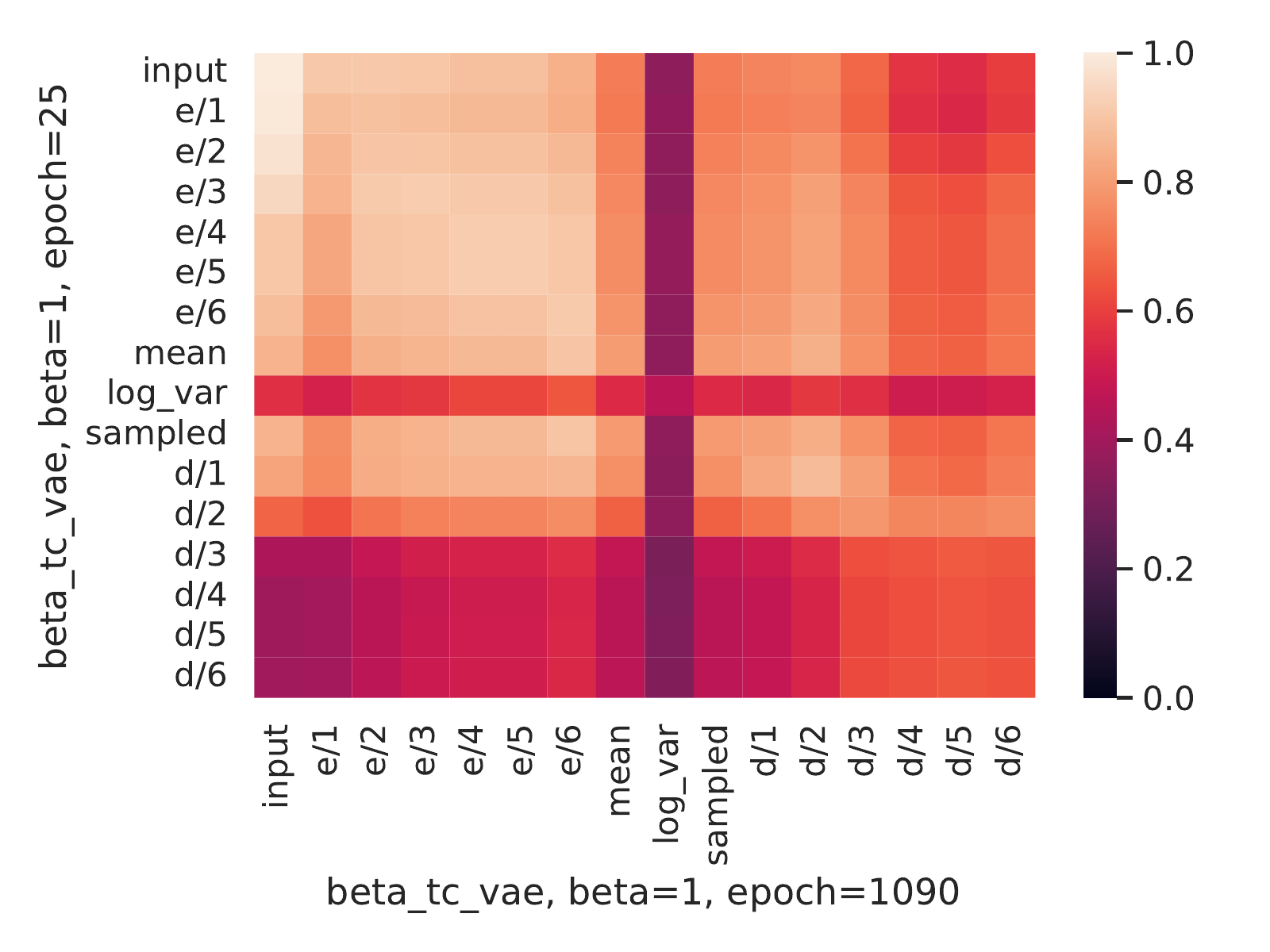}
    }%
    \subcaptionbox{Epoch 292\label{fig:cars-epoch-2}}{
        \includegraphics[width=0.5\textwidth]{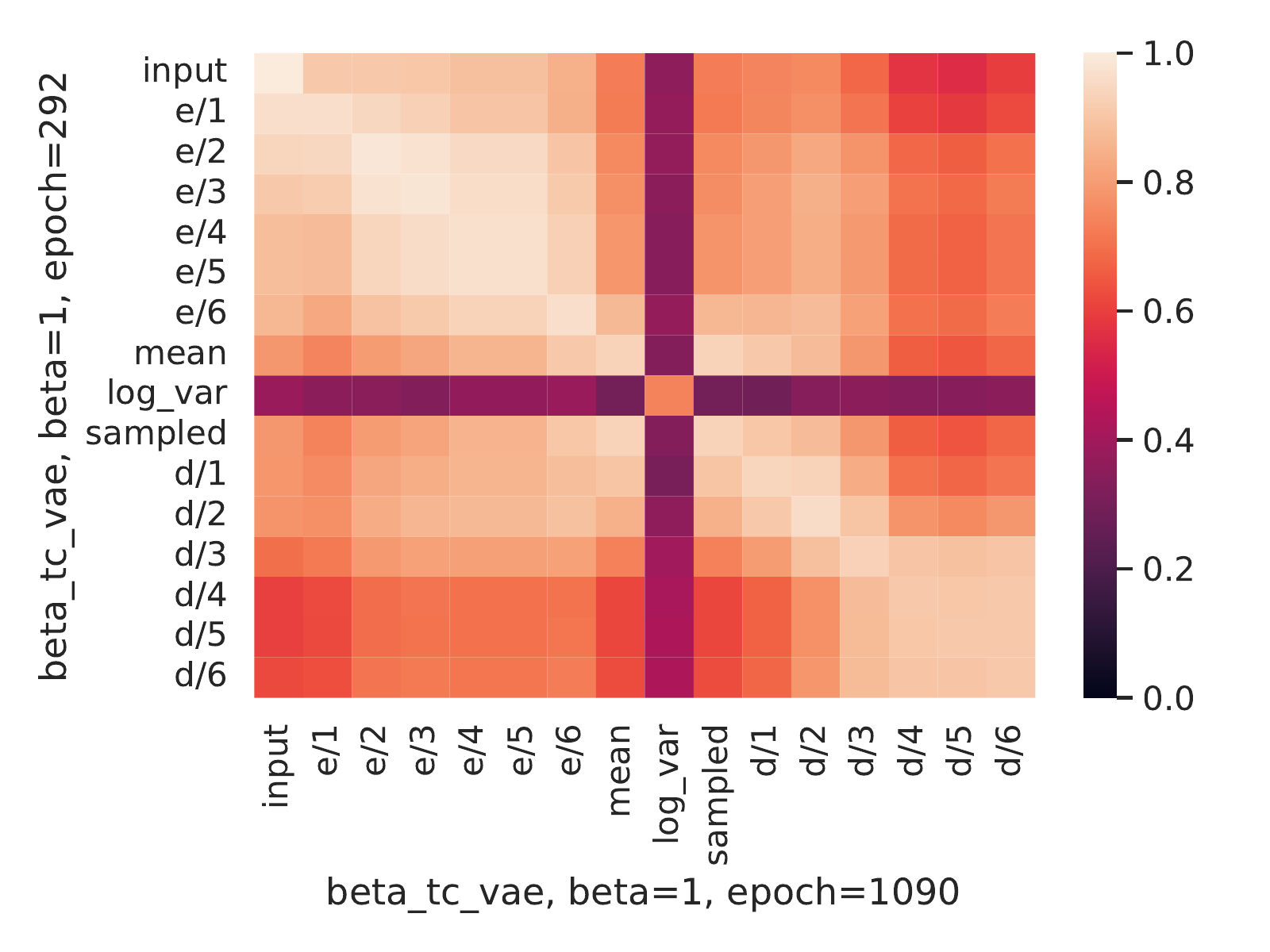}
    }\\
    \subcaptionbox{Epoch 559\label{fig:cars-epoch-3}}{
        \includegraphics[width=0.5\textwidth]{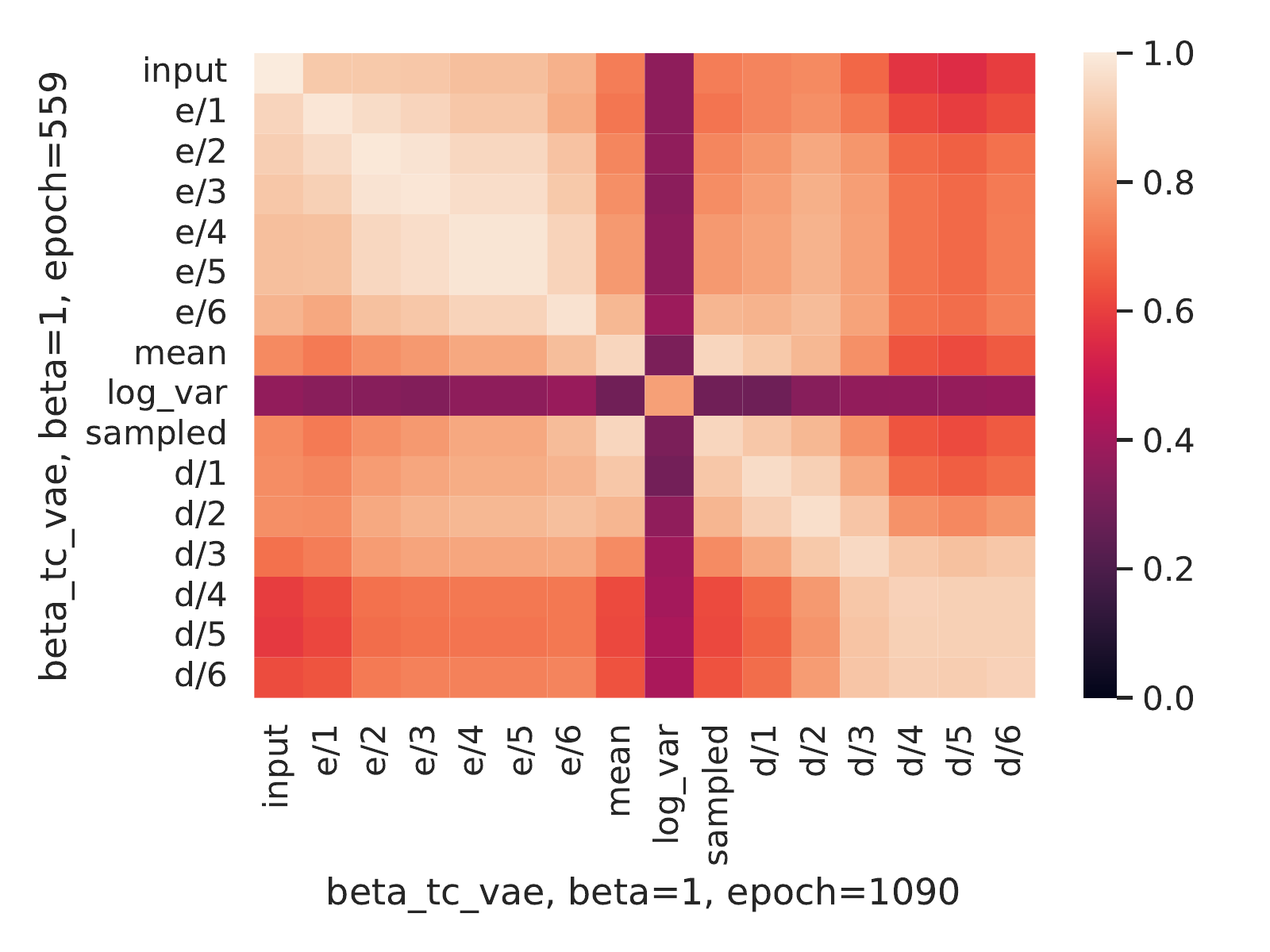}
    }%
    \subcaptionbox{Epoch 826\label{fig:cars-epoch-4}}{
        \includegraphics[width=0.5\textwidth]{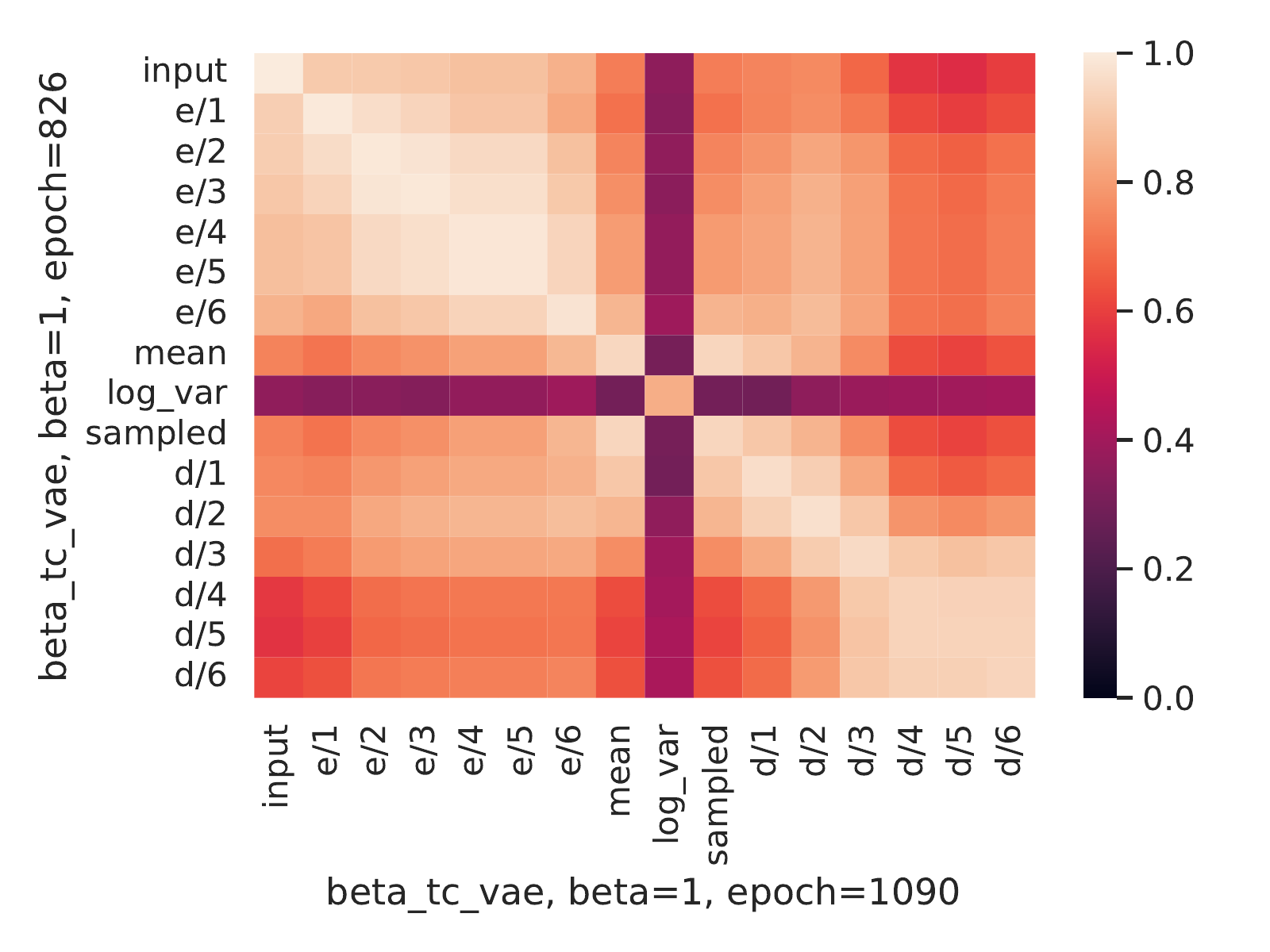}
    }\\
    \subcaptionbox{Epoch 1090\label{fig:cars-epoch-5}}{
        \includegraphics[width=0.5\textwidth]{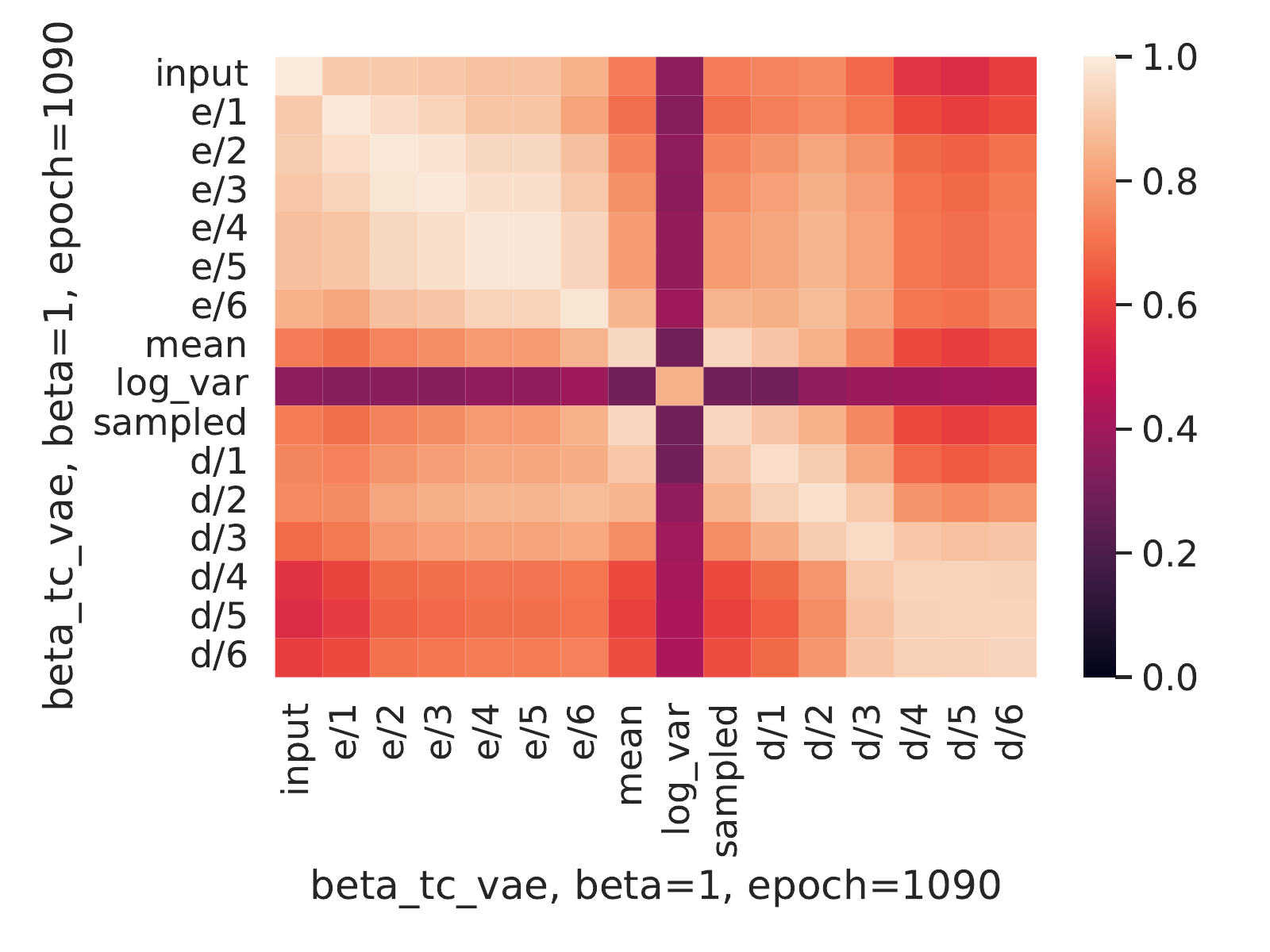}
    }
    \caption{(a), (b), (c), (d) and (e) show the representational similarity between $\beta$-TC VAE after full training, and at epochs 25, 292, 559, 826, and 1090 respectively.
    All models are trained on Cars3D and the results are averaged over 5 runs.}
    \label{fig:cars-epochs}
\end{figure}

\begin{figure}[ht!]
    \centering
    \subcaptionbox{Epoch 10\label{fig:norb-epoch-1}}{
        \includegraphics[width=0.5\textwidth]{heatmaps/smallnorb/annealed_vae_5_epoch_10_annealed_vae_5_epoch_410}
    }%
    \subcaptionbox{Epoch 110\label{fig:norb-epoch-2}}{
        \includegraphics[width=0.5\textwidth]{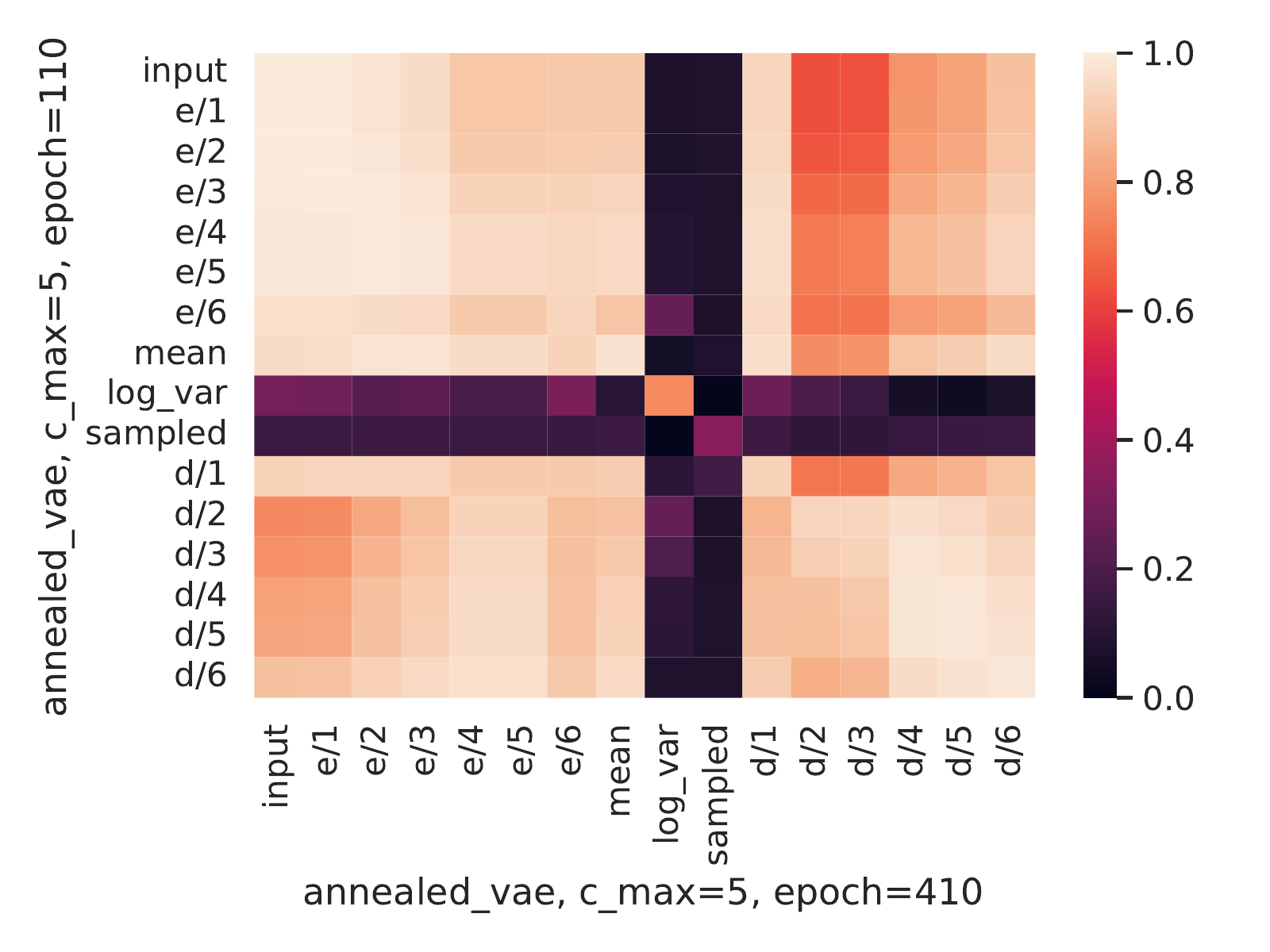}
    }\\
    \subcaptionbox{Epoch 210\label{fig:norb-epoch-3}}{
        \includegraphics[width=0.5\textwidth]{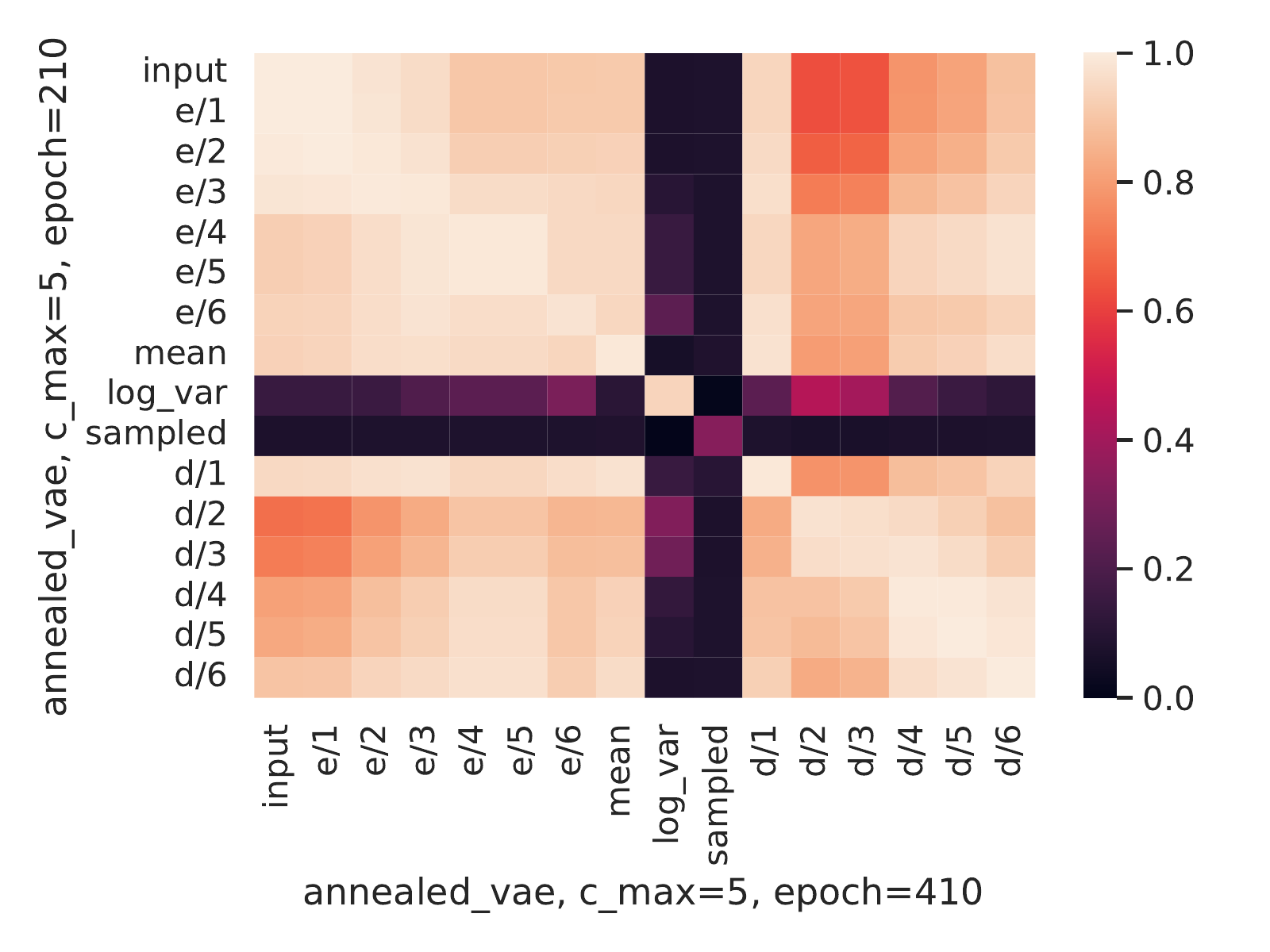}
    }%
    \subcaptionbox{Epoch 311\label{fig:norb-epoch-4}}{
        \includegraphics[width=0.5\textwidth]{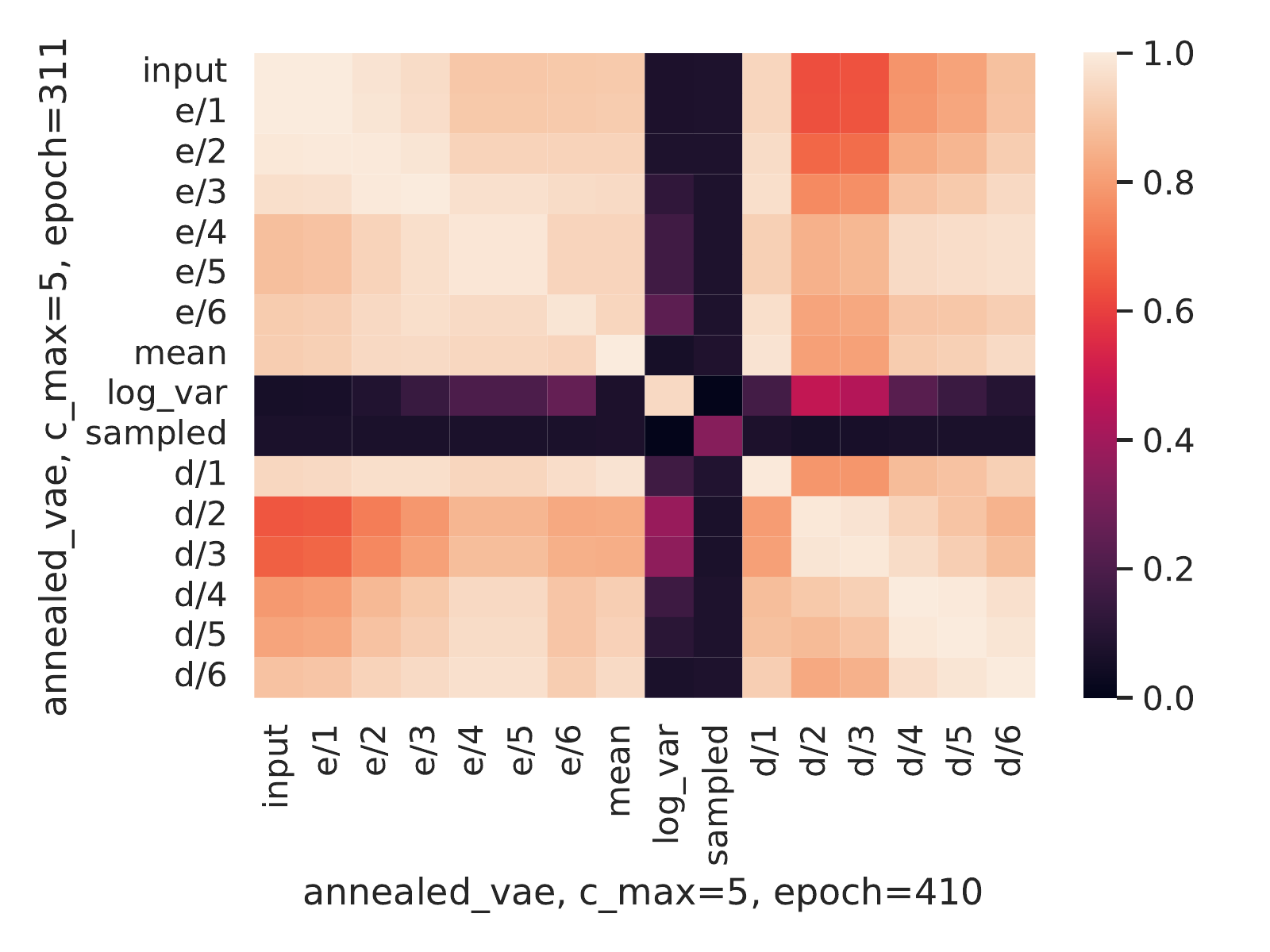}
    }\\
    \subcaptionbox{Epoch 311\label{fig:norb-epoch-5}}{
        \includegraphics[width=0.5\textwidth]{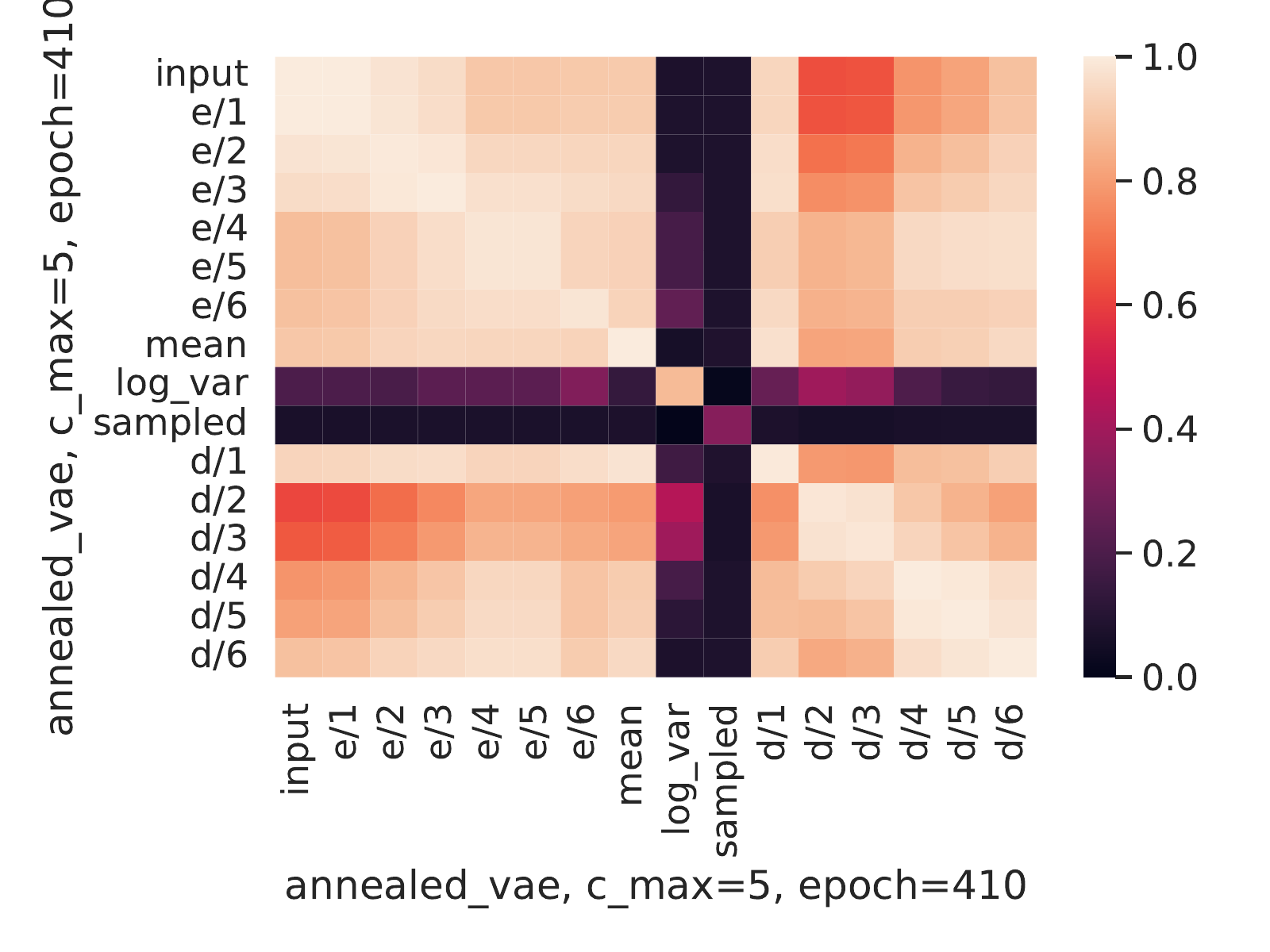}
    }
    \caption{(a), (b), (c), (d) and (e) show the representational similarity between Annealed VAE after full training, and at epochs 10, 110, 210, 311, and 410 respectively.
    All models are trained on SmallNorb and the results are averaged over 5 runs.}
    \label{fig:norb-epochs}
\end{figure}
    \clearpage
    \section{Convergence rate of different VAEs}\label{sec:app-convergence}
\begin{figure}[ht!]
    \centering
    \subcaptionbox{Convergence on Cars3D\label{fig:conv-cars}}{
        \includegraphics[width=\textwidth]{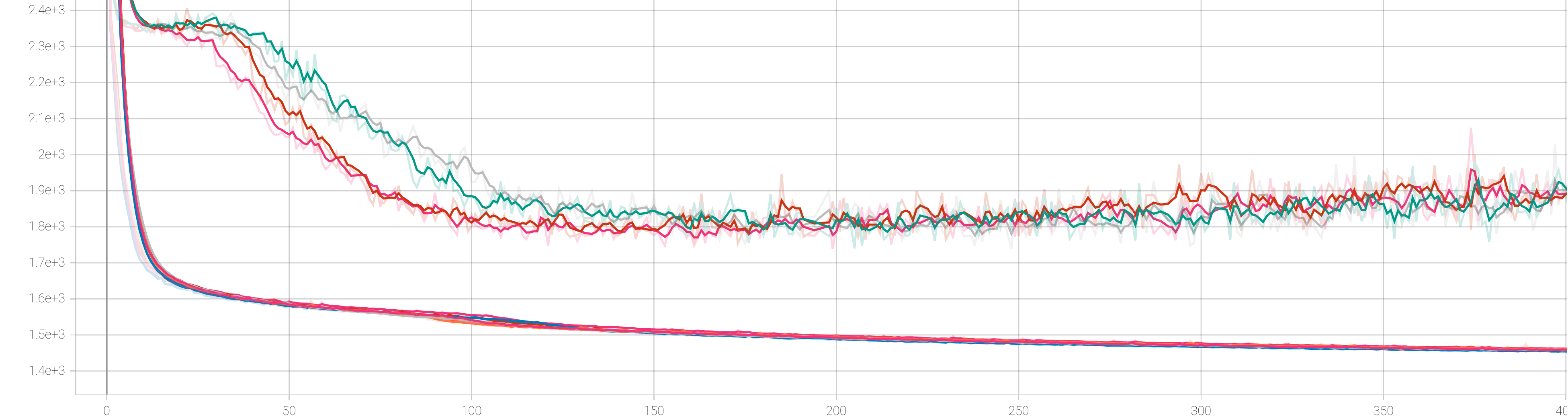}
    }\\
    \subcaptionbox{Convergence on dSprites\label{fig:conv-dsprites}}{
        \includegraphics[width=\textwidth]{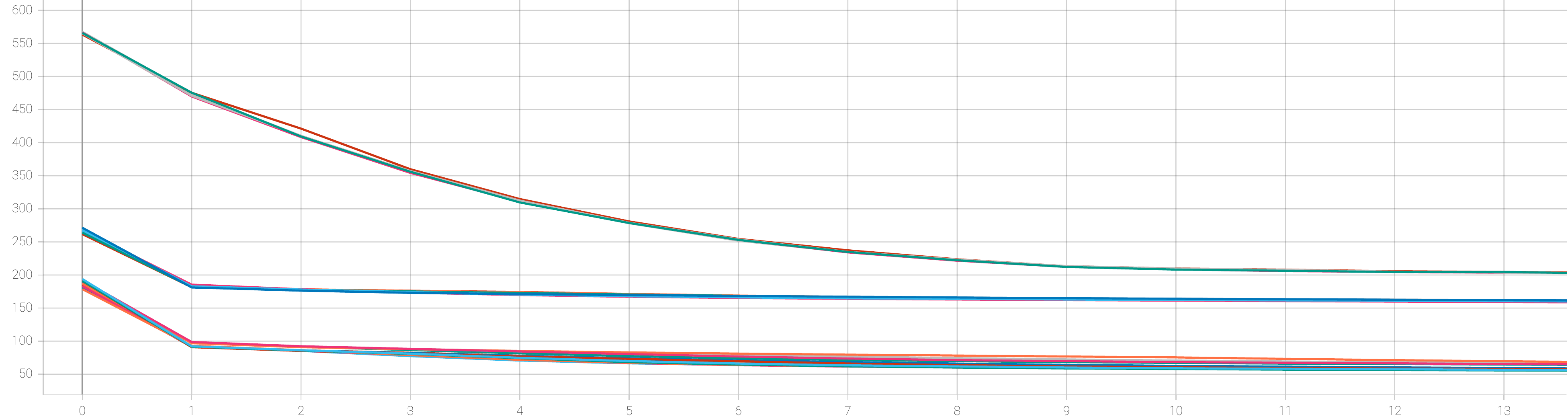}
    }\\
    \subcaptionbox{Convergence on SmallNorb\label{fig:conv-smallnorb}}{
        \includegraphics[width=\textwidth]{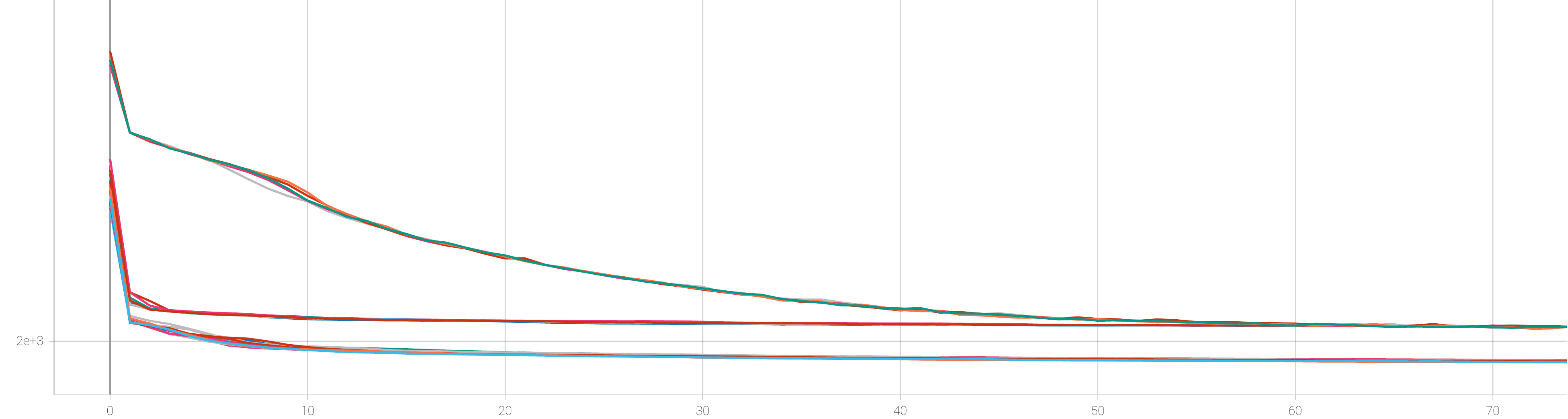}
    }\\
    \caption{ In (a), (b), and (c), we show the model loss of each model that converged when trained on Cars3D, dSprites, and SmallNorb, respectively.
    For each learning objective, we display 5 runs of the least and most regularised versions.
    }
    \label{fig:conv}
\end{figure}

We can see in~\Figref{fig:conv} that all the models converge at the same epoch, with less regularised models reaching lower losses.
While annealed VAEs start converging together with the other models, they then take longer to plateau, due to the annealing process.
We can see them distinctly in the upper part of~\Figref{fig:conv}.
Overall, the epochs at which the models start to converge are consistent with our choice of epoch for early training in~\Secref{subsec:cka-check}.

\end{document}